%% file: main.tex
\documentclass{article} % For LaTeX2e

\IfFileExists{latexml.sty}{\usepackage{latexml}}{}
\providecommand{\iflatexml}{\iffalse}
\iflatexml
  \usepackage{authblk}
\fi

\usepackage{preprint,times}

\input{math_commands.tex}

\usepackage{amsmath}
\usepackage{amssymb}
\usepackage{adjustbox}
\usepackage{array}
\usepackage{booktabs}
\usepackage{colortbl}
\usepackage{xcolor}
\usepackage{float}
\usepackage{graphicx}
\usepackage{multirow}
\usepackage{placeins}
\usepackage{tabularx}
\usepackage{titletoc}
\usepackage{hyperref}
\usepackage{microtype}
\usepackage[most]{tcolorbox}
\usepackage{url}

\titlecontents{lsection}[1.5em]{\addvspace{0.35em}\bfseries}
  {\contentslabel{1.5em}}{}{\titlerule*[0.5pc]{.}\contentspage}
\titlecontents{lsubsection}[4em]{}
  {\contentslabel{2.5em}}{}{\titlerule*[0.5pc]{.}\contentspage}
\hypersetup{
  colorlinks=false,
  pdfborder={0 0 1},
  linkbordercolor={1 0 0},
  citebordercolor={0 1 0},
  urlbordercolor={0 1 1},
  pdfauthor={Yuesheng Huang, Hoi Fan Au, Jiaqi Hou, Ruicheng Zhang, Jiarui He, Haocheng Yang, Bin Hu, Kewei Zhang, Weijian Deng, Jun Zhou, Xiu Li},
  pdfcreator={LaTeX},
  pdftitle={
    AV-FABLE: Audio-Visual Manipulation Forensics in Long-Form Videos
  }
}

\definecolor{AVFableTitleNavy}{HTML}{327FBE}
\definecolor{AVFableTitleCyan}{HTML}{7F0080}

\newcommand{\gradientavfable}{%
  {\bfseries
    \textcolor{AVFableTitleNavy!100!AVFableTitleCyan}{A}%
    \textcolor{AVFableTitleNavy!86!AVFableTitleCyan}{V}%
    \textcolor{AVFableTitleNavy!71!AVFableTitleCyan}{-}%
    \textcolor{AVFableTitleNavy!57!AVFableTitleCyan}{F}%
    \textcolor{AVFableTitleNavy!43!AVFableTitleCyan}{A}%
    \textcolor{AVFableTitleNavy!29!AVFableTitleCyan}{B}%
    \textcolor{AVFableTitleNavy!14!AVFableTitleCyan}{L}%
    \textcolor{AVFableTitleNavy!0!AVFableTitleCyan}{E}%
  }%
}

\DeclareRobustCommand{\avfabletitlebrand}{%
  \texorpdfstring{%
    \mbox{%
      \raisebox{-0.18em}{%
        \includegraphics[
          height=1.35em,
          keepaspectratio
        ]{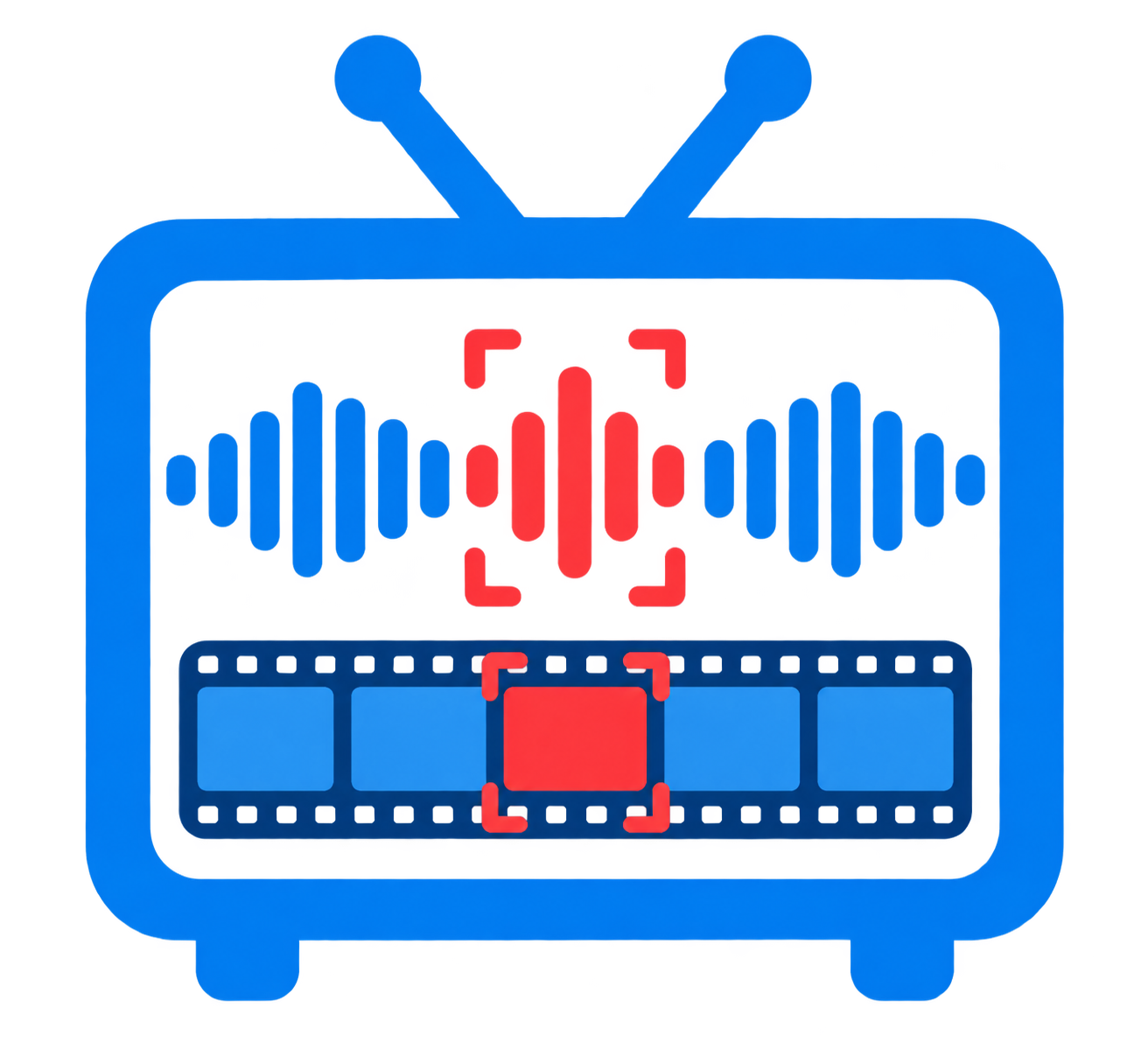}%
      }%
      \hspace{0.22em}%
      {\normalfont\gradientavfable}%
    }%
  }{AV-FABLE}%
}

\title{%
  \avfabletitlebrand:
  Audio-Visual Manipulation For\-ensics in Long-Form Videos%
}

\iflatexml
\author[1]{Yuesheng Huang\thanks{Equal contribution.}}
\author[1]{Hoi Fan Au\thanks{Equal contribution.}}
\author[1]{Jiaqi Hou}
\author[1]{Ruicheng Zhang}
\author[1]{Jiarui He}
\author[1]{Haocheng Yang}
\author[1]{Bin Hu}
\author[2]{Kewei Zhang}
\author[1]{Weijian Deng}
\author[1]{Jun Zhou\thanks{Corresponding author.}}
\author[1]{Xiu Li\thanks{Corresponding author.}}
\affil[1]{Tsinghua University}
\affil[2]{Peking University}
\else
\author{%
  Yuesheng Huang\textsuperscript{1,*}\hspace{0.6em}%
  Hoi Fan Au\textsuperscript{1,*}\hspace{0.6em}%
  Jiaqi Hou\textsuperscript{1}\hspace{0.6em}%
  Ruicheng Zhang\textsuperscript{1}\hspace{0.6em}%
  Jiarui He\textsuperscript{1}\\[0.25em]
  \bfseries Haocheng Yang\textsuperscript{1}\hspace{0.6em}%
  Bin Hu\textsuperscript{1}\hspace{0.6em}%
  Kewei Zhang\textsuperscript{2}\hspace{0.6em}%
  Weijian Deng\textsuperscript{1}\hspace{0.6em}%
  Jun Zhou\textsuperscript{1,\ensuremath{\dagger}}\hspace{0.6em}%
  Xiu Li\textsuperscript{1,\ensuremath{\dagger}}\\[0.6em]
  {\normalfont\small
    \textsuperscript{1}Tsinghua University\quad
    \textsuperscript{2}Peking University}\\[0.4em]
  {\normalfont\footnotesize
    \textsuperscript{*}Equal contribution.\quad
    \textsuperscript{\ensuremath{\dagger}}Corresponding authors.}
}
\fi

\begin{document}

\maketitle

\input{sections/abstract}
\input{sections/introduction}
\input{sections/related_work}
\input{sections/dataset}
\input{sections/method}
\input{sections/experiments}
\input{sections/conclusion}

\clearpage
\input{sections/statements}

% The appendix introduces several additional references;
% keep the bibliography compact enough to avoid a nearly empty final reference page.
\setlength{\bibsep}{2pt plus 0.5pt minus 0.5pt}

\bibliography{references}
\bibliographystyle{references}

\clearpage
\appendix
\input{appendix/appendix}

\end{document}

%% file: math_commands.tex
\usepackage{amsmath,amsfonts,bm}

\def\eqref#1{equation~\ref{#1}}
\def\1{\bm{1}}

\DeclareMathAlphabet{\mathsfit}{\encodingdefault}{\sfdefault}{m}{sl}
\SetMathAlphabet{\mathsfit}{bold}{\encodingdefault}{\sfdefault}{bx}{n}

%% file: sections/abstract.tex
\begin{abstract}
% Localized AI manipulations in long-form videos often alter brief events within otherwise authentic content, so forensic analysis must go beyond video-level classification. Existing audio-visual deepfake benchmarks focus on short talking-head clips, while long-form temporal forensics remains largely visual. We introduce AV-FABLE, a dual-timeline benchmark for sparse manipulations in open-scene long-form audio-visual videos, comprising 16,650 pristine--manipulated twin pairs (998 hours) produced with 27 editing configurations. Its separate audio and visual timelines support modality-specific temporal localization and expose failures that joint-timeline evaluation hides. On AV-FABLE, zero-shot prompting of audio-visual large language models (AV-LLMs) yields near-chance detection and poor temporal localization, whereas supervised readouts of their frozen perception features recover discriminative forensic signals. Building on this finding, we propose NFSR, a strong baseline that reads these frozen features with separate audio and visual temporal decoders. NFSR achieves the best audio-visual localization in-domain and in all three out-of-distribution (OOD) settings, one of which includes held-out videos from an external long-video forensic benchmark. The dataset, code, and model weights will be publicly released.
% \wj{
Localized AI manipulations in long-form videos often affect brief events within otherwise authentic content, making video-level classification insufficient for forensic analysis. Existing audio-visual deepfake benchmarks largely focus on short talking-head clips, while long-form temporal forensics remains predominantly visual. We introduce AV-FABLE, a dual-timeline benchmark for sparse manipulations in open-scene long-form audio-visual videos. AV-FABLE contains 16,650 pristine--manipulated twin pairs (998 hours) created with 27 editing configurations, with separate audio and visual timelines that enable modality-specific temporal localization and expose errors hidden by joint-timeline evaluation.
On AV-FABLE, zero-shot prompting of audio-visual large language models (AV-LLMs) yields near-chance detection and poor temporal localization. In contrast, supervised readouts of their frozen perception features recover strong forensic signals, suggesting manipulation evidence is encoded in these representations but not effectively exposed through language outputs. Motivated by this observation, we propose NFSR~(Native Forensics via Separated Readouts), which reads frozen AV-LLM features with separate audio and visual temporal decoders. NFSR achieves the best audio-visual localization performance in-domain and across all three out-of-distribution settings, including transfer to held-out videos from an external long-video forensic benchmark. AV-FABLE, code, and model weights will be publicly released.
% }
\end{abstract}

%% file: sections/introduction.tex
\section{Introduction}

\label{sec:introduction}

A few manipulated seconds can change the meaning of an otherwise authentic video. Replacing an utterance can alter what a speaker appears to have said; inserting a sound can suggest an event that never occurred; modifying a brief action can change the interpretation of an entire sequence. Such edits may affect the audio, the visual stream, or both, often with different temporal boundaries. A video-level authenticity label therefore reveals little about where a manipulation occurs or which modality is affected. Fine-grained forensic analysis requires identifying \textbf{when a manipulation occurred and which modality was affected}.

Existing benchmarks capture parts of this problem, but not this setting as a whole. LAV-DF and the AV-Deepfake1M series advance audio-visual deepfake detection and temporal localization in predominantly speaker-centered videos~\citep{cai2022lavdf,cai2024avdeepfake1m,cai2025avdeepfake1mpp}. TASLE and ActivityForensics extend visual forensics to localized manipulations in broader, untrimmed videos~\citep{feng2026tasle,bao2026activityforensics}, while AVID studies audio-visual inconsistency understanding and temporal grounding in long-form videos~\citep{chen2026avid}. What remains missing is a benchmark that localizes manipulations independently on the audio and visual timelines of open-scene long-form video. This distinction is important because audio and visual edits may overlap, occur asynchronously, or affect only one modality. We target this setting: sparse manipulations of speech, environmental sounds, actions, objects, and attributes embedded within largely authentic long-form content.

\begin{figure}[t]
\centering
\includegraphics[width=\linewidth]{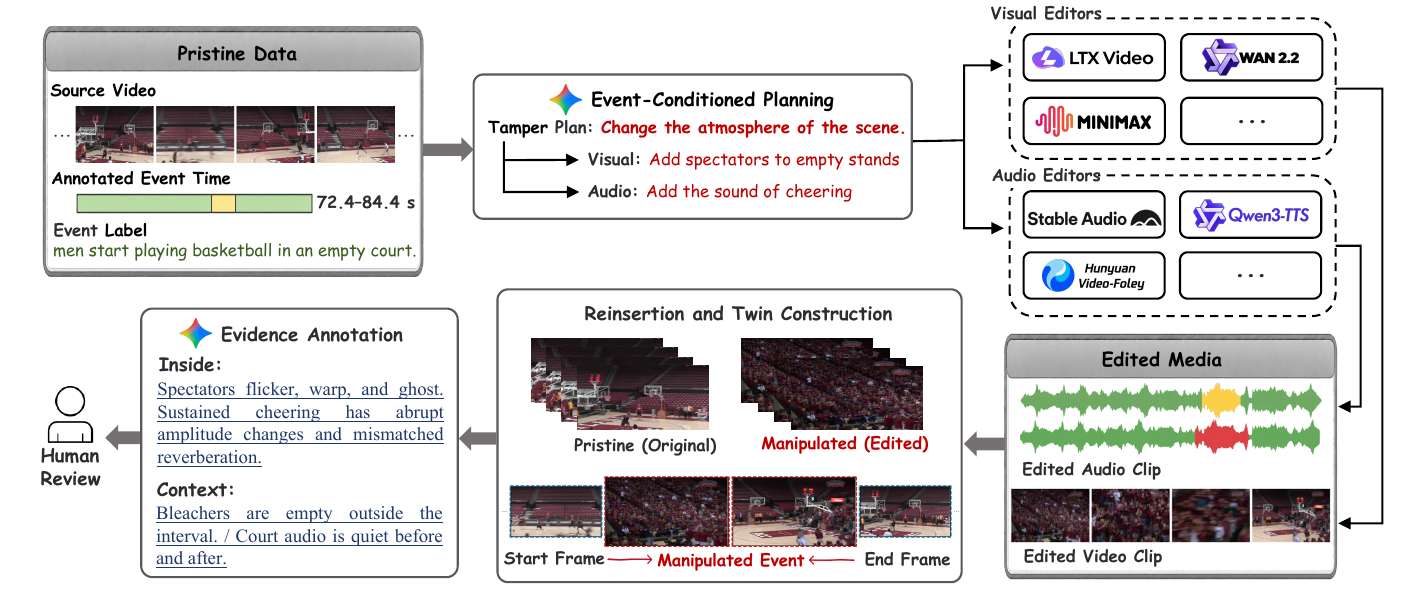}
\caption{\textbf{Overview of SEE-Forge, the event-driven pipeline for constructing AV-FABLE.}
% \wj{
Given an event in a pristine video, SEE-Forge plans audio and/or visual manipulations, executes them with editors, and reinserts them into context to create pristine--manipulated twin pairs. The executed interval provides temporal ground truth, while human-reviewed annotations record forensic cues within and around the manipulated event.}
% }
\label{fig:see_forge}
\vspace{-3mm}
\end{figure}

% The distinction between modalities is central to this setting. Audio and visual forgeries may overlap in time yet require different forensic judgments, or occur asynchronously; on a single joint timeline, a prediction can receive temporal credit without identifying the manipulated modality, and cross-modal plausibility does not establish the authenticity of either track. Evaluation should therefore preserve both temporal boundaries and modality identity.

The distinction between modalities is central to this setting. Audio and visual manipulations may overlap in time, occur asynchronously, or affect only one modality. A joint timeline can therefore reward a prediction for finding the right moment even when it identifies the wrong stream. Cross-modal consistency is also insufficient: two streams can remain plausible even when one has been altered. Evaluation should therefore preserve both temporal boundaries and modality identity.

We introduce \textbf{AV-FABLE} (Forensic Audio-visual Benchmark for Long-form Events), a dual-timeline benchmark for sparse manipulations in open-scene long-form audio-visual videos. AV-FABLE comprises \textbf{16,650 pristine--manipulated twin pairs spanning 998 hours and 27 editing configurations}; videos are minute-scale (median 88\,s), and manipulations cover a median of 9.45\% of each manipulated video. Our construction pipeline, \textbf{SEE-Forge}, selects natural events in the source footage, plans a local audio or visual edit for each, routes the edit to a suitable generator, and reinserts the result into the original video (Figure~\ref{fig:see_forge}). Each manipulated video has a pristine twin from the same source under a matched export process, reducing content and processing confounds. Separate audio and visual timelines record the interval and modality of every executed edit, and human-verified evidence records describe how each edit can be observed. The benchmark centers on temporal localization, together with detection, modality attribution, and explanation, and evaluates transfer to held-out sources, generators, and editing routes with thresholds fixed on source-domain data. Native-track evaluation exposes failures that joint-timeline evaluation hides: the visual predictions of our model alone reach 64.04 Joint-mAP, above every visual-only baseline, while missing every audio event.

When prompted zero-shot on AV-FABLE, the audio-visual large language models (AV-LLMs) Qwen2.5-Omni and Qwen3-Omni~\citep{xu2025qwen25omni,xu2025qwen3omni} reach only 51--52 detection AUROC and at most 0.30 mAP on either track. Yet linear probes on their frozen perception features separate manipulated from authentic segments with 90.8 (audio) and 80.8 (visual) AUROC. This striking gap suggests that AV-LLMs encode useful forensic evidence in their perceptual representations, but fail to reliably expose it through zero-shot language outputs. This observation motivates learning temporal readouts directly from these frozen features.

% Building on this finding, we propose \textbf{NFSR} (Native Forensics via Separated Readouts), a strong baseline for AV-FABLE. NFSR reads intermediate-layer and interface features from the frozen audio and visual encoders, conditions them on statistics of the host video, and localizes audio and visual manipulations with separate temporal decoders; the predictions on the two tracks also yield video-level detection and modality attribution. In-domain, NFSR reaches 91.30 AV-mAP and 96.04 detection AUROC, compared with 84.76 AV-mAP for UMMAFormer on the same frozen features. It also achieves the highest AV-mAP in all three OOD settings, including one that contains a source-disjoint subset of the external TASLE benchmark.
Building on this finding, we propose \textbf{NFSR} (Native Forensics via Separated Readouts), which directly reads frozen audio and visual representations for modality-specific temporal localization. NFSR combines intermediate-layer and interface features, conditions local evidence on the host video, and uses separate temporal decoders to localize audio and visual manipulations. Their predictions further support video-level detection and modality attribution. In-domain, NFSR achieves 91.30 AV-mAP and 96.04 detection AUROC, compared with 84.76 AV-mAP for UMMAFormer using the same frozen features. NFSR further achieves the highest AV-mAP across all three OOD settings, including transfer to a source-disjoint subset of the external TASLE benchmark. Our contributions are summarized as follows:

\begin{itemize}

\item \textbf{AV-FABLE and the dual-timeline task:} a large-scale benchmark for sparse audio-visual temporal localization in long-form videos, with pristine--manipulated twin pairs, separate audio and visual annotations, and three OOD settings. Its native-track evaluation exposes failures hidden by joint-timeline evaluation.

\item \textbf{A finding on AV-LLMs:} zero-shot prompting remains near chance on AV-FABLE, while supervised readouts of frozen perception features recover strong forensic signals, revealing a gap between internal representations and language outputs.

\item \textbf{NFSR:} a readout of frozen AV-LLM encoders that combines intermediate-layer and interface features, host conditioning, and separate audio and visual decoders. NFSR achieves the highest AV-mAP in-domain and across all three OOD settings.

\end{itemize}

% We will publicly release the dataset, code, and model weights to support further research.

%% file: sections/related_work.tex
\section{Related Work}
\label{sec:related_work}

\textbf{Benchmarks for manipulation detection and temporal localization.}
Facial-forgery benchmarks primarily assess image- or video-level authenticity~\citep{rossler2019faceforensics,celebdf,dfdc}, with FakeAVCeleb adding manipulated speech to talking-head video~\citep{khalid2021fakeavceleb}.
ForgeryNet and PartialSpoof provide interval-level supervision for visual and audio manipulations, respectively~\citep{forgerynet,zhang2023partialspoof}.
LAV-DF and the AV-Deepfake1M series extend temporal localization to content-driven audio-visual edits, predominantly in speaker-centered clips~\citep{cai2022lavdf,cai2023glitch,cai2024avdeepfake1m,cai2025avdeepfake1mpp}.
Beyond this setting, AVFakeBench broadens scene coverage~\citep{xia2026avfakebench}, while Omni-Fake brings several media types into a common forensic benchmark~\citep{li2026omnifake}.
Longer authentic context is central to TASLE and ActivityForensics, which embed sparse visual manipulations within untrimmed or activity-rich videos~\citep{feng2026tasle,bao2026activityforensics}.
AVID studies long-form audio-visual inconsistency grounding~\citep{chen2026avid}, but cross-modal compatibility and the authenticity of each stream remain distinct targets.
AV-FABLE combines open-scene long-form videos with separate temporal annotations and native-track evaluation, requiring both manipulation boundaries and the affected modality to be identified.

\textbf{Temporal localization and forensic representations.}
Beyond classification~\citep{chollet2017xception,jung2022aasist,chen2024demamba}, temporal forensics studies boundary prediction in audio~\citep{wu2024cfprf,zhong2024boundaryaware} and audio-visual content~\citep{cai2023glitch,chen2025hbmnet,shen2026evas}.
Audio-visual localizers exploit cross-modal cues~\citep{koutlis2025auvire,anshul2025intra}, adaptive features in UMMAFormer~\citep{zhang2023ummaformer}, or query-based and contextual modeling~\citep{zhu2025regqav,yin2025unicaclf}.
Action detection supplies interval predictors~\citep{zhang2022actionformer}, while TASLE's MSLoc uses language-assisted proposal refinement~\citep{feng2026tasle}.
Modality-decoupled detectors retain stream-specific evidence~\citep{peng2026davdet,li2026divideconquer}.
AV-LLMs provide language-based predictions~\citep{xu2025qwen25omni,xu2025qwen3omni,tang2025video}, whereas supervised probes assess forensic information accessible from pretrained features~\citep{alain2016understanding,boldisor2026selfsupervised}.
NFSR conditions frozen perception features on the input host video and learns separate audio and visual temporal readouts.
Appendix~\ref{sec:related_work_extended} expands these comparisons and reviews generation and editing techniques.

%% file: sections/dataset.tex
\section{AV-FABLE}
\label{sec:avfable}

Table~\ref{tab:dataset_comparison} compares AV-FABLE with representative
forensic benchmarks: among them, only AV-FABLE combines open-scene long-form
videos, separate audio and visual timelines, pristine twins, and
event-grounded explanations. It comprises \textbf{16,650 twin pairs}
(33,300 videos; 998.1 hours), with 10,306 audio and 11,847 visual events
across 27 editing configurations.

\subsection{Benchmark Design and Task}
\label{sec:avfable_task}

For a video $X=(A,V)$ of duration $T$, each modality $m\in\{a,v\}$ carries
its own event set $\mathcal G_m=\{(b_i^m,e_i^m,o_i^m,k_i^m)\}_{i=1}^{N_m}$,
where $b_i^m,e_i^m$ are temporal boundaries, $o_i^m$ is the manipulation
operation, and $k_i^m$ is the target category. Audio and visual events may
overlap, occur asynchronously, or appear on only one track. Given $X$, a
system returns confidence-scored intervals on each track (localization),
a video-level authenticity score (detection), one of four attribution
classes---clean, audio-only, visual-only, or both-track---and an evidence
statement for each predicted event (explanation).

\emph{Native-track evaluation} preserves modality identity: a visual
prediction cannot match an audio event, even at identical timestamps. It is
the primary localization protocol of AV-FABLE, summarized by AV-mAP, the
mean of A-mAP and V-mAP. \emph{Joint-timeline evaluation} instead merges
overlapping ground-truth intervals across tracks and matches predictions
regardless of their track (Joint-mAP); we report it as an auxiliary view.
Comparing the two separates locating a manipulated interval from
identifying the track it affects. Predictions on pristine videos count as
false positives under both protocols, as do predictions on an unmodified
track under native-track evaluation.
\vspace{-2mm}
\begin{table*}[t]
\centering
\caption{\textbf{Forensic benchmarks comparison.}
Length: short clips (Clip) or minute-scale videos (Long).
Manip.: audio (A) or visual (V) manipulations;
Loc.: temporal localization; Expl.: explanations; Pair:
pristine--manipulated twins; OOD shift: held-out sources, generators, or
manipulation types/routes. \textcolor{orange!85!black}{(\ensuremath{\checkmark})}:
partial support, or shift
without a dedicated held-out split.}

\label{tab:dataset_comparison}
\fontsize{6}{7}\selectfont
\setlength{\tabcolsep}{1.4pt}
\renewcommand{\arraystretch}{1.12}
\providecommand{\AFcmark}{\textcolor{green!45!black}{\ensuremath{\checkmark}}}
\providecommand{\AFxmark}{\textcolor{red!75!black}{\ensuremath{\times}}}
\providecommand{\AFqmark}{\textcolor{orange!85!black}{(\ensuremath{\checkmark})}}
\providecommand{\AFnr}{\AFxmark}
% Keep each benchmark name and its citation on one visual line.
\let\newline\nobreakspace
\newlength{\AFTableWidth}
\setlength{\AFTableWidth}{\dimexpr\textwidth-22\tabcolsep\relax}
\begin{tabular}{@{}
>{\raggedright\arraybackslash}m{0.270\AFTableWidth}
>{\centering\arraybackslash}m{0.175\AFTableWidth}
>{\raggedright\arraybackslash}m{0.135\AFTableWidth}
>{\centering\arraybackslash}m{0.060\AFTableWidth}
>{\centering\arraybackslash}m{0.032\AFTableWidth}
>{\centering\arraybackslash}m{0.032\AFTableWidth}
>{\centering\arraybackslash}m{0.045\AFTableWidth}
>{\centering\arraybackslash}m{0.075\AFTableWidth}
>{\centering\arraybackslash}m{0.045\AFTableWidth}
>{\centering\arraybackslash}m{0.040\AFTableWidth}
>{\centering\arraybackslash}m{0.040\AFTableWidth}
>{\centering\arraybackslash}m{0.040\AFTableWidth}
@{}}
\toprule
\multirow{2}{*}{\textbf{Dataset}} & \multirow{2}{*}{\textbf{Scale}} & \multirow{2}{*}{\textbf{Scene}} & \multirow{2}{*}{\textbf{Length}} & \multicolumn{2}{c}{\textbf{Manip.}} & \multirow{2}{*}{\textbf{Loc.}} & \multirow{2}{*}{\textbf{Expl.}} & \multirow{2}{*}{\textbf{Pair}} & \multicolumn{3}{c}{\textbf{OOD shift}} \\
\cmidrule(lr){5-6}\cmidrule(l){10-12}
 & & & & A & V & & & & Src & Gen & \shortstack{Type} \\
\midrule
FaceForensics++\newline\citep{rossler2019faceforensics} & \shortstack[r]{6,000 v} & Faces & Clip & \AFxmark & \AFcmark & \AFxmark & \AFnr & \AFqmark & \AFnr & \AFqmark & \AFqmark \\
Celeb-DF (v2)\newline\citep{celebdf} & \shortstack[r]{6,529 v} & Celebrity faces & Clip & \AFxmark & \AFcmark & \AFxmark & \AFnr & \AFqmark & \AFnr & \AFnr & \AFnr \\
DFDC\newline\citep{dfdc} & \shortstack[r]{128,154 v} & Consented actors & Clip & \AFqmark & \AFcmark & \AFxmark & \AFnr & \AFqmark & \AFqmark & \AFqmark & \AFnr \\
FakeAVCeleb\newline\citep{khalid2021fakeavceleb} & \shortstack[r]{20,000 v} & Talking heads & Clip & \AFcmark & \AFcmark & \AFxmark & \AFnr & \AFqmark & \AFnr & \AFnr & \AFnr \\
\midrule
ForgeryNet\newline\citep{forgerynet} & \shortstack[r]{221,247 v} & Faces & Clip & \AFxmark & \AFcmark & \AFcmark & \AFnr & \AFqmark & \AFnr & \AFnr & \AFnr \\
LAV-DF\newline\citep{cai2022lavdf,cai2023glitch} & \shortstack[r]{136,304 v} & Talking heads & Clip & \AFcmark & \AFcmark & \AFcmark & \AFnr & \AFqmark & \AFnr & \AFnr & \AFnr \\
AV-Deepfake1M++\newline\citep{cai2025avdeepfake1mpp} & \shortstack[r]{2,051,154 v} & Talking heads & Clip & \AFcmark & \AFcmark & \AFcmark & \AFnr & \AFqmark & \AFqmark & \AFqmark & \AFnr \\
\midrule
TASLE\newline\citep{feng2026tasle} & \shortstack[r]{12,472 v} & Open scenes & Long & \AFxmark & \AFcmark & \AFcmark & \AFcmark & \AFqmark & \AFqmark & \AFcmark & \AFnr \\
ActivityForensics\newline\citep{bao2026activityforensics} & \shortstack[r]{5,389 v} & Open activities & Long & \AFxmark & \AFcmark & \AFcmark & \AFnr & \AFqmark & \AFnr & \AFcmark & \AFqmark \\
\midrule
\rowcolor{green!6}
\textbf{AV-FABLE (ours)} & \shortstack[r]{\textbf{33,300 v}} & \textbf{Open scenes} & \textbf{Long} & \AFcmark & \AFcmark & \AFcmark & \AFcmark & \AFcmark & \AFcmark & \AFcmark & \AFcmark \\
\bottomrule

\end{tabular}
\vspace{-3mm}
\end{table*}

\subsection{Event-Driven Construction and Annotation}
\label{sec:avfable_construction}

\textbf{From natural events to executable edits.}
\textbf{SEE-Forge} (Figure~\ref{fig:see_forge}) edits events that already
occur in the host video, i.e., the source video that receives the edit,
rather than arbitrary time windows.
ActivityNet~\citep{caba2015activitynet} and
EPIC-KITCHENS-100~\citep{damen2022epic} provide third-person activities and
egocentric interactions; HDTF~\citep{zhang2021hdtf} supplies held-out
talking-head videos. For each host video, we build an event catalog from
captions, speech transcripts, acoustic onsets, motion, and object
interactions. Gemini 3.7 Flash~\citep{google2026gemini37flash} then acts as
the planner: it selects one cataloged event and specifies the target track,
interval, semantic change, and content to preserve. A second model verifies
that the target is observable and that a compatible generator can execute
the instruction.

\textbf{Editing and pristine twins.}
Each of the 27 editing configurations pairs a generator with an execution
route: frame-conditioned generation, masked or unmasked video-to-video
editing, joint audio-visual generation, audio generation, or speech
replacement (Table~\ref{tab:avc_editors}). SAM~2 masks constrain localized
visual edits~\citep{ravi2024sam2}. Generators receive the target interval
with 2--5\,s of surrounding context, and their outputs are reinserted at the
recorded coordinates with checks on motion continuity, visual anchors, and
loudness at the seams.
Each manipulated video has a pristine twin from the same host video, and
both twins pass through the same cropping, reinsertion, and export steps
(resolution, frame rate, codec, and audio format). These controls address
the content and processing shortcuts reported for earlier
datasets~\citep{smeu2025shortcuts}: metadata-only classifiers reach at most
54.6 AUROC (Table~\ref{tab:avc_metadata}). Traces that generators leave
inside the edited interval, such as local resampling, generator texture, or
synthesis residue, are beyond the reach of file-level probes and are
discussed in Appendix~\ref{app:avc_shortcuts}. The pristine twin also
serves as a content-matched reference for annotation and error analysis.

\textbf{Temporal labels and observable evidence.}
\label{sec:avfable_annotation}
Temporal labels are execution-grounded: each event records the track,
interval, operation, and target category of an executed edit, and
``ground truth'' refers to these annotations. Intervals mark where edits
were applied rather than perceptual boundaries, so a perceptible change may
cover only part of an interval or begin and end slightly off its
boundaries; reviewers check boundaries during retention but do not re-draw
them. Operations cover replacement,
insertion, regeneration, and removal; targets include speech, sounds,
actions, objects, and scene content. For 20,529 events, Gemini 3.7 Flash
drafts a candidate evidence record from a short clip around the event.
Evidence records distinguish observations
\emph{inside} the event, \emph{context} from surrounding or paired media,
and the \emph{judgment} these observations support, so the intended edit
and its observable trace are recorded separately.
Automatic checks screen decoding, synchronization, and integration failures.
Two reviewers then independently inspect all 22,153 events against the
pristine twin and the edit instruction. They decide whether the edit was
realized within its recorded interval without collateral changes, abrupt
boundary transitions, or visual breakdown (Cohen's
$\kappa=0.84$)~\citep{cohen1960coefficient} and whether the
media support its evidence statements ($\kappa=0.82$); a third reviewer
adjudicates the 7.7\% of events with disagreements
(Appendix~\ref{app:human_qc_agreement}). Events without supported evidence
remain available for detection and localization but not explanation.
\vspace{-2mm}
\subsection{Dataset Characteristics and Evaluation Splits}
\label{sec:avfable_protocol}

The dataset contains 4,903 audio-only, 6,376 visual-only, and 5,371 both-track
manipulated videos. Across manipulated videos, the union of audio and visual
event intervals covers a median of 9.45\% of the video, while 91.5\% have
at most 30\% manipulated time. Videos are minute-scale: the median lasts
88\,s and 80\% last between 33 and 207\,s. The median event lasts about 9\,s
in the training and ID splits, so each manipulation occupies a small part of
its timeline. In 561 both-track
videos, the audio and visual events do not overlap, so each track requires
its own boundaries.

We allocate 10,954/2,347/2,347/1,002 twin pairs to training, validation,
in-domain (ID) testing, and OOD testing; all derivatives of a host
video remain in the same split. Of the 27 editing configurations, 23
appear in training and four are held out. We evaluate three OOD settings
(Section~\ref{sec:nfsr_exp_ood}): \emph{source transfer} to HDTF and a
source-disjoint TASLE subset, \emph{held-out generators} on seen routes,
and \emph{held-out TI2V}, a text-and-image-conditioned video generation
route withheld together with its generator.
Evaluation covers localization, detection, attribution, and explanation,
with decision thresholds fixed on source-validation data
(Section~\ref{sec:nfsr_experiments}). Explanation is the most challenging
dimension: current baselines produce well-formed outputs but often
misidentify the operation or target (Appendix~\ref{app:avc_explanation}),
leaving substantial room for future work.
Appendices~\ref{app:dataset_details} and~\ref{app:implementation_details}
detail the composition, construction settings, and split protocols.
\vspace{-2mm}

%% file: sections/method.tex
\section{NFSR: Native Forensics via Separated Readouts}
\label{sec:method}

\begin{figure}[t]
    \centering
    \includegraphics[width=\linewidth]{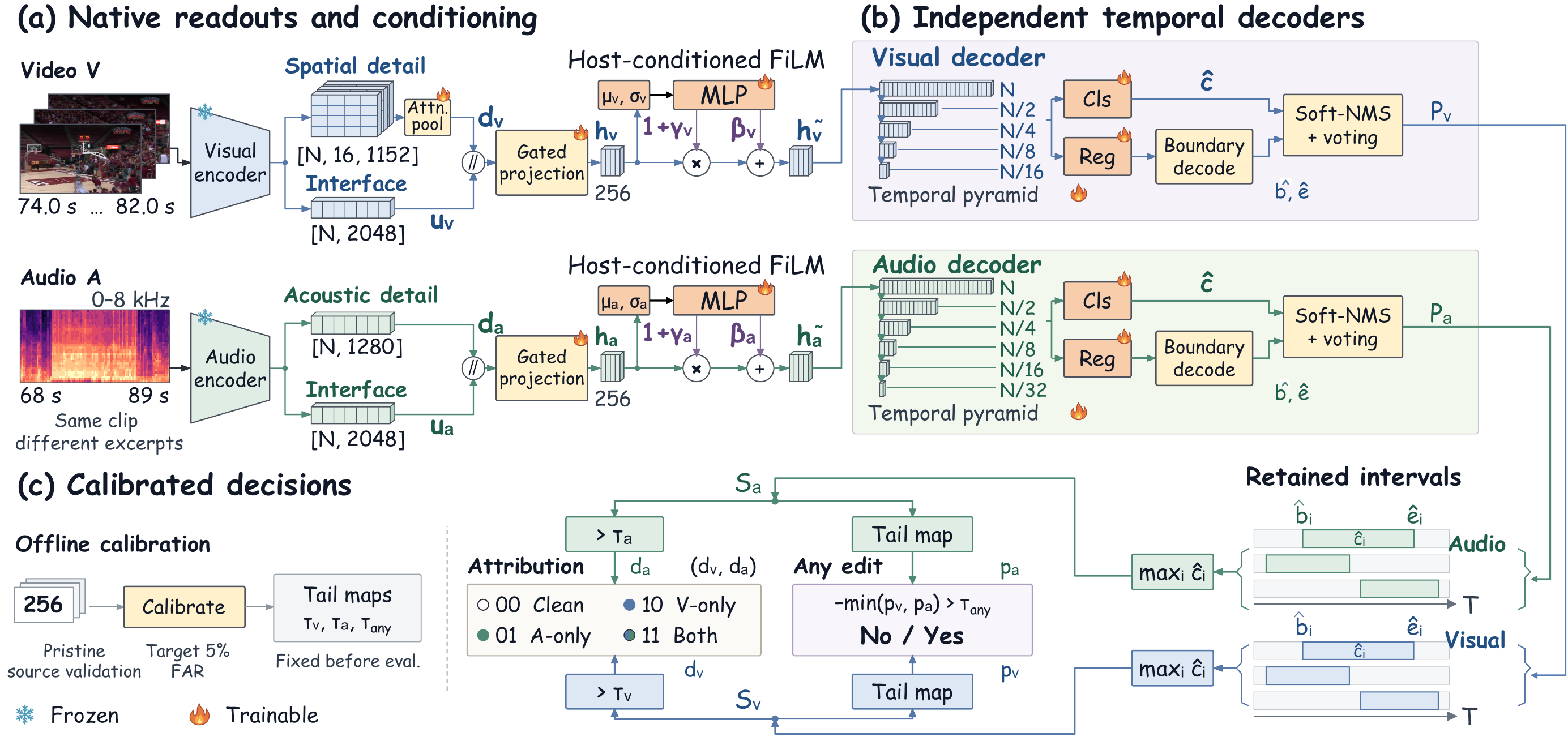}
    \caption{\textbf{Overview of NFSR.}
(a) Frozen intermediate-layer (``detail'') and interface features pass
through modality-specific gated projection and host-conditioned FiLM.
(b) Independent decoders localize audio and visual manipulations on
separate timelines.
(c) Source-calibrated thresholds give video-level detection and four-way
attribution without changing the intervals.
Audio and video excerpts come from the same clip; intervals are schematic.}
\label{fig:nfsr}
\vspace{-2mm}
\end{figure}

NFSR reads frozen perception features and localizes manipulations on
separate audio and visual timelines. Each of its three components addresses
one observation: supervised readouts recover discriminative signals from
perception features below the language interface
(Section~\ref{sec:native_readout}) whether a local change is anomalous
depends on its host video (Section~\ref{sec:host_conditioning}); and the two
modalities need not share manipulation boundaries
(Section~\ref{sec:separated_localization}).
Figure~\ref{fig:nfsr} summarizes the framework.
\vspace{-2mm}

\subsection{Reading Frozen Perception Features}
\label{sec:native_readout}

Zero-shot Qwen-Omni models remain near chance for detection through the
language interface, whereas linear probes on their frozen perception
features separate manipulated from authentic segments
(Tables~\ref{tab:nfsr_id} and~\ref{tab:avc_probes}). NFSR therefore reads
the perception features directly. We instantiate it with the frozen audio
and visual encoders of Qwen3-Omni~\citep{xu2025qwen3omni}. For each modality
$m\in\{a,v\}$, we extract \emph{intermediate-layer features} from a fixed
intermediate layer and \emph{interface features} from the encoder output
passed to the language model. The former retain fine perceptual detail,
while the latter provide a complementary representation aligned with the
language interface.

For a video of duration $T$, both feature types are aligned to
$N=\lceil T/\Delta\rceil$ positions with $\Delta=0.25\,\mathrm{s}$.
Visual spatial tokens are attention-pooled at each position; audio features
are temporal vectors. Denoting the intermediate-layer and interface features
by $\mathbf d_t^m$ and $\mathbf u_t^m$, a modality-specific gated projection
forms $\mathbf h_t^m=\phi_m(\mathbf d_t^m,\mathbf u_t^m)\in\mathbb R^D$,
with $D=256$. Reading both layers lets the temporal model combine their
information rather than committing to a single representation depth.

\subsection{Host-Conditioned Feature Modulation}
\label{sec:host_conditioning}

A local sound, motion, or texture change takes on different significance
across video conditions and scene content. With sparse manipulation,
the extended host video provides context for interpreting such changes.
We therefore condition local features on the input video's channel-wise
mean $\boldsymbol\mu_m$ and standard deviation $\boldsymbol\sigma_m$,
computed over valid temporal positions. Following feature-wise linear
modulation~\citep{perez2018film}, we apply
\begin{equation}
    (\boldsymbol\gamma_m,\boldsymbol\beta_m)
    =g_m([\boldsymbol\mu_m;\boldsymbol\sigma_m]),\qquad
    \widetilde{\mathbf h}_t^m
    =\mathbf h_t^m\odot(\mathbf 1+\boldsymbol\gamma_m)+\boldsymbol\beta_m.
    \label{eq:host_film}
\end{equation}
Here, $g_m$ is a two-layer MLP, $[\cdot;\cdot]$ denotes concatenation,
and $\odot$ is element-wise multiplication. The scale and shift vectors
$\boldsymbol\gamma_m,\boldsymbol\beta_m\in\mathbb R^D$ are shared across
positions within a video but vary with its context. The MLP thus learns
how host statistics should reweight local evidence while keeping the
features' absolute scale, which per-video standardization would discard.
The statistics come from the input video alone, in training and at
inference.

\subsection{Modality-Separated Temporal Localization}
\label{sec:separated_localization}

An event in one modality does not establish manipulation in the other.
We use independently parameterized decoders $D_a$ and $D_v$, each receiving
only its own conditioned feature sequence. This keeps the evidence path
modality-specific while allowing different event boundaries on the two
tracks. Following multi-scale temporal localization
\citep{zhang2022actionformer,yang2024dyfadet}, each decoder constructs a
temporal pyramid with masked dilated convolutions and predicts a
classification logit and nonnegative left/right distances at each point:
\begin{equation}
    (\hat b_j^m,\hat e_j^m,\hat c_j^m)
    =\left(
        u_j-r_j\hat d_j^{m,-},\;
        u_j+r_j\hat d_j^{m,+},\;
        \operatorname{sigmoid}(z_j^m)
      \right).
    \label{eq:boundary_decoding}
\end{equation}
Here, $j$ indexes points across pyramid levels, $u_j$ is the point's
timestamp, and $r_j$ is its level's temporal stride in seconds.
The decoder outputs $z_j^m$ and distances $\hat d_j^{m,-},\hat d_j^{m,+}$
in stride units. Thus, the first two terms recover event boundaries in
video time, and the third gives classification confidence. Temporal
Soft-NMS~\citep{bodla2017soft} combines proposals across scales; surviving
intervals are clipped to video bounds to form $\widehat{\mathcal S}_m$.

\textbf{Training objective.}
Each branch is supervised by its own temporal annotations:
\begin{equation}
    \mathcal L_{\mathrm{readout}}
    =\sum_{m\in\{a,v\}}
      \left(\mathcal L_{\mathrm{focal}}^m
      +\lambda_{\mathrm{reg}}\mathcal L_{\mathrm{DIoU}}^m\right).
    \label{eq:readout_loss}
\end{equation}
Here, $\mathcal L_{\mathrm{focal}}^m$ is sigmoid focal
loss~\citep{lin2017focal} over valid temporal points, and
$\mathcal L_{\mathrm{DIoU}}^m$ is one-dimensional distance-IoU
loss~\citep{zheng2020diou} on points assigned to ground-truth events;
$\lambda_{\mathrm{reg}}=1$ balances the terms. Pristine videos and
unmodified tracks supply negative supervision. Assignment rules and
optimization settings appear in Appendix~\ref{app:implementation_details}.

\textbf{Detection and attribution.}
\label{sec:detection_attribution}
Each track's video score $S_m$ is its highest proposal confidence. We
convert $S_m$ to an upper-tail probability $p_m$ relative to 256 pristine
source-validation videos and score any-manipulation detection as
$S_{\mathrm{any}}=-\min\{p_a,p_v\}$. Detection and each track decision use their own
threshold, calibrated once at a 5\% false-alarm budget and reused on every
test split. The two track decisions give the attribution class (clean,
audio-only, visual-only, or both-track) without altering either event set;
Appendix~\ref{app:implementation_details} gives the full scoring rules.

%% file: sections/experiments.tex
% Experimental results; extended analyses appear in appendix/appendix.tex.
\section{Experiments}
\label{sec:nfsr_experiments}

\begin{table}[t]
\centering
\caption{\textbf{ID main results on AV-FABLE (\%).}
Blocks are visual-only, audio-only, and audio-visual. Outputs are separate
audio/visual intervals (A/V), one track (A or V), fixed-window intervals
(win), modality-labeled language timestamps (Text), or modality-unlabeled
intervals (Unified). -- marks unsupported metrics; absent tracks count as
misses in AV-mAP and Joint-timeline metrics. NFSR, UMMAFormer, and
TASLE-trained ActionFormer are three-seed means (Table~\ref{tab:avc_seeds});
others are single runs. \textbf{Bold} and \underline{underlined} values mark
the best and second-best results per column; FAR is an operating-point
diagnostic and is not ranked.}
\vspace{2pt}
\label{tab:nfsr_id}
\begingroup
\scriptsize
\setlength{\tabcolsep}{1.8pt}
\renewcommand{\arraystretch}{1.05}
\resizebox{\linewidth}{!}{%
\begin{tabular}{@{}llc*{10}{r}@{}}
\toprule
Method & Training Form & Output & AUROC & FAR & Attr. F1
& AV-mAP & A-mAP & V-mAP & Joint-mAP
& A-IoU & V-IoU & Joint-IoU\\
\midrule
% HTML_DATA_BEGIN main
Xception & Official weights & V$_{\mathrm{win}}$ & 50.71 & 4.94 & 17.62 & 0.01 & -- & 0.02 & 0.02 & -- & 0.22 & 0.24 \\
MSLoc-PG & Official weights & V & 57.34 & 3.54 & 20.37 & 0.95 & -- & 1.90 & 1.36 & -- & 4.55 & 3.40 \\
ActionFormer & TASLE & V & 67.94 & 2.78 & 20.99 & 5.06 & -- & 10.13 & 7.56 & -- & 5.56 & 4.55 \\
Xception & AV-FABLE & V$_{\mathrm{win}}$ & 66.27 & 4.94 & 23.27 & 0.04 & -- & 0.08 & 0.06 & -- & 4.87 & 3.98 \\
DeMamba (Stage-1 window) & AV-FABLE & V$_{\mathrm{win}}$ & 74.35 & 4.26 & 28.77 & 24.75 & -- & 49.49 & 26.83 & -- & 35.21 & 28.54 \\
ActionFormer & AV-FABLE & V & 84.55 & 5.84 & 34.95 & 44.61 & -- & \underline{89.22} & \underline{59.93} & -- & \textbf{80.66} & 62.71 \\
\midrule
AASIST & Official weights & A$_{\mathrm{win}}$ & 55.17 & 5.07 & 18.65 & 0.00 & 0.01 & -- & 0.04 & 0.51 & -- & 0.38 \\
AASIST & AV-FABLE & A$_{\mathrm{win}}$ & 75.95 & 5.62 & 29.42 & 0.03 & 0.06 & -- & 0.07 & 16.26 & -- & 12.06 \\
\midrule
Qwen2.5-Omni-7B (ZS) & Official weights & Text & 51.04 & 0.04 & 10.92 & 0.28 & 0.25 & 0.30 & 0.26 & 0.00 & 0.00 & 0.00 \\
Qwen3-Omni-30B-A3B (ZS) & Official weights & Text & 52.28 & 4.13 & 20.42 & 0.03 & 0.02 & 0.04 & 0.03 & 0.56 & 0.47 & 0.60 \\
video-SALMONN 2+ 7B (ZS) & Official weights & Text & 49.98 & 0.43 & 0.59 & 0.00 & 0.00 & 0.00 & 0.00 & 0.01 & 0.02 & 0.02 \\
DAV-Det & Official weights & A/V$_{\mathrm{win}}$ & 56.52 & 4.90 & 22.30 & 0.40 & 0.59 & 0.21 & 0.57 & 2.90 & 0.86 & 2.68 \\
BA-TFD+ & AV-FABLE & A/V & 66.71 & 4.52 & 27.07 & 0.88 & 1.54 & 0.22 & 0.63 & 37.54 & 1.23 & 23.58 \\
UMMAFormer + Qwen3 & AV-FABLE & A/V & \underline{89.18} & 5.75 & \underline{74.83} & \underline{84.76} & \underline{89.90} & 79.62 & 58.42 & \underline{79.77} & 65.25 & \underline{67.44} \\
Qwen3 direct timestamp SFT & AV-FABLE & Text & 76.53 & 4.35 & 47.66 & 37.80 & 57.30 & 18.30 & 19.20 & 56.06 & 26.86 & 33.99 \\
AuViRe & AV-FABLE & Unified & 64.87 & 6.31 & -- & -- & -- & -- & 0.42 & -- & -- & 11.21 \\
ICS-Sync & AV-FABLE & Unified & 59.61 & 6.43 & -- & -- & -- & -- & 5.47 & -- & -- & 8.39 \\
DAV-Det + AV-FABLE & AV-FABLE & A/V$_{\mathrm{win}}$ & 80.56 & 4.05 & 48.02 & 1.50 & 0.80 & 2.21 & 1.51 & 10.49 & 16.85 & 14.79 \\
\rowcolor{gray!10}
NFSR + Qwen3 (Ours) & AV-FABLE & A/V & \textbf{96.04} & 3.52 & \textbf{83.61} & \textbf{91.30} & \textbf{92.24} & \textbf{90.36} & \textbf{72.33} & \textbf{84.72} & \underline{76.90} & \textbf{81.21} \\
% HTML_DATA_END main
\bottomrule
\end{tabular}%
}
\endgroup
% \vspace{-3mm}
\end{table}

We evaluate whether frozen perception features support sparse
audio-visual temporal localization, how the resulting readouts transfer to
the three OOD settings, and how feature depth, host conditioning, and
modality separation affect these results.

\textbf{Evaluation protocol.}
We use AV-FABLE's source-disjoint splits. Localization uses
confidence-ranked mAP@[0.30:0.05:0.70] and Event-IoU, the mean best
overlap per event, with zero assigned to misses. AV-mAP,
headline measure, averages A-mAP and V-mAP over tracks with
events; a method without output on a track receives zero AP there.
Joint-timeline metrics (Joint-mAP and Joint-IoU;
Section~\ref{sec:avfable_task}) apply to outputs without track labels,
but count a correct co-timed prediction on the second track as a
duplicate, so we report them as an auxiliary view
(Appendix~\ref{app:avc_timeline}). Detection uses AUROC and
realized false-alarm rate (FAR). Attribution reports macro-F1 over the
four classes (clean, audio-only, visual-only, both-track) present in
each evaluation set's ground truth (Eq.~\ref{eq:attr_f1}); held-out TI2V,
for example, contains only clean and visual-only videos, so attribution
is compared within rather than across settings. Thresholds are set on 256
pristine source-validation videos at a nominal 5\% FAR budget and held
fixed on ID and OOD data (Appendix~\ref{app:implementation_details}).

\textbf{Baselines and implementation.}
We compare three prediction approaches: language-interface inference
with Qwen2.5-Omni-7B, Qwen3-Omni-30B-A3B, and video-SALMONN~2+~7B
\citep{xu2025qwen25omni,xu2025qwen3omni,tang2025video};
single-modality detection with Xception~\citep{chollet2017xception},
AASIST~\citep{jung2022aasist}, and DeMamba~\citep{chen2024demamba};
and temporal or audio-visual modeling with BA-TFD+~\citep{cai2023glitch},
ActionFormer~\citep{zhang2022actionformer},
UMMAFormer~\citep{zhang2023ummaformer},
AuViRe~\citep{koutlis2025auvire},
ICS-Sync~\citep{anshul2025intra},
and DAV-Det~\citep{peng2026davdet}.
Qwen3-Omni timestamp SFT additionally tests supervised language-based
localization. We include the released MSLoc proposal generator
\citep{feng2026tasle}; its full pipeline and baseline adaptation details
appear in Appendix~\ref{app:additional_experiments} and
Appendix~\ref{app:implementation_details}, respectively.
These groups differ in role and output form (Table~\ref{tab:nfsr_id}):
AV-LLMs test whether prompted outputs expose manipulations,
single-modality detectors are track-specific references, and localizers
with separate audio and visual interval outputs are the direct comparison.
For fixed-window,
single-track, unified, and text outputs, localization scores also reflect
the fit of the output form to the task; AV-FABLE-trained AASIST, for
example, reaches 75.95 AUROC but 0.06 A-mAP after window-to-interval
conversion.
NFSR trains 256-dimensional readouts on frozen Qwen3-Omni features
at 0.25\,s resolution for 12 epochs using AdamW
($10^{-4}$ learning rate)~\citep{loshchilov2017decoupled};
UMMAFormer shares its features and training budget.
All architecture choices, layer choices, hyperparameters, and model
selection were determined without access to OOD test results; checkpoints
are selected by source-validation mAP.

% \FloatBarrier
\subsection{Forensic Readout and In-Domain Results}
\label{sec:nfsr_exp_id}

\textbf{Supervised readouts recover signals that prompted outputs miss.}
Zero-shot AV-LLM predictions yield near-chance detection and at most 0.30
mAP on either track (Table~\ref{tab:nfsr_id}), partly because of their
text-timestamp output. Timestamp SFT keeps this output form and raises
A-mAP to 57.30 but V-mAP only to 18.30. On the same frozen features,
linear probes outperform unsupervised feature-deviation and temporal-jump
scores on both tracks (Table~\ref{tab:avc_probes}), and NFSR reaches 92.24
A-mAP and 90.36 V-mAP. The forensic evidence is therefore accessible to
supervised readouts even when prompted outputs do not reveal it.

\textbf{NFSR improves localization on shared frozen features.}
With the same Qwen3 features and training budget, NFSR raises AV-mAP over
UMMAFormer from 84.76 to 91.30, the highest in Table~\ref{tab:nfsr_id},
with the largest gain on the visual track (+10.74 V-mAP). It also improves
detection AUROC by 6.86 points and Event-IoU on both tracks, so the gain
covers event coverage as well as ranking. NFSR also achieves the highest
attribution F1.

\textbf{Native-track evaluation exposes missing modalities.}
The AV-FABLE-trained ActionFormer is competitive on the visual track
(89.22 V-mAP and higher V-IoU than NFSR); its lower AV-mAP reflects its
visual-only output rather than weaker visual localization. Conversely,
NFSR's visual predictions alone reach 64.04 Joint-mAP, above every
visual-only baseline, despite zero A-mAP (Table~\ref{tab:avc_timeline});
only native-track evaluation exposes this missing modality.

% \FloatBarrier
\subsection{Transfer Across OOD Settings}
\label{sec:nfsr_exp_ood}

We evaluate the three OOD settings of Section~\ref{sec:avfable_protocol}
(composition in Table~\ref{tab:avc_ood}) without target-domain adaptation
or threshold adjustment. Source transfer changes both domain and modality
composition, and held-out TI2V changes both route and generator; each
result therefore measures transfer to the setting as a whole.

\begin{table*}[t]
\centering
\caption{\textbf{OOD performance (\%).}
Metrics, source-calibrated thresholds, rows, --, and bold/underline follow
Table~\ref{tab:nfsr_id}. For source transfer, detection averages HDTF/TASLE,
attribution and V-mAP pool them, and A-mAP uses HDTF (384 audio events).
Held-out TI2V has visual edits only (AV-mAP${}={}$V-mAP). Attribution F1
uses three, four, and two classes, respectively. $^{\dagger}$TASLE overlaps
the model's training distribution. Joint-timeline metrics and Event-IoU:
Table~\ref{tab:avc_ood_joint}.}
\vspace{2pt}
\label{tab:nfsr_ood}
\begingroup
\scriptsize
\setlength{\tabcolsep}{1.2pt}
\renewcommand{\arraystretch}{1.04}
\resizebox{\linewidth}{!}{%
\begin{tabular}{@{}ll*{16}{r}@{}}
\toprule
\multirow{2}{*}{Method} & \multirow{2}{*}{Training Form}
& \multicolumn{6}{c}{Source transfer}
& \multicolumn{6}{c}{Held-out generators}
& \multicolumn{4}{c}{Held-out TI2V}\\
\cmidrule(lr){3-8}\cmidrule(lr){9-14}\cmidrule(l){15-18}
& & AUROC & FAR & Attr.\ F1 & AV-mAP & A-mAP & V-mAP
& AUROC & FAR & Attr.\ F1 & AV-mAP & A-mAP & V-mAP
& AUROC & FAR & Attr.\ F1 & AV-mAP\\
\midrule
% HTML_DATA_BEGIN ood-main
Xception & Official weights & 48.41 & 3.53 & 22.27 & 0.30 & -- & 0.59 & 50.59 & 5.47 & 18.96 & 0.11 & -- & 0.23 & 50.18 & 5.33 & 37.21 & 0.25 \\
MSLoc-PG & Official weights & 72.37 & 0.97 & 47.97 & 26.51 & -- & \underline{53.01}$^{\dagger}$ & 57.86 & 3.22 & 22.82 & 3.82 & -- & 7.65 & 56.88 & 5.00 & 40.18 & 0.92 \\
ActionFormer & TASLE & 73.82 & 0.00 & 47.55 & 37.50 & -- & \textbf{75.00}$^{\dagger}$ & \underline{77.50} & 2.10 & 24.05 & 10.52 & -- & 21.05 & 66.78 & 2.89 & 44.00 & 15.15 \\
Xception & AV-FABLE & 52.63 & 1.60 & 24.24 & 0.15 & -- & 0.31 & 53.26 & 3.86 & 19.44 & 0.25 & -- & 0.49 & 53.91 & 4.33 & 38.46 & 0.67 \\
DeMamba (Stage-1 window) & AV-FABLE & 54.94 & 2.23 & 25.65 & 0.12 & -- & 0.23 & 55.75 & 1.93 & 18.16 & 1.18 & -- & 2.37 & 59.66 & 4.33 & 39.11 & 2.17 \\
ActionFormer & AV-FABLE & 62.31 & 3.74 & 40.01 & 19.13 & -- & 38.26 & 69.10 & 4.50 & 34.26 & 25.60 & -- & \textbf{51.21} & \textbf{85.46} & 6.67 & \textbf{72.03} & \underline{61.57} \\
\midrule
AASIST & Official weights & 52.43 & 0.00 & 22.24 & 0.00 & 0.00 & -- & 52.08 & 5.14 & 19.20 & 0.04 & 0.07 & -- & 50.00 & 4.67 & 32.80 & 0.00 \\
AASIST & AV-FABLE & 71.61 & 1.04 & 52.46 & 0.03 & 0.05 & -- & 61.70 & 3.22 & 26.29 & 0.27 & 0.55 & -- & 50.00 & 6.67 & 32.56 & 0.00 \\
\midrule
Qwen2.5-Omni-7B (ZS) & Official weights & 48.92 & 0.00 & 10.96 & 0.92 & 0.34 & 1.50 & 49.11 & 0.00 & 10.27 & 0.62 & 0.56 & 0.68 & 51.30 & 0.00 & 18.45 & 0.86 \\
Qwen3-Omni-30B-A3B (ZS) & Official weights & 51.98 & 7.70 & 35.41 & 1.14 & 0.01 & 2.26 & 50.98 & 4.50 & 19.99 & 0.04 & 0.05 & 0.04 & 51.31 & 6.00 & 32.56 & 0.12 \\
video-SALMONN 2+ 7B (ZS) & Official weights & 50.02 & 0.00 & 22.24 & 0.00 & 0.00 & 0.00 & 50.00 & 0.32 & 18.50 & 0.00 & 0.00 & 0.00 & 49.83 & 0.33 & 33.30 & 0.00 \\
DAV-Det & Official weights & 64.78 & 24.03 & 45.68 & 10.47 & 0.10 & 20.84 & 53.23 & 4.50 & 20.50 & 0.81 & 0.56 & 1.07 & 51.24 & 3.67 & 38.41 & 1.27 \\
BA-TFD+ & AV-FABLE & 52.43 & 2.21 & 23.99 & 0.41 & 0.65 & 0.17 & 57.23 & 3.22 & 21.88 & 0.59 & 1.09 & 0.09 & 50.00 & 6.00 & 34.84 & 0.24 \\
UMMAFormer + Qwen3 & AV-FABLE & \underline{78.81} & 12.21 & \underline{71.78} & \underline{60.48} & \textbf{98.91} & 22.05 & 68.71 & 4.18 & \underline{35.23} & \underline{33.98} & \underline{40.98} & 26.97 & 63.46 & 5.00 & 48.01 & 36.06 \\
Qwen3 direct timestamp SFT & AV-FABLE & 76.40 & 0.66 & 47.16 & 23.94 & 47.78 & 0.09 & 59.52 & 6.11 & 31.16 & 10.80 & 21.30 & 0.30 & 51.63 & 4.33 & 34.61 & 0.26 \\
AuViRe & AV-FABLE & 51.99 & 1.04 & -- & -- & -- & -- & 54.82 & 6.75 & -- & -- & -- & -- & 50.36 & 8.67 & -- & -- \\
ICS-Sync & AV-FABLE & 50.56 & 1.30 & -- & -- & -- & -- & 54.10 & 6.75 & -- & -- & -- & -- & 49.43 & 7.00 & -- & -- \\
DAV-Det + AV-FABLE & AV-FABLE & 61.52 & 40.45 & 40.79 & 8.60 & 0.14 & 17.05 & 60.06 & 5.47 & 25.49 & 3.36 & 1.15 & 5.56 & 63.32 & 4.00 & 43.75 & 10.66 \\
\rowcolor{gray!10}
NFSR + Qwen3 (Ours) & AV-FABLE & \textbf{89.86} & 2.97 & \textbf{77.21} & \textbf{70.15} & \underline{98.37} & 41.92 & \textbf{79.40} & 3.86 & \textbf{49.06} & \textbf{49.12} & \textbf{48.95} & \underline{49.29} & \underline{84.56} & 4.67 & \underline{68.06} & \textbf{68.07} \\
% HTML_DATA_END ood-main
\bottomrule
\end{tabular}%
}
\endgroup
% \vspace{-3mm}
\end{table*}

\textbf{Localization gains hold in OOD settings.}
NFSR achieves the highest AV-mAP in every setting
(Table~\ref{tab:nfsr_ood}) and exceeds the matched-feature UMMAFormer, mainly through higher V-mAP; the two are on par on HDTF
audio. NFSR also has higher attribution F1 and Event-IoU than UMMAFormer
in all three settings (Table~\ref{tab:avc_ood_joint}). Visual-only
baselines remain competitive on their own track: the AV-FABLE-trained
ActionFormer slightly exceeds NFSR in V-mAP on held-out generators and
leads in AUROC, attribution F1, and V-IoU on held-out TI2V, and the
TASLE-trained models lead source-transfer V-mAP because those events come
almost entirely from their training distribution. Without audio output,
however, their AV-mAP is at most 37.50 in the two settings with audio edits.

\textbf{False alarms are read together with coverage.}
Low FAR alone does not indicate reliable decisions, since a conservative
detector can miss manipulations; we therefore read FAR together with
attribution F1 and Event-IoU, which give no credit for missed events at the
same operating point. Without recalibration, NFSR's realized FAR stays
within 2.97--4.67\% on all evaluation sets, close to the 5\% budget. On
source transfer, UMMAFormer's 12.21\% FAR comes with lower coverage than
NFSR on both tracks (Table~\ref{tab:avc_ood_joint}). Visual coverage under
shift remains the main open problem: NFSR reaches 26.42--41.21 V-IoU, and
the two DAV-Det rows cover more visual events on source transfer but at
24.03\% and 40.45\% FAR. Corruption can likewise lower recall and FAR
together (Appendix~\ref{app:avc_robustness}).

\subsection{Ablation Studies}
\label{sec:nfsr_exp_ablation}

We remove one component at a time under the same training and
model-selection protocol (Table~\ref{tab:nfsr_ablation}).

\begin{table}[t]
\centering
\caption{\textbf{NFSR core component ablation.}
Three-seed mean $\pm$ sample standard deviation with matched 12-epoch
training, splits, scoring, and thresholds. OOD-All averages
Table~\ref{tab:nfsr_ood}; A-mAP uses its two audio-edit settings
(Appendix~\ref{app:implementation_details}). Bold/underline mark the best
and second-best means (FAR not ranked).
Joint-timeline metrics and OOD-All Event-IoU: Table~\ref{tab:avc_ablation_aux}.}
\vspace{2pt}
\label{tab:nfsr_ablation}
\begingroup
\scriptsize
\setlength{\tabcolsep}{2.0pt}
\renewcommand{\arraystretch}{1.12}
\resizebox{\linewidth}{!}{%
\begin{tabular}{@{}l*{14}{r}@{}}
\toprule
\multirow{2}{*}{Variant}
& \multicolumn{8}{c}{ID}
& \multicolumn{6}{c}{OOD-All}\\
\cmidrule(lr){2-9}\cmidrule(l){10-15}
& AUROC & FAR & Attr.\ F1 & AV-mAP & A-mAP & V-mAP & A-IoU & V-IoU
& AUROC & FAR & Attr.\ F1 & AV-mAP & A-mAP & V-mAP\\
\midrule
% HTML_DATA_BEGIN ablation

w/o intermediate-layer features
& \shortstack{92.73\\$\pm$ 0.17}
& \shortstack{4.93\\$\pm$ 0.69}
& \shortstack{75.91\\$\pm$ 0.52}
& \shortstack{84.75\\$\pm$ 0.13}
& \shortstack{88.94\\$\pm$ 0.12}
& \shortstack{80.56\\$\pm$ 0.20}
& \shortstack{80.05\\$\pm$ 0.71}
& \shortstack{65.24\\$\pm$ 1.21}
& \shortstack{79.43\\$\pm$ 0.25}
& \shortstack{5.29\\$\pm$ 0.66}
& \shortstack{60.92\\$\pm$ 0.69}
& \shortstack{53.73\\$\pm$ 0.47}
& \shortstack{68.53\\$\pm$ 0.67}
& \shortstack{42.66\\$\pm$ 0.65} \\
\hline
w/o interface features
& \shortstack{\underline{95.58}\\$\pm$ 0.16}
& \shortstack{6.42\\$\pm$ 0.80}
& \shortstack{\underline{83.35}\\$\pm$ 0.51}
& \shortstack{\underline{90.61}\\$\pm$ 0.10}
& \shortstack{\underline{92.12}\\$\pm$ 0.24}
& \shortstack{\underline{89.10}\\$\pm$ 0.12}
& \shortstack{\textbf{84.81}\\$\pm$ 0.30}
& \shortstack{\textbf{79.76}\\$\pm$ 1.51}
& \shortstack{81.83\\$\pm$ 0.24}
& \shortstack{8.84\\$\pm$ 0.16}
& \shortstack{\textbf{64.82}\\$\pm$ 1.74}
& \shortstack{\underline{58.38}\\$\pm$ 0.40}
& \shortstack{\underline{72.95}\\$\pm$ 0.32}
& \shortstack{47.31\\$\pm$ 0.76} \\
\hline
w/o host conditioning
& \shortstack{95.41\\$\pm$ 0.08}
& \shortstack{5.00\\$\pm$ 0.93}
& \shortstack{82.25\\$\pm$ 0.48}
& \shortstack{89.86\\$\pm$ 0.57}
& \shortstack{91.36\\$\pm$ 0.86}
& \shortstack{88.36\\$\pm$ 0.32}
& \shortstack{83.77\\$\pm$ 0.33}
& \shortstack{74.13\\$\pm$ 0.78}
& \shortstack{\underline{81.90}\\$\pm$ 0.82}
& \shortstack{4.45\\$\pm$ 1.15}
& \shortstack{62.24\\$\pm$ 1.05}
& \shortstack{57.72\\$\pm$ 1.55}
& \shortstack{69.24\\$\pm$ 2.93}
& \shortstack{\underline{47.89}\\$\pm$ 0.83} \\
\hline
\rowcolor{gray!10}
Full NFSR
& \shortstack{\textbf{96.04}\\$\pm$ 0.10}
& \shortstack{3.52\\$\pm$ 0.33}
& \shortstack{\textbf{83.61}\\$\pm$ 0.04}
& \shortstack{\textbf{91.30}\\$\pm$ 0.13}
& \shortstack{\textbf{92.24}\\$\pm$ 0.16}
& \shortstack{\textbf{90.36}\\$\pm$ 0.28}
& \shortstack{\underline{84.72}\\$\pm$ 0.55}
& \shortstack{\underline{76.90}\\$\pm$ 0.78}
& \shortstack{\textbf{84.61}\\$\pm$ 0.12}
& \shortstack{3.83\\$\pm$ 0.44}
& \shortstack{\underline{64.77}\\$\pm$ 0.72}
& \shortstack{\textbf{62.45}\\$\pm$ 0.08}
& \shortstack{\textbf{73.66}\\$\pm$ 0.79}
& \shortstack{\textbf{53.10}\\$\pm$ 0.56} \\

% HTML_DATA_END ablation
\bottomrule
\end{tabular}%
}
\endgroup
% \vspace{-3mm}
\end{table}

\textbf{Each component matters more under distribution shift.}
Removing any component lowers OOD-All AV-mAP, with larger drops than in-domain. Intermediate-layer features contribute most: removing them reduces AV-mAP by 6.55 points on ID and 8.72 on OOD-All. Removing interface features has little effect on ID AV-mAP but reduces OOD-All AV-mAP by 4.07 points and more than doubles OOD FAR. Removing host conditioning lowers OOD-All AV-mAP by 4.73 points. The ablated variants also achieve higher Joint-mAP on ID but lower under shift (Table~\ref{tab:avc_ablation_aux}). Appendix~\ref{app:avc_residual} further shows that both interface-predictable and residual intermediate-layer features remain discriminative.

\textbf{Separated readouts preserve modality-specific decisions.}
On audio-only speech replacements (Appendix~\ref{app:avc_cross_track}),
the separated visual readout fires on 2.9\% of both edited videos and
pristine twins, whereas the visual head of a cross-modal readout, whose
branches exchange information through cross-modal attention, fires on
4.6\% of pristine twins but 11.8--15.1\% of the edited videos.

\FloatBarrier

%% file: sections/conclusion.tex
% \section{Conclusion}
% \label{sec:conclusion}

% We presented AV-FABLE, a dual-timeline benchmark for sparse temporal manipulation localization in open-scene long-form audio-visual videos. Native-track evaluation reveals modality-specific failures that joint-timeline evaluation hides. AV-LLMs prompted zero-shot stay near chance on AV-FABLE, yet supervised readouts of their frozen representations recover discriminative forensic signals; NFSR builds on this finding and achieves the highest AV-mAP in-domain and in all three OOD settings. \emph{Limitations.} AV-FABLE is synthetically constructed, its execution-grounded temporal labels may deviate slightly from perceptible manipulation boundaries, and its release depends on the availability of third-party source videos. Robustness to media corruption and grounded explanation remain open challenges for current models, including NFSR. We will publicly release the dataset, code, and model weights to support further research.

\section{Conclusion}
\label{sec:conclusion}
% \wj{
We introduce AV-FABLE, a dual-timeline benchmark for localizing sparse audio and visual manipulations in open-scene long-form videos. By preserving both temporal boundaries and modality identity, AV-FABLE exposes failures that joint-timeline evaluation can miss. Our experiments reveal a clear gap in current AV-LLMs: zero-shot prompting remains near chance, while supervised readouts of frozen perceptual representations recover strong forensic signals. Building on this observation, NFSR achieves the highest AV-mAP in-domain and across all three OOD settings, showing the effectiveness of modality-separated temporal readouts for long-form audio-visual forensics.
% }

% \wj{
\textit{Limitation and future direction.} AV-FABLE also opens several directions for future research. Extending the benchmark to broader real-world editing processes, refining perceptual boundary annotations, improving robustness to media corruption, and developing more faithful evidence-grounded explanations are promising next steps. We will publicly release the dataset, code, and model weights to support further work in long-form audio-visual forensics.
% }

%% file: sections/statements.tex
\subsection*{AI use statement}

Generative AI models were used in the SEE-Forge pipeline to construct AV-FABLE, including event-level edit planning, localization of source events, edit-feasibility checks, audio and visual media generation and editing, and drafting candidate evidence annotations (Section~\ref{sec:avfable_construction} and Appendix~\ref{app:dataset_details}). Temporal manipulation labels were derived from the executed edits. Two reviewers independently assessed edit validity and evidence support using the edited media, pristine counterparts, and edit instructions, with disagreements adjudicated by a third reviewer. Events without supported evidence records were retained for detection and localization but not explanation. AI tools also provided language assistance during manuscript preparation. The authors take responsibility for the final manuscript, results, and accompanying artifacts.

\subsection*{Ethics statement}
This work aims to advance research on audio-visual manipulation detection, localization, and forensic analysis in long-form videos. AV-FABLE builds on existing research datasets and introduces controlled manipulations using generative models. We follow the access and usage terms of the corresponding source datasets and do not redistribute source media, or excerpts thereof, when such redistribution is restricted by third-party licenses. The public release will instead focus on event-level annotations, source identifiers and timestamps, provenance metadata, independently releasable generated assets, and reconstruction utilities. Beyond long-form localization, these generated segments, produced by recent video, audio, and joint audio-visual generators across 27 editing configurations, can also serve as a standalone resource: labeled with generator, route, and operation and paired through source timestamps with their pristine counterparts, they support real-versus-generated classification, and those with human-verified evidence records additionally support evidence-grounded explanation. Given the dual-use nature of deepfake research, we do not provide end-to-end pipelines intended for impersonation, fraud, or other malicious uses, and we avoid intentionally constructing defamatory or deceptive content targeting real individuals. Human review is used solely for data and annotation quality control. Additional details on data licensing, human review, and release policies are provided in the Appendix.

\subsection*{Reproducibility statement}
The reproducibility materials comprise the complete AV-FABLE dataset manifest; the split, OOD-setting, and threshold-calibration lists; an independent evaluator with unit tests; and the model code with the training configuration. AV-FABLE, the code, and model weights will be publicly released in accordance with the source-data terms described in the Ethics statement. Additional implementation and dataset construction details are provided in the Appendix.

%% file: appendix/appendix.tex
% Appendix manuscript fragment. The project preamble supplies amsmath,
% amssymb, booktabs, graphicx, array, tabularx, colortbl, placeins,
% tcolorbox, hyperref, and natbib.
%
% Visual language used only in this appendix. The restrained blue-gray
% palette improves navigation throughout the appendix.
\definecolor{AVCBlue}{HTML}{315B7D}
\definecolor{AVCBlueLight}{HTML}{EEF4F8}
\definecolor{AVCBluePale}{HTML}{F7FAFC}
\definecolor{AVCLine}{HTML}{AABCC9}
\colorlet{AVCTableHead}{AVCBlueLight}
\newcolumntype{L}[1]{>{\raggedright\arraybackslash}p{#1}}
\newcommand{\AVCtablefont}{%
 \footnotesize
 \setlength{\tabcolsep}{4pt}%
 \renewcommand{\arraystretch}{1.12}}
\newcommand{\AVCcompacttablefont}{%
 \scriptsize
 \setlength{\tabcolsep}{3pt}%
 \renewcommand{\arraystretch}{1.02}}

\setlength{\textfloatsep}{12pt plus 2pt minus 3pt}
\setlength{\floatsep}{10pt plus 2pt minus 2pt}
\setlength{\intextsep}{10pt plus 2pt minus 2pt}
\setlength{\abovecaptionskip}{5pt}
\setlength{\belowcaptionskip}{1pt}

% Align appendix page bottoms using the available display and float spacing.
\flushbottom

\newcommand{\AVCplaceholder}[2]{%
 \begin{tcolorbox}[
  enhanced,
  width=0.96\linewidth,
  colback=AVCBluePale,
  colframe=AVCLine,
  boxrule=0.5pt,
  arc=1.2mm,
  left=3mm,right=3mm,top=2.5mm,bottom=2.5mm]
  \centering\small\color{AVCBlue}\textbf{#1}
  \par\smallskip
  \raggedright\color{black!82}#2
 \end{tcolorbox}}

\providecommand{\AVCfigure}[3]{%
 \begin{figure}[H]
 \centering
 \AVCplaceholder{Figure placeholder}{#3}
 \caption{#2}
 \label{#1}
 \end{figure}}

\clearpage
\addcontentsline{toc}{section}{Appendix}
\startcontents[appendix]
\begingroup
\normalsize
\setlength{\parskip}{0pt}
\setlength{\baselineskip}{16pt}
\printcontents[appendix]{l}{1}{%
 \section*{\centering Appendix Contents}}
\endgroup

\newpage

\begin{figure}[H]
\centering
\includegraphics[width=\linewidth]{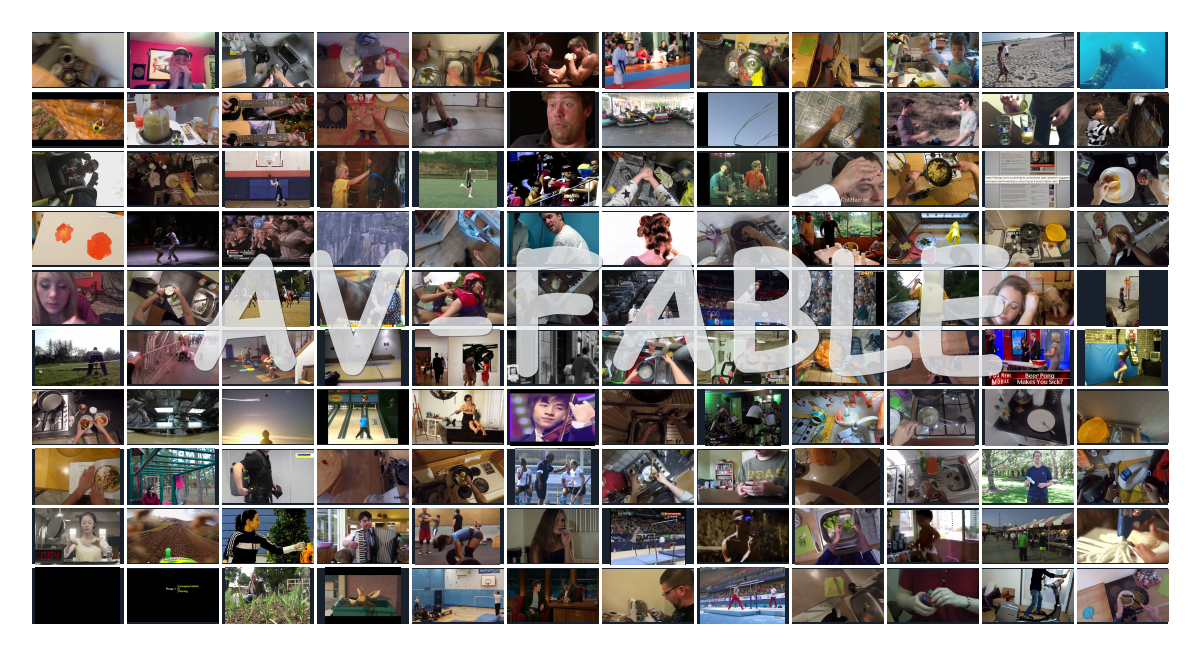}
\caption{Scene diversity in AV-FABLE. Example frames span indoor and
outdoor environments, human activities, objects, food preparation, sports,
and other contexts represented in the twin pairs.}
\label{fig:avc_teaser}
\end{figure}

\section{AV-FABLE: Construction and Data Quality}
\label{app:dataset_details}
\label{app:avc_data}

\subsection{Sources, Composition, and Labels}
ActivityNet-Raw supplies open-scene third-person videos, while
EPIC-KITCHENS-100 supplies egocentric recordings
\citep{caba2015activitynet,damen2022epic}.
The EPIC construction pool contains 400 source recordings from
37 participants, totaling 67.3 hours; natural event clips are extracted
before local manipulation and matched export.
HDTF provides held-out talking-head hosts \citep{zhang2021hdtf}.
Source qualification checks decodability, both media tracks,
synchronization, usable context, and observable candidate events.

The release comprises 16,650 twin pairs and 22,153
track-specific events (Table~\ref{tab:avc_statistics}).
The total of 998.1 hours sums both members of every pair; counting each
pair once gives 499.1 hours. Video-level duration and coverage distributions
count each pair once, whereas event statistics count the annotated intervals
on each modality separately.
The 16,650 manipulated videos contain 4,903 audio-only, 6,376
visual-only, and 5,371 both-track examples. Independent timelines allow
events to overlap or occur asynchronously; jointly generated media can
yield separate audio and visual event annotations.
Event records include identity, track, interval, operation, target type,
execution route, generator, and evidence status.
Operations comprise replacement (14,914 events), insertion (5,555),
regeneration (961), and removal (723). Targets cover sound events,
actions, objects, attributes, ambience, speech, scene regions, text,
and background.

\begin{table}[!htbp]
\centering
\AVCtablefont
\caption{Release statistics. Counts are per pair or annotated event as indicated; multi-event rate is over manipulated videos. External TASLE and supplementary TTS examples are excluded.}
\label{tab:avc_statistics}
\begingroup
\adjustbox{max width=\linewidth}{%
\begin{tabular}{@{}lrrrr@{}}
\toprule
\rowcolor{AVCTableHead}
Quantity & Train & Val. & ID test & OOD \\
\midrule
Twin pairs & 10,954 & 2,347 & 2,347 & 1,002 \\
Audio events & 6,843 & 1,451 & 1,482 & 530 \\
Visual events & 7,911 & 1,729 & 1,695 & 512 \\
Video duration: median (s) & 88 & 88 & 91 & 87 \\
Video duration: P10--P90 (s) & 36--204 & 33--205 & 36--207 & 17--238 \\
Event duration: median (s) & 9.4 & 9.3 & 9.5 & 5.7 \\
Event duration: P10--P90 (s) & 4.9--12.7 & 5.0--12.6 & 5.0--12.9 & 4.5--10.3 \\
Multi-event rate & 34.2\% & 35.0\% & 35.0\% & 2.8\% \\
Events with candidate evidence & 13,628 & 2,948 & 2,951 & 1,002 \\
\bottomrule
\end{tabular}
}
\endgroup
\end{table}

\subsection{Event-Conditioned Construction}
SEE-Forge first forms an event catalog from captions/ASR, audio onsets,
motion, shot boundaries, and interactions. A timeline localizer verifies
natural event boundaries. The planner selects an existing event and
specifies its modality, interval, single-action edit instruction,
operation, target, preservation constraints, execution route, and
expected evidence. Gemini 3.7 Flash plans~\citep{google2026gemini37flash},
Qwen2.5-VL localizes~\citep{bai2025qwen25vl}, and Qwen3-Omni reviews plan
validity and executability~\citep{xu2025qwen3omni}. After execution,
Gemini 3.7 Flash also drafts candidate evidence records from a clip around
each event. Checks establish that the target is observable, the requested
change is compatible with its context, and the selected generator can
execute the plan.

Routing matches the edit to a generator: masked V2V for local
object/attribute changes, V2V for process changes, frame-conditioned
generation for endpoint-constrained events, joint generation for
audio-visual changes, and Foley/text-to-audio/TTS for audio edits.
Generators receive the target interval and 2--5 seconds of context.
Reinsertion preserves non-target content, checks visual anchors and
motion continuity, and matches audio loudness with short cross-fades.
Persistent entity changes require consistent editing across their
visible intervals or replanning. Diversity memory controls repeated
scene, target, operation, and generator combinations.

\begin{figure}[H]
\centering
\includegraphics[width=\linewidth]{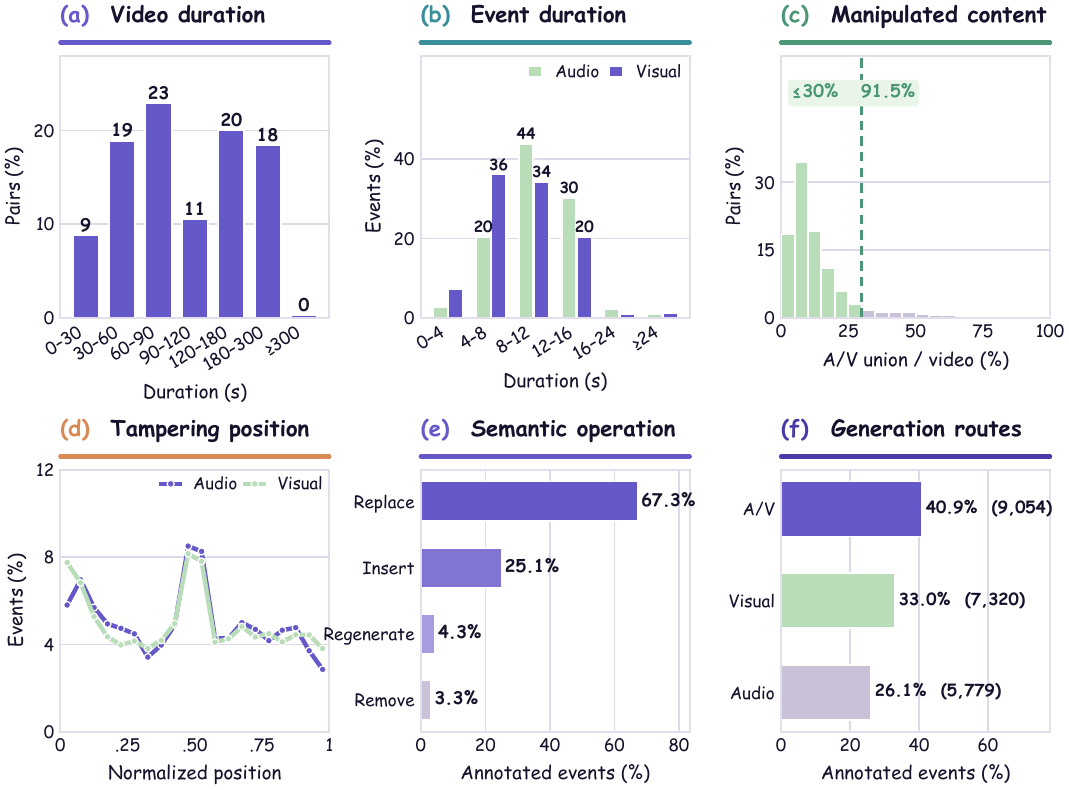}
\caption{Statistics and manipulation coverage in AV-FABLE. The six panels
show video duration, event duration by modality, manipulated temporal
coverage, tampering position, semantic operation, and aggregated generation
route. Video-level distributions count each twin pair once;
coverage is computed over the union of audio and visual intervals.}
\label{fig:avc_data}
\end{figure}

\begin{figure}[H]
\centering
\includegraphics[width=\linewidth]{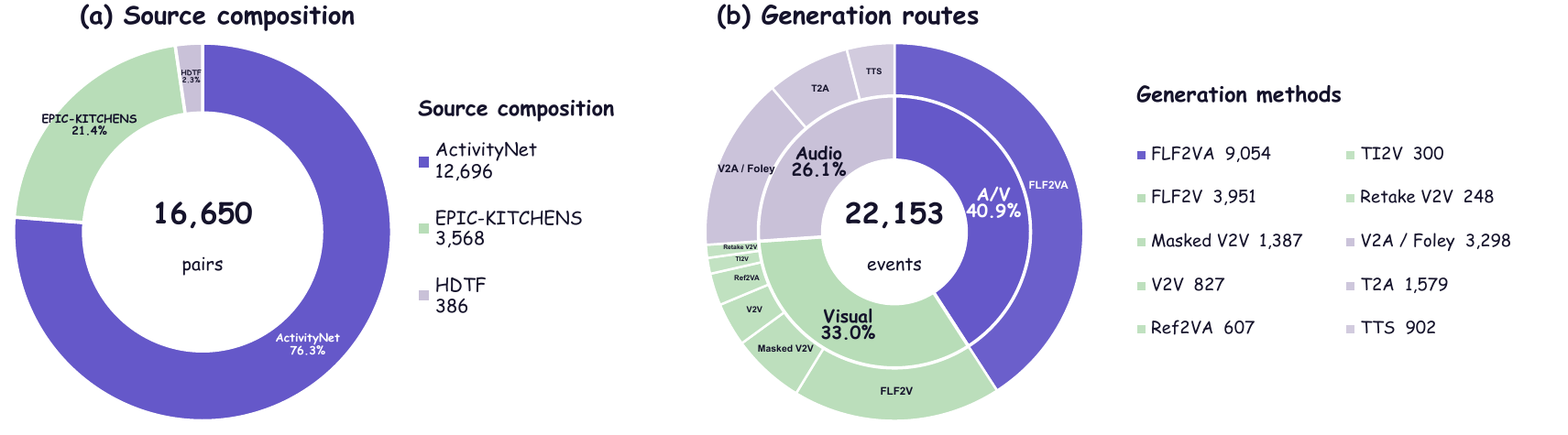}
\caption{Source and generation composition in AV-FABLE. The left donut
shows source-domain shares; the right donut shows modality attribution and
the aggregated generation methods used to produce the edited events. The
full list of editing configurations is given in Table~\ref{tab:avc_editors}.}
\label{fig:avc_coverage}
\end{figure}

Both twins then pass through a common export profile, matching
container, codec, resolution, frame rate, sample rate, channel count,
and encoding policy. Pristine twins follow the same cropping,
reinsertion, and export steps while retaining the original target content.
Figure~\ref{fig:see_forge} summarizes the pipeline;
Figures~\ref{fig:avc_data} and~\ref{fig:avc_coverage} provide the
dataset statistics and detailed construction coverage.
\begin{table}[!htbp]
\centering
\AVCtablefont
\caption{Generator coverage across the 23 training and four OOD-only editing configurations. Each configuration pairs a generator with an execution route.}
\label{tab:avc_editors}
\begingroup
\adjustbox{max width=\linewidth}{%
\begin{tabular}{@{}L{0.42\linewidth}L{0.34\linewidth}L{0.16\linewidth}@{}}
\toprule
\rowcolor{AVCTableHead}
Generator group & Routes & Role \\
\midrule
MiniMax H3~\citep{minimax2026h3}; LTX 2.3/2.5~\citep{ltxvideo} & FLF2V, FLF2VA, reference / retake & Train / ID \\
Wan Fun / VACE~\citep{wan,vace}; CogVideoX-Fun~\citep{cogvideox} & Frame-conditioned, masked V2V, V2V & Train / ID \\
MMAudio~\citep{mmaudio}; FoleyCrafter~\citep{foleycrafter}; HunyuanVideo-Foley~\citep{shan2025hunyuanvideo}; ThinkSound~\citep{liu2025thinksound} & Video-to-audio Foley & Train / ID \\
Stable Audio~\citep{evans2025stable}; TangoFlux~\citep{hung2026tangoflux}; AudioX~\citep{tian2026audiox}; Dasheng AudioGen~\citep{mei2026dasheng} & Text-to-audio & Train / ID \\
Qwen3-TTS; F5-TTS v1~\citep{qwen3tts,f5tts} & Speech replacement & Train / ID \\
Omni2Sound~\citep{dai2026omni2sound}; Wan-FLF2V~\citep{wan}; OmniWeaving~\citep{pan2026omniweaving} & Foley, FLF2V, V2V & OOD generator \\
Wan2.2-TI2V-5B~\citep{wan} & TI2V & OOD route + generator \\
\bottomrule
\end{tabular}
}
\endgroup
\end{table}

\subsection{Speech Replacement}
The speech-replacement subset contains 518 ActivityNet examples
(351/77/90 in train/validation/ID test), each combining a speech
replacement with a visual manipulation.
Eligible hosts contain 5--12 seconds of continuous English speech and
a separate same-speaker reference. Qwen3-TTS and F5-TTS produce
258 and 260 replacements, using unrelated transcripts of similar length
\citep{qwen3tts,f5tts}. Vocal separation preserves background audio;
loudness matching and 20-ms cross-fades preserve the timeline.
The median replacement lasts 7.3 seconds, and these examples carry
both-track attribution labels.

A separate audio-only supplement uses 242 pristine ID hosts with
matched interval/text budgets: 239 Qwen3-TTS, 239 F5-TTS, and
238 XTTS-v2 examples~\citep{casanova2024xtts}. The 90 ID-test speech replacements are built on
these hosts and reuse their generated speech. The language-verification
analyses (Appendix~\ref{app:avc_language}) use an earlier training release
without speech replacement.

\subsection{Human Review and Decision Agreement}
\label{app:human_qc_agreement}
Two reviewers independently assessed all 22,153 events using the edited
media, pristine twins, and edit instructions. Retention
(retain vs.\ reject/rework) required the intended edit to be realized while
preserving non-target content and avoiding unintended integration failures.
This decision also covered boundary quality: events whose perceptible change
clearly mismatched the recorded interval, or whose interval boundaries showed
abrupt transitions or visual breakdown, were rejected or reworked. Reviewers
did not re-draw boundaries, so retained intervals remain the executed edit
intervals.
Explanation eligibility required candidate evidence statements to be supported
by the media; otherwise valid events remained usable for detection/localization
only. Table~\ref{tab:qc_agreement} reports initial agreement on these two
binary decisions before adjudication, together with the overall rate of
referral to a third reviewer.

\begin{table}[t]
\centering
\caption{Initial agreement on retention and explanation eligibility,
and the overall third-reviewer adjudication rate.
Rates use the complete reviewed event pool as the denominator.}
\label{tab:qc_agreement}
\begingroup
\small
\setlength{\tabcolsep}{6pt}
\renewcommand{\arraystretch}{1.12}
\begin{tabular}{@{}lr@{}}
\toprule
Metric & Value \\
\midrule
Retention: raw agreement & 96.4\% \\
Retention: Cohen's $\kappa$ & 0.841 \\
Explanation eligibility: Cohen's $\kappa$ & 0.823 \\
Overall third-reviewer adjudication rate & 7.7\% \\
\bottomrule
\end{tabular}
\endgroup
\end{table}

\subsection{Metadata-Only Shortcut Probes}
\label{app:avc_shortcuts}
We test whether file-level properties distinguish pristine and manipulated
videos after matched export. Logistic regression and gradient-boosted
trees receive container format, codec, resolution, frame rate, timebase,
bitrate, audio channels and sample rate, frame count, duration, and encoder
tags, without access to decoded frames or waveforms. Five-fold evaluation
keeps each twin pair in the same fold.
\begin{table}[!htbp]
\centering
\AVCtablefont
\caption{Metadata-only authenticity probes: AUROC (\%). Ranges summarize
the evaluated multifield configurations. The OOD pool, used throughout the
appendix analyses, pools all 1,002 AV-FABLE OOD pairs across the source,
generator, and TI2V settings, without TASLE or equal weighting of settings.}
\label{tab:avc_metadata}
\begin{tabular}{@{}lrr@{}}
\toprule
\rowcolor{AVCTableHead}
Control & ID & OOD pool \\
\midrule
Multifield probes: AUROC range & 53.9--54.6 & 51.0--53.0 \\
Strongest single field & 51.9 & -- \\
\bottomrule
\end{tabular}
\end{table}
The multifield probes remain close to chance on both ID and the OOD pool
(Table~\ref{tab:avc_metadata}), and the strongest single field reaches
only 51.9 ID AUROC. File-level properties therefore carry little label
information after matched export. The probes do not inspect decoded
frames or waveforms, however, so they bound only this class of shortcut.

\paragraph{Controlled and remaining shortcut risks.}
By construction, each twin pair shares its host content, and both twins
undergo the same cropping, reinsertion, and export steps; source-disjoint
splits prevent shared-source leakage; and the metadata probes show that
file-level format carries little label information. These controls do not
exclude decoded-signal traces that generators may introduce inside the
edited interval beyond the intended semantic change, such as local
resampling by generators that operate at a different resolution, frame
rate, or sample rate, generator-specific texture, and acoustic residue
from synthesis. Some of these traces also accompany real manipulations
and are legitimate forensic evidence, whereas others may be specific to
the generators used here. The held-out-generator setting measures transfer
to unseen generators on seen routes but does not separate these two cases,
so pipeline-specific cues cannot be fully excluded.

\subsection{Extended Benchmark Comparison}
Table~\ref{tab:avc_benchmark_comparison} complements the main-text
comparison with a broader view of dataset scale, temporal structure,
annotation coverage, and evaluation design.

\begin{table*}[!t]
\centering
\caption{\textbf{Extended comparison with forensic datasets.} Additional
video, audio, and image benchmarks. Columns and symbols follow
Table~\ref{tab:dataset_comparison}.}
\label{tab:avc_benchmark_comparison}
\fontsize{6}{6.7}\selectfont
\setlength{\tabcolsep}{1.8pt}
\renewcommand{\arraystretch}{1.0}
\providecommand{\AFcmark}{\textcolor{green!45!black}{\ensuremath{\checkmark}}}
\providecommand{\AFxmark}{\textcolor{red!75!black}{\ensuremath{\times}}}
\providecommand{\AFqmark}{\textcolor{orange!85!black}{(\ensuremath{\checkmark})}}
\providecommand{\AFnr}{\AFxmark}
% Keep each benchmark name and its citation on one visual line.
\let\newline\nobreakspace
\ifdefined\AFTableWidth\else\newlength\AFTableWidth\fi
\setlength{\AFTableWidth}{\dimexpr\textwidth-20\tabcolsep\relax}
\begin{tabular}{@{}
>{\raggedright\arraybackslash}m{0.360\AFTableWidth}
>{\raggedleft\arraybackslash}m{0.150\AFTableWidth}
>{\raggedright\arraybackslash}m{0.140\AFTableWidth}
>{\centering\arraybackslash}m{0.032\AFTableWidth}
>{\centering\arraybackslash}m{0.032\AFTableWidth}
>{\centering\arraybackslash}m{0.045\AFTableWidth}
>{\centering\arraybackslash}m{0.080\AFTableWidth}
>{\centering\arraybackslash}m{0.040\AFTableWidth}
>{\centering\arraybackslash}m{0.040\AFTableWidth}
>{\centering\arraybackslash}m{0.040\AFTableWidth}
>{\centering\arraybackslash}m{0.040\AFTableWidth}
@{}}
\toprule
\multirow{2}{*}{\textbf{Dataset}} & \multirow{2}{*}{\textbf{Scale}} & \multirow{2}{*}{\textbf{Scene}} & \multicolumn{2}{c}{\textbf{Manip.}} & \multirow{2}{*}{\textbf{Loc.}} & \multirow{2}{*}{\textbf{Expl.}} & \multirow{2}{*}{\textbf{Pair}} & \multicolumn{3}{c}{\textbf{OOD shift}} \\
\cmidrule(lr){4-5}\cmidrule(l){9-11}
 & & & A & V & & & & Src & Gen & \shortstack{Type/\\route} \\
\midrule
\multicolumn{11}{@{}l}{\textbf{Temporal and audio--visual benchmarks}} \\[1pt]
AV-Deepfake1M\newline\citep{cai2024avdeepfake1m} & \shortstack[r]{1,146,760 v} & Talking heads & \AFcmark & \AFcmark & \AFcmark & \AFnr & \AFqmark & \AFnr & \AFnr & \AFnr \\
DDL-AV\newline\citep{miao2025ddl} & \shortstack[r]{330K v} & Face-centric AV & \AFcmark & \AFcmark & \AFcmark & \AFnr & \AFqmark & \AFqmark & \AFqmark & \AFnr \\
ArEnAV\newline\citep{kuckreja2026arenav} & \shortstack[r]{387,072 v} & Arabic/English speech & \AFcmark & \AFcmark & \AFcmark & \AFnr & \AFqmark & \AFnr & \AFnr & \AFnr \\
X-AVFake (FakeHunter)\newline\citep{chen2025fakehunter} & \shortstack[r]{$>5,700$ v} & Open AV scenes & \AFcmark & \AFcmark & \AFqmark & \AFcmark & \AFqmark & \AFnr & \AFnr & \AFnr \\
AVFakeBench\newline\citep{xia2026avfakebench} & \shortstack[r]{3,000 v\\12,000 QA} & Human + general & \AFcmark & \AFcmark & \AFxmark & \AFcmark & \AFnr & \AFnr & \AFnr & \AFnr \\
Omni-Fake\newline\citep{li2026omnifake} & \shortstack[r]{$\sim$1.35M items} & I / A / V / AV-TH & \AFcmark & \AFcmark & \AFqmark & \AFcmark & \AFnr & \AFqmark & \AFqmark & \AFqmark \\
\midrule
\multicolumn{11}{@{}l}{\textbf{Video datasets and evaluation suites}} \\[1pt]
DeeperForensics-1.0\newline\citep{deeperforensics} & \shortstack[r]{60,000 v} & Faces + perturb. & \AFxmark & \AFcmark & \AFxmark & \AFnr & \AFqmark & \AFnr & \AFnr & \AFnr \\
DF40\newline\citep{df40} & \shortstack[r]{$>0.1$M v\\$>1$M img} & Faces; 40 methods & \AFxmark & \AFcmark & \AFxmark & \AFnr & \AFqmark & \AFcmark & \AFcmark & \AFqmark \\
GenVideo (DeMamba)\newline\citep{chen2024demamba} & \shortstack[r]{2,314,182 v} & Open scenes & \AFxmark & \AFcmark & \AFxmark & \AFnr & \AFnr & \AFqmark & \AFqmark & \AFnr \\
GenVidBench\newline\citep{ni2026genvidbench} & \shortstack[r]{6,784,490 v} & Open scenes & \AFxmark & \AFcmark & \AFxmark & \AFnr & \AFqmark & \AFqmark & \AFcmark & \AFnr \\
AIGVDBench\newline\citep{ma2026aigvdbench} & \shortstack[r]{$>440$K v} & Open scenes & \AFxmark & \AFcmark & \AFxmark & \AFnr & \AFqmark & \AFnr & \AFcmark & \AFnr \\
VidGuard-R1 data\newline\citep{park2025vidguard} & \shortstack[r]{140,000 v} & Open scenes & \AFxmark & \AFcmark & \AFxmark & \AFqmark & \AFqmark & \AFqmark & \AFqmark & \AFnr \\
MintVid\newline\citep{tan2026videoveritas} & \shortstack[r]{4,992 v} & General / face / fact & \AFxmark & \AFcmark & \AFxmark & \AFnr & \AFqmark & \AFqmark & \AFqmark & \AFnr \\
MVAD / DDL-GAV (reported)\newline\citep{peng2026davdet} & \shortstack[r]{205,758 AV} & Realistic + anime & \AFcmark & \AFcmark & \AFxmark & \AFnr & \AFnr & \AFnr & \AFnr & \AFnr \\
EVID-Bench\newline\citep{yu2026evid} & \shortstack[r]{222 v} & Web misinformation & \AFxmark & \AFcmark & \AFxmark & \AFcmark & \AFqmark & \AFnr & \AFnr & \AFnr \\
DeepfakeBench (v2)\newline\citep{deepfakebench} & \shortstack[r]{9 base sets\\+ DF40} & Benchmark suite & \AFxmark & \AFcmark & \AFxmark & \AFnr & \AFqmark & \AFqmark & \AFqmark & \AFnr \\
FVBench (released subset)\newline\citep{wang2026fvbench} & \shortstack[r]{8,398 rows} & Public video subset & \AFxmark & \AFcmark & \AFxmark & \AFxmark & \AFxmark & \AFqmark & \AFcmark & \AFqmark \\
\midrule
\multicolumn{11}{@{}l}{\textbf{Audio-only forensic datasets}} \\[1pt]
PartialSpoof\newline\citep{zhang2023partialspoof} & \shortstack[r]{121,461 utt} & English speech & \AFcmark & \AFxmark & \AFcmark & \AFnr & \AFqmark & \AFnr & \AFcmark & \AFnr \\
HAD (partial track)\newline\citep{yi2021had} & \shortstack[r]{107,224 trials} & Mandarin speech & \AFcmark & \AFxmark & \AFcmark & \AFnr & \AFqmark & \AFnr & \AFcmark & \AFnr \\
ADD 2023 (Track 2)\newline\citep{add2023} & \shortstack[r]{120,917 utt} & Mandarin speech & \AFcmark & \AFxmark & \AFcmark & \AFnr & \AFqmark & \AFnr & \AFqmark & \AFnr \\
ASVspoof 2019\newline\citep{asvspoof2019} & \shortstack[r]{121,461 LA trials\\218,430 PA trials} & Speech / replay & \AFcmark & \AFxmark & \AFxmark & \AFnr & \AFqmark & \AFqmark & \AFcmark & \AFnr \\
ASVspoof 2021\newline\citep{asvspoof2021} & \shortstack[r]{181,566 LA trials\\943,110 PA trials\\611,829 DF trials} & Speech / replay & \AFcmark & \AFxmark & \AFxmark & \AFnr & \AFqmark & \AFqmark & \AFqmark & \AFnr \\
SceneFake\newline\citep{scenefake} & \shortstack[r]{84,480 trials} & Speech + soundscape & \AFcmark & \AFxmark & \AFxmark & \AFnr & \AFqmark & \AFqmark & \AFqmark & \AFnr \\
WaveFake\newline\citep{wavefake} & \shortstack[r]{117,985 fake utt} & English / Japanese & \AFcmark & \AFxmark & \AFxmark & \AFnr & \AFqmark & \AFqmark & \AFcmark & \AFnr \\
\midrule
\multicolumn{11}{@{}l}{\textbf{Image / explanation resources}} \\[1pt]
DDL-I\newline\citep{miao2025ddl} & \shortstack[r]{1.433M img} & Faces / varied images & \AFxmark & \AFcmark & \AFxmark & \AFnr & \AFqmark & \AFqmark & \AFqmark & \AFnr \\
SID-Set (SIDA)\newline\citep{huang2025sida} & \shortstack[r]{300K img\\3K explanations} & Open images & \AFxmark & \AFcmark & \AFxmark & \AFqmark & \AFqmark & \AFnr & \AFqmark & \AFnr \\
MMTD-Set-34k (FakeShield)\newline\citep{xu2025fakeshield} & \shortstack[r]{34,470 ann.} & PS / face / AIGC edit & \AFxmark & \AFcmark & \AFxmark & \AFcmark & \AFqmark & \AFqmark & \AFqmark & \AFnr \\
HydraFake (Veritas)\newline\citep{tan2026veritas} & \shortstack[r]{100,000 img} & Faces; multi-forgery & \AFxmark & \AFcmark & \AFxmark & \AFqmark & \AFqmark & \AFqmark & \AFcmark & \AFqmark \\
Impostor\newline\citep{impostor} & \shortstack[r]{100K fake img} & Open image edits & \AFxmark & \AFcmark & \AFxmark & \AFnr & \AFqmark & \AFqmark & \AFqmark & \AFnr \\
DFBench\newline\citep{dfbench} & \shortstack[r]{540K img} & Open images & \AFxmark & \AFcmark & \AFxmark & \AFnr & \AFqmark & \AFnr & \AFqmark & \AFnr \\
$X^2$-DFD annotation set\newline\citep{chen2025x2dfd} & \shortstack[r]{--} & Faces / VQA & \AFxmark & \AFcmark & \AFxmark & \AFcmark & \AFqmark & \AFqmark & \AFqmark & \AFnr \\
Tariq et al. evaluation\newline\citep{tariq2025llmsdeepfake} & \shortstack[r]{100 img} & Portrait images & \AFxmark & \AFcmark & \AFxmark & \AFnr & \AFnr & \AFnr & \AFnr & \AFnr \\
\bottomrule
\end{tabular}
\end{table*}

\section{Extended Related Work}
\label{sec:related_work_extended}

\subsection{Benchmark scope and temporal supervision}

\paragraph{From facial authenticity to general video forensics.}
Visual deepfake evaluation has largely developed around facial authenticity, as in FaceForensics++, Celeb-DF, and DFDC~\citep{rossler2019faceforensics,celebdf,dfdc}.
DeeperForensics-1.0 and DF40 expand this setting through challenging perturbations and a broader range of manipulation techniques~\citep{deeperforensics,df40}.
Temporal supervision also appears in ForgeryNet, alongside image- and video-level face-forgery tasks~\citep{forgerynet}.
General generated-video resources such as GenVideo, GenVidBench, and AIGVDBench broaden the content and generators considered~\citep{chen2024demamba,ni2026genvidbench,ma2026aigvdbench}.
Moving from wholly generated clips to sparse edits within a video requires both broad content coverage and interval annotations.

\paragraph{Partial speech manipulation and acoustic scenes.}
Speech anti-spoofing provides a parallel route from recording-level decisions to within-recording localization.
ASVspoof 2019 and ASVspoof 2021 establish protocols covering synthesized, converted, replayed, and deepfake speech~\citep{asvspoof2019,asvspoof2021}; WaveFake supplies synthetic speech from multiple generation methods~\citep{wavefake}.
PartialSpoof, HAD, and ADD 2023 Track~2 focus more directly on partial manipulation or the localization of manipulated regions~\citep{zhang2023partialspoof,yi2021had,add2023}.
Here, authentic and manipulated content can coexist within one recording, making global authenticity labels insufficient for boundary estimation.
SceneFake broadens the manipulated content from speech to the surrounding acoustic scene~\citep{scenefake}.
This distinction is relevant to environmental-sound edits, whose forensic evidence need not involve a speaker or lip synchronization.

\paragraph{Audio-visual benchmarks beyond talking heads.}
FakeAVCeleb combines facial and speech manipulations, establishing a multimodal detection setting~\citep{khalid2021fakeavceleb}.
Within speaker-centered video, LAV-DF and AV-Deepfake1M provide content-driven edits with temporal annotations~\citep{cai2022lavdf,cai2024avdeepfake1m}.
AV-Deepfake1M++ extends this line with larger-scale evaluation under realistic perturbations~\citep{cai2025avdeepfake1mpp}.
DDL-AV also retains fine-grained localization annotations across audio, visual, and joint manipulations~\citep{miao2025ddl}, while ArEnAV addresses Arabic--English code-switching and dialect variation~\citep{kuckreja2026arenav}.
More general audio-video content is represented by DDL-GAV, built on MVAD and described in DAV-Det's evaluation~\citep{peng2026davdet}.
The emphasis shifts toward forensic interpretation in X-AVFake and AVFakeBench, which include explanatory annotations or question-based evaluation beyond conventional talking-head detection~\citep{chen2025fakehunter,xia2026avfakebench}.
Omni-Fake brings several media types into one benchmark, but its individual-modality localization tasks differ from its binary audio-video talking-head task~\citep{li2026omnifake}.
These benchmarks vary not only in content coverage but also in whether their labels supervise clip-level decisions or modality-specific intervals.

\paragraph{Long-form context and modality-specific targets.}
Sparse edits within extended authentic context are central to TASLE and ActivityForensics, which study visual manipulation localization in untrimmed or activity-rich videos~\citep{feng2026tasle,bao2026activityforensics}.
AVID brings long-form context to audio-visual inconsistency understanding, including temporal grounding and reasoning about conflicting streams~\citep{chen2026avid}.
This shares the need to identify brief events, but an inconsistency target is not the same as establishing manipulation in each modality.
A replacement can remain semantically compatible with the other stream, and a mismatch need not determine which stream was altered.
AV-FABLE instead assigns executed edits to separate audio and visual timelines, including asynchronous events within the same host video.
Native-track evaluation matches each prediction against its own modality, so a temporally correct prediction on the wrong stream receives no credit.

\paragraph{Reasoning resources and evaluation suites.}
Language-based forensic evaluation also depends on what evidence a model can access.
The datasets accompanying VidGuard-R1 and VideoVeritas, including the latter's MintVid, support reasoning-oriented generated-video detection~\citep{park2025vidguard,tan2026videoveritas}.
EVID-Bench requires web retrieval and cross-video comparison to verify video misinformation, including cases not resolvable from the input video alone~\citep{yu2026evid}.
It therefore differs from detecting local synthesis traces, even when AI-generated content contributes to the misinformation.
DeepfakeBench standardizes detector implementation and evaluation~\citep{deepfakebench}, whereas FVBench assesses the deepfake-video detection capabilities of large multimodal models~\citep{wang2026fvbench}.
The extended comparison in Table~\ref{tab:avc_benchmark_comparison} includes these evaluation resources alongside datasets without assuming that they share a localization protocol.

\paragraph{Spatial localization and explanation precedents.}
Image forensics provides relevant precedents for localized evidence, although its annotations refer to spatial regions rather than temporal intervals.
DDL-I supplies image-forgery data and spatial annotations, complementing the DDL-AV branch~\citep{miao2025ddl}.
SID-Set and MMTD-Set-34k support SIDA and FakeShield, respectively, linking image detection and localization to explanatory supervision~\citep{huang2025sida,xu2025fakeshield}.
HydraFake, introduced with Veritas, emphasizes generalization across forgery techniques and domains~\citep{tan2026veritas}; Impostor addresses realistic localized AI image editing~\citep{impostor}.
DFBench and the evaluation by Tariq et al. assess multimodal models on deepfake-image detection~\citep{dfbench,tariq2025llmsdeepfake}.
The $X^2$-DFD annotation set further supports explainable deepfake detection~\citep{chen2025x2dfd}.
They provide precedents for coupling localized evidence with explanation, while the temporal dimension requires a different annotation target.

\paragraph{Construction controls and generalization.}
The forensic task can be distorted when incidental processing traces predict the label.
Circumventing Shortcuts identifies exploitable artifacts in audio-visual datasets~\citep{smeu2025shortcuts}, while Fake or JPEG? and B-Free examine preprocessing and training-data biases in visual detection~\citep{grommelt2024jpeg,guillaro2025biasfree}.
These findings motivate matched processing of pristine and manipulated examples without implying that every synthesis trace should be removed.
Likewise, a held-out generator tests a different shift from a new source domain or editing route.
When several factors change together, transfer measures the combined setting rather than an isolated causal effect of any one factor.

\subsection{Generation and localized editing}

Video generation models provide general synthesis capabilities~\citep{cogvideox,ltxvideo,wan}, with further advances in controllable generation and world modeling~\citep{ma2025controllable,wang2026subjectanchor,zhang2026zo3t,zhang2025mindv,zhang2026robostereo}.
Related editing methods support controlled and streaming editing~\citep{vace,wang2026liveedit}, physics-grounded manipulation~\citep{hu2026physeditworld}, layer-decomposed editing~\citep{liu2025controllable}, as well as joint editing~\citep{ma2026group}.
Diffusion-based methods also support controllable portrait animation, identity-consistent synthesis, and talking-head generation through visual or audio-visual conditioning~\citep{xu2025hunyuanportrait,hu2026identity,hong2025audio}.
Speech can be replaced using synthesis systems such as F5-TTS and Qwen3-TTS~\citep{f5tts,qwen3tts}.
For non-speech audio, MMAudio and FoleyCrafter condition generation on visual content~\citep{mmaudio,foleycrafter}, allowing an altered soundtrack to remain compatible with the visible scene.
The same generator may produce a fully synthetic clip or a brief replacement inserted into an authentic host.
The latter changes the forensic problem by retaining surrounding context and requiring event boundaries.
In AV-FABLE, the executed edit coordinates define temporal ground truth, while the evidence records describe observable changes within and around those intervals.

\subsection{Temporal localization and forensic representations}

\paragraph{From classification scores to event intervals.}
Single-modality detection provides useful references but does not always train an explicit boundary predictor.
Xception supplies a visual classification backbone~\citep{chollet2017xception}, AASIST targets speech anti-spoofing~\citep{jung2022aasist}, and DeMamba models spatiotemporal evidence in generated videos~\citep{chen2024demamba}.
Applying these detectors to successive windows produces local scores, but their conversion into intervals remains an additional modeling choice.
A correct video-level decision can therefore coexist with inaccurate boundaries or predictions assigned to the wrong modality.
For partially spoofed audio, CFPRF combines frame-level predictions with proposal refinement~\citep{wu2024cfprf}; boundary-aware attention explicitly models genuine--spoofed transitions~\citep{zhong2024boundaryaware}.
These methods are closer to the interval-prediction objective than recording-level classification alone.

\paragraph{Cross-modal evidence and temporal context.}
Audio-visual localization can use relationships between streams as forensic evidence, as in dissonance-based detection~\citep{notmadeforeachother}.
Cross-modal reconstruction and synchronization offer related cues in AuViRe and ICS-Sync~\citep{koutlis2025auvire,anshul2025intra}.
Such cues must still be converted into temporal predictions: BA-TFD and BA-TFD+ emphasize boundary learning~\citep{cai2022lavdf,cai2023glitch}, and HBMNet combines frame-level and proposal-level boundary information~\citep{chen2025hbmnet}.
UMMAFormer and RegQAV develop adaptive multimodal representations and query-based predictions, respectively~\citep{zhang2023ummaformer,zhu2025regqav}.
UniCaCLF relates genuine and forged instants to global context through contrastive learning~\citep{yin2025unicaclf}, while EVAS combines audio-visual interaction with boundary calibration~\citep{shen2026evas}.
Together, these approaches connect evidence aggregation to boundary prediction, establishing context modeling as a relevant precedent for long-form localization.

\paragraph{Temporal decoding and host conditioning.}
Temporal action detection provides architectures that predict event intervals from feature sequences, including ActionFormer and DyFADet~\citep{zhang2022actionformer,yang2024dyfadet}.
Their relevance is architectural: recognizing an action and deciding which part of it was manipulated remain different supervision targets.
Action boundaries need not coincide with manipulation boundaries, even when the same temporal decoding architecture is used.
Feature-wise linear modulation provides a general mechanism for conditioning a representation on contextual information~\citep{perez2018film}.
NFSR uses the input host video's feature statistics to modulate local evidence before modality-specific interval prediction.
The pristine twin supports benchmark construction and annotation, but it is not required as a reference at inference.

\paragraph{Modality-specific decisions.}
Cross-modal evidence is informative but does not establish that both streams have been altered.
DAV-Det models audio and visual forensic evidence independently~\citep{peng2026davdet}, and Divide and Conquer combines modality-specific detectors and localizers through score fusion~\citep{li2026divideconquer}.
Decoupled video-level detection and separate temporal predictions address different output requirements.
The latter accommodates events with different boundaries in each stream, making modality identity part of the localization target.

\paragraph{Frozen representations and language interfaces.}
Representation quality is a different question from the behavior of a language-based prediction interface.
Linear probes assess information accessible through restricted supervised readouts~\citep{alain2016understanding}; self-supervised deepfake studies examine the complementarity of audio and visual features~\citep{boldisor2026selfsupervised}.
Qwen2.5-Omni, Qwen3-Omni, and video-SALMONN offer audio-visual language interfaces~\citep{xu2025qwen25omni,xu2025qwen3omni,tang2025video}, while AV-LMMDetect adapts an AV-LLM for authenticity classification~\citep{cao2026avlmmdetect}.
These settings differ in supervision, so weak zero-shot performance alone cannot establish that the underlying features lack useful forensic information.
Section~\ref{sec:nfsr_exp_id} compares temporal localizers on the same frozen features, separating the choice of encoder from the design of its readout.

\paragraph{Language-assisted localization and explanation.}
A language model may revise intervals or explain them, and these roles should be distinguished.
TASLE's MSLoc uses language-assisted proposal refinement as part of visual localization~\citep{feng2026tasle}.
VidForensics-M1 incorporates verifiable temporal grounding into reinforcement learning through evidence-guided reward redistribution~\citep{liu2026vidforensicsm1}.
PATE-Forensics separates forensic perception from explanation generation~\citep{li2026pateforensics}; FakeHunter retrieves contextual evidence and invokes tools to support multimodal reasoning~\citep{chen2025fakehunter}.
Image-level systems likewise connect detected spatial regions to natural-language explanations~\citep{huang2025sida,xu2025fakeshield,chen2025x2dfd}.
More broadly, hierarchical multimodal reasoning workflows decompose
complex interpretations into staged cognitive steps, providing a useful
precedent for structuring multimodal explanation pipelines~\citep{huang2025affective}.
In the separate explanation evaluation of Appendix~\ref{app:avc_explanation}, language outputs describe fixed events and do not alter localization or detection decisions.
Fluent descriptions do not themselves establish that predicted intervals or modality labels are correct, motivating separate evaluation of these outputs.

\FloatBarrier
\section{Evaluation and Implementation}
\label{app:implementation_details}
\label{app:avc_protocol}
\subsection{Splits and OOD Evaluation Protocol}
All twins and derived samples from the same host video remain in one split.
Training, validation, and ID testing use ActivityNet and EPIC-KITCHENS
with shared manipulation categories and editing configurations. The three
OOD settings are defined by their held-out components. Source transfer
uses HDTF and a source-disjoint TASLE subset. Held-out generators retain
editing routes represented in training while using unseen generators.
Held-out TI2V jointly withholds the text-and-image-conditioned video
generation route and its generator, Wan2.2-TI2V-5B. This setting evaluates
transfer to that route--generator configuration, without separating their
contributions to the distribution shift.
The calibration subset contains 256 pristine and 256 manipulated videos
from the source-validation set used for checkpoint selection. Thresholds
are estimated from its pristine videos only; test and OOD data are used
solely for evaluation.
\begin{table}[!htbp]
\centering
\AVCtablefont
\caption{OOD evaluation sets and their held-out components. Source-transfer detection and Joint-timeline results average HDTF and TASLE with equal domain weight, whereas native A-mAP uses HDTF and native V-mAP pools both domains; held-out-generator and TI2V results are reported on their respective sets.}
\label{tab:avc_ood}
\begingroup
\adjustbox{max width=\linewidth}{%
\begin{tabular}{@{}lrrl@{}}
\toprule
\rowcolor{AVCTableHead}
Setting / domain & Manipulated & Pristine & Shift \\
\midrule
Source / HDTF & 386 & 386 & New domain; predominantly audio edits \\
Source / TASLE & 358 & 360 & External dataset; visual edits \\
Generator / three models & 311 & 311 & Seen routes, new generators \\
TI2V / Wan2.2 & 300 & 300 & Held-out route and generator \\
\bottomrule
\end{tabular}
}
\endgroup
\end{table}
The source-transfer setting combines different domains and modality
compositions: HDTF contains
384 audio and two visual manipulated examples, while TASLE supplies
visual forgeries. The TASLE subset contains 718 videos after excluding
54 examples from 15 EPIC source recordings shared with AV-FABLE training, using
source metadata to determine membership. The held-out-generator set contains
311 of 317 candidate pairs after deduplication and restriction to common
prediction coverage. TASLE is external to the AV-FABLE release counts.
Equal weighting of HDTF and TASLE prevents the larger source subset
from dominating the source-transfer result. Attribution macro-F1 is computed
on the pooled 1,490 source videos instead, because each domain alone lacks
some attribution classes; the pooled set contains clean, audio-only, and
visual-only videos but no both-track videos (Eq.~\ref{eq:attr_f1}). Native localization cannot use equal domain
weighting either: HDTF contains all 384 audio events of the setting but
only two visual events. Source-transfer A-mAP is therefore computed on
HDTF, and V-mAP on the 393 pooled visual events of both domains. The resulting source-transfer scores reflect
both domain and modality composition. Because none of the three settings
isolates a single shift factor, differences between their scores do not
rank the difficulty of individual shift factors.
Each OOD setting uses the same
checkpoint, decoding procedure, and source-calibrated thresholds.

\subsection{Scoring and Operating Points}
Ranking and fixed-threshold evaluation measure complementary aspects of
temporal prediction. Native-track mAP averages one-to-one,
confidence-ranked interval AP over
tIoU thresholds $0.30{:}0.05{:}0.70$. Matching is within the same video
and modality; unmatched and duplicate predictions, including predictions
on pristine twins, count as false positives.
At the operating point, Event-IoU averages each ground-truth event's best
overlap with submitted same-track intervals, assigning zero to misses.
Both video score and interval confidence must strictly exceed the
source threshold. Event-IoU measures coverage; mAP additionally accounts
for extra predictions.

A track score is the highest post-Soft-NMS proposal confidence,
$S_m(X)=\max_i\hat c_i^m$, or zero if no proposal exists. The branches
can have different genuine-score distributions. We place their video
scores on a common reference scale using $n=256$ pristine calibration videos:
\begin{equation}
\begin{aligned}
p_m(X)&=\frac{1+\sum_{i=1}^n\mathbf1[S_m(X_i^{\mathrm{cal}})\geq S_m(X)]}{n+1},\\
S_{\mathrm{any}}(X)&=-\min\{p_a(X),p_v(X)\}.
\end{aligned}
\label{eq:tail_probability}
\end{equation}
Here, $X_i^{\mathrm{cal}}$ denotes a pristine source-validation video and
$\mathbf1[\cdot]$ is the indicator function. Smaller $p_m$ indicates a
more extreme score relative to genuine videos, so larger $S_{\mathrm{any}}$
indicates stronger evidence of manipulation in either track.
At budget $\alpha$, the threshold is the
$(\lfloor n\alpha\rfloor+1)$-st largest pristine calibration score,
with strict exceedance; at 5\% this is the thirteenth-largest score.
Any-manipulation detection uses $S_{\mathrm{any}}>\tau_{\mathrm{any}}$,
with its own calibrated threshold. Attribution instead uses
$d_m=\mathbf1[S_m>\tau_m]$: $(d_a,d_v)=(0,0),(1,0),(0,1),(1,1)$ correspond
to clean, audio-only, visual-only, and both-track manipulation, respectively.
For an evaluation set $D$ with ground-truth class set $\mathcal C_D$,
\begin{equation}
\mathrm{F1}_{\mathrm{attr}}(D)=\frac{1}{|\mathcal C_D|}\sum_{c\in\mathcal C_D}\mathrm{F1}_c,
\label{eq:attr_f1}
\end{equation}
where $\mathrm{F1}_c$ is the one-versus-rest F1 of class $c$. A prediction
of a class outside $\mathcal C_D$ still lowers the recall of the true
class. Classes without ground-truth support are excluded because their
recall is undefined; including them with zero F1 would only rescale the
score by $|\mathcal C_D|/4$. Here, $\mathcal C_D$ contains all four
classes on ID and held-out generators, three on pooled source transfer
(746 clean, 384 audio-only, and 360 visual-only videos), and two on
held-out TI2V (300 clean and 300 visual-only videos).
The decision-level combination leaves native event intervals and proposal
scores unchanged.
Recall and realized FAR are measured on each evaluation population,
including distribution shifts, at this fixed source-domain operating point.

For detection, attribution, and Joint-timeline metrics, the OOD
macro-average (OOD-All) weights source transfer, held-out generators, and
held-out TI2V equally; for attribution, it averages
per-setting scores computed over different class sets. Native OOD A-mAP averages
source transfer and held-out generators, the settings with audio edits,
and native OOD V-mAP averages all three settings. OOD AV-mAP averages the
per-setting AV-mAP over the three settings; in each setting, AV-mAP
averages A-mAP and V-mAP over the tracks with annotated events, so it
equals V-mAP on held-out TI2V.
Variation across training runs is reported as sample standard deviation;
pair-bootstrap intervals quantify uncertainty over sampled video pairs.

\subsection{Native-Track and Joint-Timeline Evaluation}
\label{app:avc_timeline}
Temporal overlap alone does not establish which modality was manipulated.
We quantify this distinction on all 4,694 ID test videos using predictions
from the three NFSR training seeds. For each model, we evaluate the complete
prediction set and two subsets retaining only audio or visual proposals,
without changing their boundaries. Native-track evaluation matches
predictions to ground-truth events on the same track. Joint-timeline
evaluation instead merges overlapping audio and visual ground-truth
intervals within each video and matches proposals without modality labels.
Both use mAP@[0.30:0.05:0.70].

For joint-timeline evaluation of the complete prediction set, proposal
confidences are mapped to track-specific empirical anomaly percentiles using
256 pristine source-validation videos, allowing cross-track ranking.
Single-track prediction sets and native-track evaluation use the original
confidences. This protocol defines the Joint-mAP reported in
Table~\ref{tab:nfsr_id}.
\begin{table}[!htbp]
\centering
\AVCtablefont
\caption{\textbf{Modality coverage under native-track and joint-timeline evaluation.}
Results are mean $\pm$ sample standard deviation over three seeds (\%).
A missing prediction track has zero native mAP.}
\label{tab:avc_timeline}
\begingroup
\adjustbox{max width=\linewidth}{%
\begin{tabular}{@{}lrrr@{}}
\toprule
\rowcolor{AVCTableHead}
Predictions & A-mAP & V-mAP & Joint-mAP \\
\midrule
Audio + visual & $92.24\pm0.16$ & $90.36\pm0.28$ & $72.33\pm0.10$ \\
Audio only & $92.24\pm0.16$ & 0.00 & $57.36\pm0.12$ \\
Visual only & 0.00 & $90.36\pm0.28$ & $64.04\pm0.23$ \\
\bottomrule
\end{tabular}}
\endgroup
\end{table}

The visual-only subset reaches 64.04 Joint-mAP, above every visual-only
baseline in Table~\ref{tab:nfsr_id}, and the audio-only subset reaches
57.36, above every audio-only baseline, although each subset produces no
events on the discarded track (Table~\ref{tab:avc_timeline}). Native-track
evaluation makes this missing coverage explicit: the absent track receives
zero mAP, while performance on the retained track is unchanged. A high
Joint-mAP can thus coexist with a complete modality-specific blind spot.
Separate timelines preserve the distinction between locating a
manipulated interval and identifying its affected track.
Merging also penalizes complete predictions on both tracks, which explains
why the complete model's Joint-mAP is below both native-track scores. In
727 of the 818 ID both-track videos, the audio and visual events overlap
and merge into one reference interval, so the co-timed proposal from the
second track counts as a duplicate false positive; without merging, the
same predictions reach 91.28 mAP. Joint-mAP therefore combines boundary
quality, cross-track confidence ranking, and redundancy between tracks.
We use the native-track A-/V- metrics as the primary localization measures
and report Joint-timeline metrics as an auxiliary view that also covers
outputs without modality labels.
Table~\ref{tab:avc_ood_joint} reports Joint-timeline metrics and native-track
Event-IoU for the OOD comparison in Table~\ref{tab:nfsr_ood}, and
Table~\ref{tab:avc_ablation_aux} reports them for the ablation. Under
Joint-mAP, NFSR also exceeds UMMAFormer in all three settings.
\begin{table}[!htbp]
\centering
\AVCtablefont
\caption{\textbf{Auxiliary OOD localization metrics (\%).}
Joint-mAP and Joint-IoU evaluate the joint timeline, with source-transfer
values averaging HDTF and TASLE with equal weight. A-IoU and V-IoU are
native-track Event-IoU at the source-calibrated operating point, with the
source-transfer composition of Table~\ref{tab:nfsr_ood}. Rows and -- follow
Table~\ref{tab:nfsr_ood}.}
\label{tab:avc_ood_joint}
\begingroup
\setlength{\tabcolsep}{2.0pt}
\adjustbox{max width=\linewidth}{%
\begin{tabular}{@{}ll*{11}{r}@{}}
\toprule
\rowcolor{AVCTableHead}
& & \multicolumn{4}{c}{Source transfer} & \multicolumn{4}{c}{Held-out generators} & \multicolumn{3}{c}{Held-out TI2V} \\
\cmidrule(lr){3-6}\cmidrule(lr){7-10}\cmidrule(l){11-13}
Method & Training Form & Joint-mAP & Joint-IoU & A-IoU & V-IoU & Joint-mAP & Joint-IoU & A-IoU & V-IoU & Joint-mAP & Joint-IoU & V-IoU \\
\midrule
Xception & Official weights & 0.86 & 0.42 & -- & 0.32 & 0.23 & 0.26 & -- & 0.19 & 0.25 & 0.12 & 0.12 \\
MSLoc-PG & Official weights & 26.78 & 19.37 & -- & 38.53 & 4.62 & 5.78 & -- & 9.18 & 0.92 & 2.23 & 2.23 \\
ActionFormer & TASLE & 37.98 & 20.04 & -- & 39.87 & 13.89 & 7.54 & -- & 10.31 & 15.15 & 7.25 & 6.73 \\
Xception & AV-FABLE & 1.73 & 0.30 & -- & 0.47 & 0.41 & 0.35 & -- & 0.49 & 0.67 & 0.80 & 0.80 \\
DeMamba (Stage-1 window) & AV-FABLE & 0.29 & 0.70 & -- & 0.94 & 1.77 & 0.72 & -- & 1.19 & 2.17 & 1.12 & 1.12 \\
ActionFormer & AV-FABLE & 21.38 & 13.32 & -- & 26.27 & 33.23 & 24.79 & -- & 40.03 & 61.57 & 42.93 & 42.93 \\
\midrule
AASIST & Official weights & 0.79 & 0.00 & 0.00 & -- & 0.27 & 0.24 & 0.42 & -- & 0.28 & 0.26 & -- \\
AASIST & AV-FABLE & 1.10 & 7.51 & 15.09 & -- & 0.51 & 3.53 & 8.01 & -- & 0.48 & 0.83 & -- \\
\midrule
Qwen2.5-Omni-7B (ZS) & Official weights & 0.89 & 0.00 & 0.00 & 0.00 & 0.73 & 0.00 & 0.00 & 0.00 & 0.86 & 0.00 & 0.00 \\
Qwen3-Omni-30B-A3B (ZS) & Official weights & 1.15 & 3.48 & 0.07 & 6.74 & 0.04 & 0.47 & 0.09 & 0.69 & 0.04 & 0.38 & 0.38 \\
video-SALMONN 2+ 7B (ZS) & Official weights & 0.00 & 0.00 & 0.00 & 0.00 & 0.00 & 0.04 & 0.11 & 0.07 & 0.00 & 0.03 & 0.03 \\
DAV-Det & Official weights & 10.53 & 23.42 & 0.00 & 46.60 & 1.12 & 1.14 & 0.14 & 1.05 & 1.18 & 0.92 & 0.81 \\
BA-TFD+ & AV-FABLE & 0.85 & 0.99 & 0.65 & 1.32 & 0.55 & 4.86 & 10.15 & 0.26 & 0.07 & 0.61 & 0.46 \\
UMMAFormer + Qwen3 & AV-FABLE & 61.15 & 50.45 & 81.64 & 22.23 & 34.29 & 15.96 & 26.69 & 15.12 & 22.39 & 4.96 & 12.30 \\
Qwen3 direct timestamp SFT & AV-FABLE & 15.80 & 26.31 & 52.07 & 1.96 & 6.40 & 10.05 & 24.11 & 1.84 & 0.31 & 0.31 & 1.21 \\
AuViRe & AV-FABLE & 0.21 & 0.00 & -- & -- & 0.39 & 2.39 & -- & -- & 0.07 & 0.75 & -- \\
ICS-Sync & AV-FABLE & 1.19 & 0.13 & -- & -- & 1.13 & 1.99 & -- & -- & 0.74 & 0.74 & -- \\
DAV-Det + AV-FABLE & AV-FABLE & 8.92 & 25.78 & 0.14 & 49.25 & 3.48 & 4.93 & 2.32 & 5.56 & 7.63 & 5.51 & 5.51 \\
NFSR + Qwen3 (Ours) & AV-FABLE & 73.70 & 56.22 & 90.86 & 26.42 & 48.55 & 26.03 & 33.40 & 32.92 & 64.92 & 38.83 & 41.21 \\
\bottomrule
\end{tabular}}
\endgroup
\end{table}
\begin{table}[!htbp]
\centering
\AVCtablefont
\caption{\textbf{Auxiliary metrics for the ablation.}
Mean $\pm$ sample standard deviation over three seeds (\%). OOD-All
Joint-timeline metrics average the three OOD settings equally; A-IoU and
V-IoU follow the native-track OOD-All definition of
Table~\ref{tab:nfsr_ablation}.}
\label{tab:avc_ablation_aux}
\begingroup
\adjustbox{max width=\linewidth}{%
\begin{tabular}{@{}lrrrrrr@{}}
\toprule
\rowcolor{AVCTableHead}
& \multicolumn{2}{c}{ID} & \multicolumn{4}{c}{OOD-All} \\
\cmidrule(lr){2-3}\cmidrule(l){4-7}
Variant & Joint-mAP & Joint-IoU & Joint-mAP & Joint-IoU & A-IoU & V-IoU \\
\midrule
w/o intermediate-layer features & $73.41\pm0.04$ & $73.35\pm0.83$ & $53.22\pm0.40$ & $35.07\pm2.02$ & $56.98\pm1.36$ & $27.74\pm1.68$ \\
w/o interface features & $77.49\pm1.09$ & $81.91\pm1.22$ & $57.00\pm0.26$ & $43.46\pm1.04$ & $62.72\pm1.33$ & $36.33\pm2.80$ \\
w/o host conditioning & $72.93\pm1.85$ & $78.86\pm0.58$ & $56.80\pm1.41$ & $37.01\pm2.45$ & $58.97\pm2.69$ & $28.44\pm0.66$ \\
Full NFSR & $72.33\pm0.10$ & $81.21\pm0.14$ & $62.39\pm0.21$ & $40.36\pm1.63$ & $62.13\pm0.71$ & $33.52\pm1.36$ \\
\bottomrule
\end{tabular}}
\endgroup
\end{table}

\subsection{Readout and Baselines}
The Qwen3-Omni encoders are frozen. Cached features use a 0.25-second grid:
visual intermediate-layer $[N,16,1152]$, visual interface $[N,2048]$,
audio intermediate-layer $[N,1280]$, and audio interface $[N,2048]$,
where $N$ denotes the number of temporal positions.
The audio intermediate layer is layer 16 (of 32) in both tested Qwen-Omni
generations. The readout uses these fixed layers; layer selection is
evaluated separately using training-video cross-validation
(Appendix~\ref{app:avc_representations}).

\paragraph{Spatial pooling and gated projection.}
A learned linear scorer assigns a weight to each of the 16 visual cells;
softmax across cells yields the weighted-sum intermediate-layer vector
$\mathbf d_t^v$. Audio intermediate-layer vectors require no spatial
pooling. With
$\mathbf q_t^m=[\mathbf d_t^m;\mathbf u_t^m]$, the fusion is
\begin{equation}
 \mathbf h_t^m=\operatorname{Dropout}\!\left(
 P_m\!\left(\operatorname{LN}(\mathbf q_t^m)\right)
 \odot\operatorname{sigmoid}\!\left(G_m(\mathbf q_t^m)\right)
 \right).
 \label{eq:native_fusion}
\end{equation}
Here, $P_m$ and $G_m$ are learned affine projections to $D=256$ channels,
and LN denotes layer normalization. The gate modulates projected channels;
dropout uses probability 0.1. The two branches have separate parameters.

\paragraph{Host statistics and modulation.}
For the valid temporal positions $\mathcal T_X$, let $M=|\mathcal T_X|$.
Channel-wise statistics are computed as
\begin{equation}
\begin{aligned}
 \boldsymbol\mu_m&=\frac{1}{M}\sum_{t\in\mathcal T_X}\mathbf h_t^m,\\
 \boldsymbol\sigma_m&=\left(
 \frac{1}{M}\sum_{t\in\mathcal T_X}
 (\mathbf h_t^m-\boldsymbol\mu_m)^{\odot2}
 +\epsilon\mathbf1\right)^{1/2}.
\end{aligned}
\label{eq:host_statistics}
\end{equation}
Here, powers and roots are element-wise and $\epsilon=10^{-6}$ stabilizes
the standard deviation. Padding is excluded from both statistics.
A two-layer GELU MLP produces the scale and shift in
Equation~\ref{eq:host_film}; the input video's statistics are recomputed
at inference while learned parameters remain fixed.

\paragraph{Temporal decoding and supervision.}
The audio/visual pyramids have six/five levels with stride-two downsampling,
masked dilated convolutions, and dynamic classification/regression heads
\citep{zhang2022actionformer,yang2024dyfadet}.
We adopt selective state space modeling~\citep{gu2023mamba} as an efficient temporal backbone; compared with standard Transformers, its linear-time sequence scaling and continuous hidden-state transitions effectively preserve extended host context while sensitively isolating brief, local forgery artifacts~\citep{chen2024demamba,mambacontour}.
A positive point lies within the event's center-sampling window
(radius 1.5 level strides, clipped to the event) and satisfies the level's
regression range. For level $\ell$, the lower distance bound is zero at
$\ell=0$ and $2^{\ell+1}$ otherwise; the upper bound is $2^{\ell+2}$,
or 10,000 at the final level, in 0.25-second grid units. The larger
left/right target distance is checked against these bounds; overlapping
eligible events are resolved by selecting the shortest. Regression targets
are normalized by level stride. Unassigned valid points are negatives,
and padded points are excluded. Classification and regression losses share
a running positive-count normalizer (initial value 100, momentum 0.9).
Sigmoid focal loss uses $\alpha=0.25$ and $\gamma=2$; regression is evaluated
only at positive points. Feature fusion, host conditioning, and temporal
decoding are trained jointly within each branch.
\begin{table}[!htbp]
\centering
\AVCtablefont
\caption{NFSR readout architecture, optimization, and decoding settings.}
\label{tab:avc_recipe}
\begingroup
\adjustbox{max width=\linewidth}{%
\begin{tabular}{@{}L{0.25\linewidth}L{0.67\linewidth}@{}}
\toprule
\rowcolor{AVCTableHead}
Setting & Value \\
\midrule
Width / dropout & 256 / 0.1 \\
Optimizer & AdamW; learning rate $10^{-4}$; weight decay 0.025 \\
Training schedule & 12 epochs; 2 warmup epochs; cosine decay; batch 8 \\
Stabilization & Gradient clipping 1.0; EMA 0.999 \\
Heads / assignment & Three layers; kernel 5; center-sampling radius 1.5 \\
Losses & Sigmoid focal classification + 1D DIoU regression \\
Decoding & Pre-NMS 0.001; top-$k$ 2,000; Soft-NMS $\sigma=0.5$ \\
Boundary voting & tIoU 0.7 \\
Model selection & Validation mAP averaged over supported tracks \\
Seeds & Three training seeds \\
Core ablation & Same three seeds and 12-epoch budget \\
\bottomrule
\end{tabular}
}
\endgroup
\end{table}
Focal loss, DIoU, and Soft-NMS follow
\citet{lin2017focal,zheng2020diou,bodla2017soft}.
The highest-scoring valid proposals from each level enter temporal Soft-NMS
with boundary voting. Retained intervals are converted to seconds and clipped
to video duration. Empty or zero-length intervals are discarded. The
source-validation localization score selects checkpoints before test and
OOD evaluation.

\paragraph{Baseline adaptation.}
Baseline inputs follow their supported modalities, with architecture-specific
adaptation schedules. Xception uses two full frames per two-second cell and full-model
AdamW fine-tuning (learning rate/weight decay $10^{-5}$, batch 32,
up to 20 epochs, validation cell-AUROC selection).
The AV-FABLE-trained DeMamba Stage-1 baseline uses eight frames per
two-second window, four training epochs, about 42.9k updates, and fixed
proposal conversion. The MSLoc-PG row instead uses the TASLE-trained
proposal generator; the released proposal-plus-refinement system is
evaluated in Appendix~\ref{app:avc_external}.
DAV-Det adaptation trains LoRA adapters in the official audio and visual
encoders for up to ten epochs with validation-AUROC selection. Both
DAV-Det rows score consecutive three-second windows on each track, with
four frames per visual window; each window becomes an interval proposal
with its probability as confidence, and the maximum window probability
gives the track score.
BA-TFD+ uses official visual/audio encoders, $96^2$ full frames at
2 fps, four epochs and 87,632 updates.
The AV-FABLE-trained ActionFormer is trained on the complete training set for 24 epochs. UMMAFormer uses the same frozen Qwen3 features as
NFSR, a 12-epoch budget, and validation-selected EMA weights.
NFSR, UMMAFormer, and the TASLE-trained ActionFormer are evaluated over
three training seeds; the AV-FABLE-trained ActionFormer and the other
adapted baselines use one seed.
Native-track scores are reported for methods producing modality-specific
intervals, including timestamped language events with modality labels.
Unified predictions without modality labels are scored only by
joint-timeline evaluation; output types are given in the Output column of
Table~\ref{tab:nfsr_id}.
For single-modality methods, events on the unsupported track remain
misses in joint-timeline evaluation rather than being excluded.

\FloatBarrier
\section{Forensic Evidence in Frozen Representations}
\label{app:avc_representations}
We examine the representation evidence underlying NFSR through linear
probes, encoder replacement, and decomposition of features across depth.
\subsection{Probe Design and Layer Selection}
To distinguish learned forensic discrimination from generic feature
anomalies, we compare within-video normalized feature deviation, temporal
feature jumps, and a supervised linear probe \citep{alain2016understanding}.
Logistic regression uses 3,000 training videos (1,500 per class), all
positive cells, a 2:1 negative-to-positive cell ratio, standardized
features, and L2 regularization $10^{-3}$.
Audio inputs concatenate intermediate-layer and interface features; the visual
probe uses interface features. Evaluation measures cell-level forensic
separability.
\begin{table}[!htbp]
\centering
\AVCtablefont
\caption{Cell-level forensic separability: AUROC (\%). OOD denotes the
OOD pool (Table~\ref{tab:avc_metadata}).}
\label{tab:avc_probes}
\begingroup
\adjustbox{max width=\linewidth}{%
\begin{tabular}{@{}lrrr@{}}
\toprule
\rowcolor{AVCTableHead}
Track / split & Deviation & Jump & Linear \\
\midrule
Audio / ID & 68.5 & 44.7 & 90.8 \\
Audio / OOD & 56.2 & 47.9 & 77.3 \\
Visual / ID & 62.5 & 49.8 & 80.8 \\
Visual / OOD & 64.4 & 53.7 & 68.8 \\
\bottomrule
\end{tabular}
}
\endgroup
\end{table}
\paragraph{Discriminative structure beyond feature anomalies.}
Linear probes outperform both unsupervised scores on every track and
split (Table~\ref{tab:avc_probes}). Audio AUROC reaches 90.8 on ID and
77.3 on OOD, while temporal jumps remain near chance. Supervised probes
therefore recover discriminative signals along a learned feature-space
direction, beyond generic deviations or changes, and these signals
transfer to the OOD pool without updating the encoder.

\begin{figure}[!htbp]
\centering
\includegraphics[width=\linewidth]{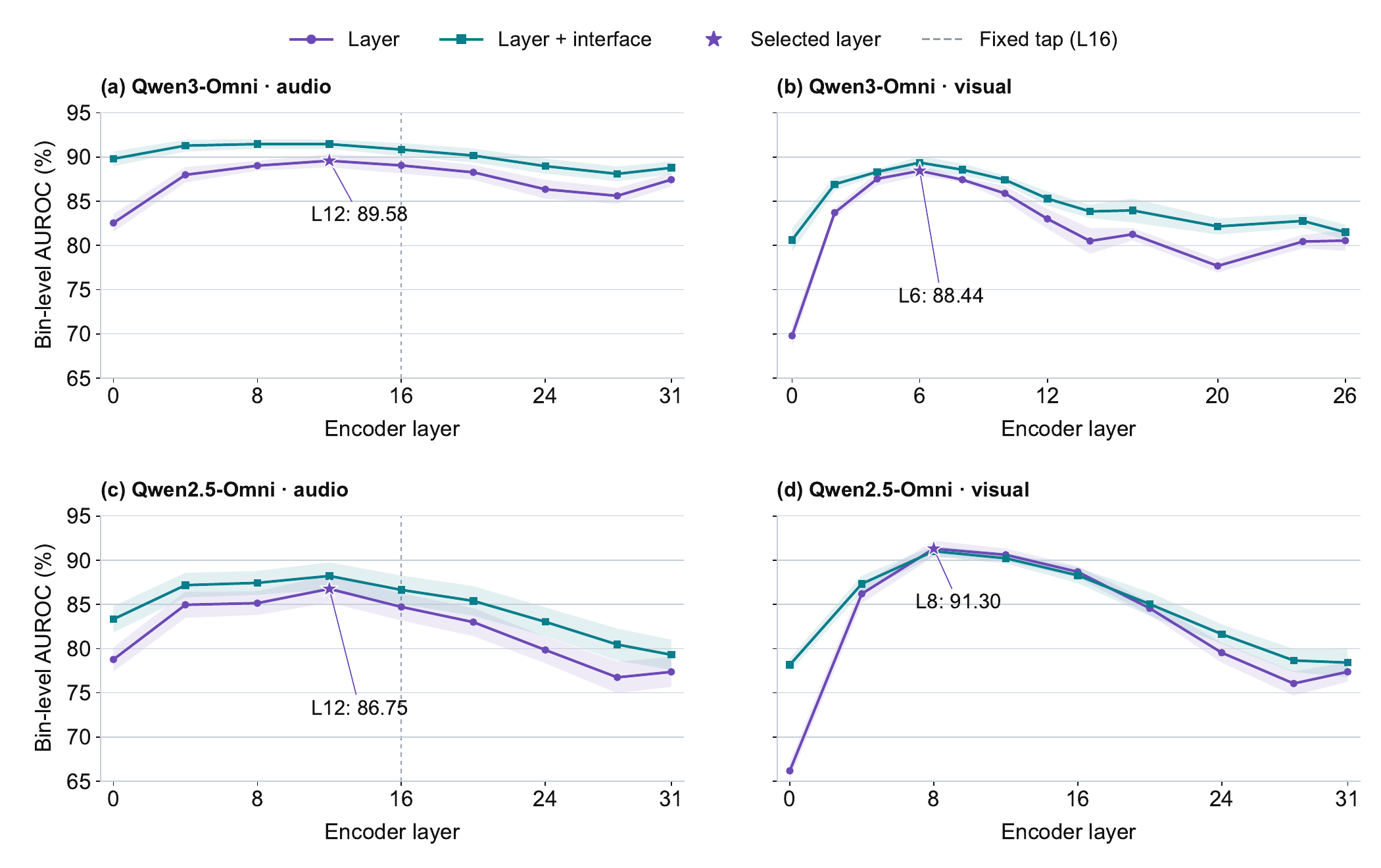}
\caption{Layer-wise forensic separability of Qwen3-Omni (a, b) and
Qwen2.5-Omni (c, d). Each panel plots cell-level probe AUROC for each layer
alone and concatenated with interface features; shading shows
cross-validation variation, stars mark the selected layer, and dashed lines
mark the fixed audio layer (L16). Visual features are spatially pooled; audio and
visual depth indices are separate.}
\label{fig:avc_layers}
\end{figure}

\paragraph{Forensic accessibility across depth.}
Layer selection uses five-fold cross-validation grouped by training video
and selects the layer with the highest held-out cell AUROC. It selects
audio layer 12 in both Qwen-Omni generations: Qwen3 obtains
$89.6\pm0.7$ compared with $89.1\pm0.9$ at layer 16, and Qwen2.5
obtains $86.8\pm1.6$ compared with $84.7\pm1.5$.
These results locate useful audio evidence within the perception encoder
and show that its accessibility varies with representation depth.
For Qwen3-Omni, layers 12 and 16 differ by 0.5 AUROC, within fold
variation, so NFSR retains layer 16 rather than tuning the layer on probe
results (Figure~\ref{fig:avc_layers}).

The Qwen3-Omni visual sweep applies the same grouped five-fold procedure
to spatially mean-pooled features from 700 training twin pairs, keeping
both members of each pair in the same fold. Among 12 probed visual
blocks (zero-indexed; the last is block 26), block 6 yields the highest
cell AUROC:
$88.44\pm0.58$ alone and $89.37\pm0.74$ when concatenated with
post-merger interface features (mean $\pm$ standard deviation across
folds). Later blocks fluctuate at lower values, with the final block
reaching $80.55\pm1.16$ alone. Both visual variants therefore peak
before the final block: greater encoder depth does not necessarily
make forensic evidence more linearly accessible.

\subsection{Encoder Replacement}
We apply temporal readouts to frozen encoders with different pretraining
objectives to assess how the available representation affects localization
and transfer. Table~\ref{tab:avc_encoders} reports native-track mAP so
that audio and visual behavior can be examined separately.
\begin{table}[!htbp]
\centering
\AVCtablefont
\caption{Frozen-encoder comparison: native-track mAP (\%). HDTF reports audio
localization on the held-out talking-head domain; generator and TI2V
columns report their respective modality-specific evaluations.}
\label{tab:avc_encoders}
\begingroup
\adjustbox{max width=\linewidth}{%
\begin{tabular}{@{}lrrrrrr@{}}
\toprule
\rowcolor{AVCTableHead}
Encoder & ID A & ID V & HDTF A & Gen. A & Gen. V & TI2V V \\
\midrule
Whisper-large-v3~\citep{radford2023robust} & 77.3 & -- & 86.8 & 31.4 & -- & -- \\
BEATs~\citep{chen2022beats} + CLIP~\citep{radford2021learning} & 93.9 & 83.4 & 85.9 & 56.1 & 49.2 & 49.6 \\
Qwen2.5-Omni-7B & 87.5 & 93.7 & 94.6 & 37.2 & 58.8 & 72.3 \\
Qwen3-Omni-30B & 92.2 & 90.4 & 98.4 & 48.9 & 49.3 & 68.1 \\
\bottomrule
\end{tabular}
}
\endgroup
\end{table}
Both Qwen-Omni generations support localization on held-out hosts and
editing configurations. Qwen3 leads on HDTF audio, Qwen2.5 on the visual
evaluations, and BEATs+CLIP on ID and generator-shifted audio. Thus,
forensic readout extends across these encoders, with complementary
strengths by modality and domain. Qwen2.5/Qwen3 use two/three seeds;
their generator-shift audio/visual mAP standard deviations are
1.5/3.3 and 1.7/0.7, respectively.

\subsection{Complementary Evidence across Representation Stages}
\label{app:avc_residual}
To characterize the evidence shared between representation stages, we
decompose audio intermediate-layer features into a component predicted from
the interface and a residual. A ridge map
$E:\mathbb R^{2048}\rightarrow\mathbb R^{1280}$ is fitted from interface
features $\mathbf a^s$ to intermediate-layer features $\mathbf a^d$ using 636,000
pristine training cells. The regularizer is
$10^{-3}\operatorname{tr}(\Sigma_{ss})/2048$. It separates a linearly
reconstructed component from a residual:
\begin{equation}
 \mathbf e_t=E(\mathbf a_t^s),\qquad
 \mathbf r_t=\mathbf a_t^d-E(\mathbf a_t^s).
 \label{eq:avc_residual}
\end{equation}
Here, $\mathbf e_t$ is the component reconstructed from interface features
and $\mathbf r_t$ is the remaining intermediate-layer feature vector.
At layer 16, reconstruction on held-out pristine cells gives $R^2=0.71$.
Linear probes evaluate the two components and their concatenations with
the interface. Temporal decoding is also evaluated with the residual in
place of the raw intermediate-layer features under a 12-epoch training budget.

\begin{table}[!htbp]
\centering
\AVCtablefont
\caption{Audio representation decomposition. Top: cell-level probe AUROC
(\%). Bottom: temporal mAP (\%) under a 12-epoch budget. Intermediate +
interface reports three-seed mean and standard deviation; alternatives use one seed.
EPIC and ANet-gen denote held-out-generator edits on EPIC-KITCHENS and
ActivityNet hosts; OOD pool is defined in Table~\ref{tab:avc_metadata}.}
\label{tab:avc_residual}
\begingroup
\adjustbox{max width=\linewidth}{%
\begin{tabular}{@{}lrrrrr@{}}
\toprule
\rowcolor{AVCTableHead}
Probe input & ID & HDTF & XTTS-v2 & EPIC & ANet-gen \\
\midrule
Interface $\mathbf a^s$ & 86.1 & 84.6 & 84.2 & 84.8 & 68.7 \\
Intermediate $\mathbf a^d$ & 89.4 & 89.1 & 93.2 & 83.4 & 71.3 \\
Residual $\mathbf r$ & 84.1 & 84.8 & 88.0 & 71.5 & 61.0 \\
Reconstructed $E(\mathbf a^s)$ & 86.0 & 85.3 & 85.3 & 84.6 & 68.8 \\
$[\mathbf r,\mathbf a^s]$ & 90.6 & 92.3 & 93.0 & 84.4 & 72.1 \\
$[\mathbf a^d,\mathbf a^s]$ & 90.7 & 92.4 & 93.2 & 85.1 & 71.9 \\
Residual norm $\|\mathbf r\|_2^2$ & 45.7 & 48.1 & 51.5 & 61.5 & 47.8 \\
\midrule
Decoder input & ID & HDTF & XTTS-v2 & Omni2Sound & OOD pool \\
\midrule
Intermediate + interface & $92.2\pm0.2$ & $98.4\pm0.6$ & $92.3\pm0.6$ & $41.4\pm1.1$ & $85.6\pm1.0$ \\
Intermediate only & 91.5 & 98.5 & 91.2 & 37.8 & 84.4 \\
Residual + interface & 88.0 & 93.2 & 82.3 & 28.6 & 76.6 \\
\bottomrule
\end{tabular}}
\endgroup
\end{table}

\paragraph{Predictive information spans both components.}
The residual remains discriminative on HDTF and XTTS-v2, but its norm
is a weak authenticity score (Table~\ref{tab:avc_residual}). The relevant
signal is therefore structured within the residual rather than captured
by its magnitude. The reconstructed component is also predictive, and
the original intermediate-layer features outperform the residual alone on
every listed domain. Concatenating intermediate-layer and interface features
further
improves probe AUROC on ID, HDTF, and ANet-gen, consistent with
complementary evidence at the two stages.

\paragraph{Feature accessibility and temporal decoding are distinct.}
For a fixed $E$, $[\mathbf r,\mathbf a^s]$ and
$[\mathbf a^d,\mathbf a^s]$ are invertible linear reparameterizations
with similar probe scores. Nevertheless, residual inputs reduce the
temporal decoder's OOD-pool mAP from 85.6 to 76.6 under the tested
budget. A linearly information-preserving transformation need not be
equally effective for temporal decoding, supporting the use of the
original features rather than subtraction of their shared component.

\FloatBarrier
\section{Sparse Evidence in Long Videos}
\label{app:avc_scaling}
Sparse manipulation requires preserving local evidence while controlling
false alarms over long authentic context. Mean aggregation and local
triggering exhibit complementary failure modes in this setting.
Consider a video of $T$ temporal cells with a manipulated event occupying
$L\ll T$ cells. Suppose manipulated cells have a local evidence mean shift
$\delta$, and the whole-video mean has variance $\sigma^2/T_{\mathrm{eff}}$,
where $T_{\mathrm{eff}}$ summarizes temporal dependence.
The standardized separation of the mean scores is
\begin{equation}
 z(T)=\frac{\delta L\sqrt{T_{\mathrm{eff}}}}{\sigma T}.
 \label{eq:avc_dilution}
\end{equation}
For independent cells, $T_{\mathrm{eff}}=T$, giving
$z(T)=\delta L/(\sigma\sqrt T)$. With fixed event length, adding authentic
context reduces this mean-score separation. Under an additional
equal-variance Gaussian score model,
$\operatorname{AUROC}=\Phi(z/\sqrt2)$, where $\Phi$ is the standard
normal cumulative distribution function.

At the opposite extreme, triggering on any local alarm creates more
opportunities for false positives. For $N$ independent decisions with
local false-positive probability $q$,
\begin{equation}
 \Pr(\text{one or more false alarms})=1-(1-q)^N
 \simeq Nq,\qquad Nq\ll1.
 \label{eq:avc_alarm}
\end{equation}
Here, $N$ counts local decisions, not necessarily independent video frames;
temporal correlations change the effective number of alarm opportunities.
Mean aggregation can dilute a fixed-length event, whereas unrestricted
local triggering can accumulate false alarms as context grows. These
stylized cases motivate retaining evidence at event scale while evaluating
false alarms at video scale. NFSR follows this design through temporal
proposals and source-validation calibration of the resulting video scores.

\FloatBarrier
\section{Generalization and Sensitivity Analyses}
\label{app:additional_experiments}
\label{app:avc_additional}
We analyze training variability, supervision coverage, external transfer,
and robustness to degraded inputs.
\subsection{Seed Stability}
We repeat the shared-feature comparison across three seeds to assess
whether the readout advantage persists under training variation.
\begin{table}[!htbp]
\centering
\AVCtablefont
\caption{Three-seed stability: mean $\pm$ sample standard deviation.
ID and OOD localization are evaluated on the native audio and visual
tracks. OOD AUROC averages the three OOD settings equally; OOD AV-mAP,
A-mAP, and V-mAP follow the OOD-All definitions of
Table~\ref{tab:nfsr_ablation}.}
\label{tab:avc_seeds}
\begingroup
\adjustbox{max width=\linewidth}{%
\begin{tabular}{@{}lrrrrrrrrrr@{}}
\toprule
\rowcolor{AVCTableHead}
Method & ID AUC & ID AV-mAP & ID A-mAP & ID V-mAP & ID A-IoU & ID V-IoU & OOD AUC & OOD AV-mAP & OOD A-mAP & OOD V-mAP \\
\midrule
ActionFormer (TASLE) & 67.94$\pm$0.15 & 5.06$\pm$2.59 & -- & 10.13$\pm$5.18 & -- & 5.56$\pm$2.25 & 72.70$\pm$0.05 & 21.06$\pm$8.31 & -- & 37.07$\pm$14.00 \\
UMMAFormer + Qwen3 & 89.18$\pm$0.29 & 84.76$\pm$0.19 & 89.90$\pm$0.22 & 79.62$\pm$0.16 & 79.77$\pm$1.90 & 65.25$\pm$0.20 & 70.33$\pm$0.80 & 43.50$\pm$0.64 & 69.94$\pm$1.50 & 28.36$\pm$1.17 \\
NFSR + Qwen3 & 96.04$\pm$0.10 & 91.30$\pm$0.13 & 92.24$\pm$0.16 & 90.36$\pm$0.28 & 84.72$\pm$0.55 & 76.90$\pm$0.78 & 84.61$\pm$0.12 & 62.45$\pm$0.08 & 73.66$\pm$0.79 & 53.10$\pm$0.56 \\
\bottomrule
\end{tabular}
}
\endgroup
\end{table}
NFSR's ID advantage over UMMAFormer (6.54 AV-mAP points, with 2.34 A-mAP
and 10.74 V-mAP points) exceeds the standard deviations of these mAP
values, which are at most 0.28 (Table~\ref{tab:avc_seeds}).
The OOD localization advantage (18.95 AV-mAP points, with 3.72 A-mAP and
24.74 V-mAP points) likewise exceeds the observed standard deviations of
at most 1.50, supporting the stability of the shared-feature comparison.
The TASLE-trained ActionFormer has stable detection AUROC but localization
that varies substantially across seeds, particularly under distribution
shift.

\subsection{Value of AV-FABLE Supervision}
\label{app:avc_corpus}
We compare visual-only NFSR readouts trained on TASLE and on AV-FABLE to
examine the effect of the training corpus on visual-event localization.
Both readouts keep only the visual branch and are trained with visual
supervision alone, matching TASLE, which contains only visual edits; their
values therefore differ from those of the audio-visual NFSR in the main
tables.
Both are evaluated on AV-FABLE and TASLE; DAV-Det provides a separate
test of adapting a video-authenticity detector to AV-FABLE.
\begin{table}[!htbp]
\centering
\AVCtablefont
\caption{Training-corpus comparison (\%). All rows report visual-track
detection AUROC, which separates videos with visual events from pristine
videos, and V-mAP. Models are trained separately on the listed
corpora; both NFSR rows are visual-only readouts, not the audio-visual model
of Table~\ref{tab:nfsr_id}. OOD denotes the OOD pool. TASLE cells are
AUROC / visual mAP.}
\label{tab:avc_corpus}
\begingroup
\adjustbox{max width=\linewidth}{%
\begin{tabular}{@{}lrrrrl@{}}
\toprule
\rowcolor{AVCTableHead}
Model / training corpus & ID AUC & ID V-mAP & OOD AUC & OOD V-mAP & TASLE \\
\midrule
NFSR (visual-only) / TASLE & 73.2 & 25.2 & 83.0 & 38.8 & 98.30 / 93.71 \\
NFSR (visual-only) / AV-FABLE & 96.9 & 91.0 & 87.5 & 60.3 & 84.4 / 59.0 \\
DAV-Det / MVAD & 52.77 & 0.21 & 64.98 & 0.84 & 78.87 / 21.00 \\
DAV-Det / AV-FABLE & 86.33 & 2.21 & 68.73 & 2.98 & 70.46 / 17.76 \\
\bottomrule
\end{tabular}}
\endgroup
\end{table}
Corpus coverage strongly affects transfer (Table~\ref{tab:avc_corpus}).
The TASLE-trained visual readout performs well on TASLE but reaches only
25.2 visual mAP on AV-FABLE ID events. AV-FABLE supervision raises
this to 91.0 and OOD-pool mAP from 38.8 to 60.3, while transferring
to TASLE at 59.0 mAP without target-dataset training. AV-FABLE supervision
thus transfers to an external benchmark, although TASLE-trained models
remain stronger on their own benchmark. Adapting DAV-Det mainly improves
ID detection (52.77 to 86.33 AUROC), with a smaller OOD gain and lower
TASLE scores, while its visual mAP remains below 3 on AV-FABLE. The
localization benefit of AV-FABLE supervision therefore also depends on the
prediction model.

\subsection{Training-Set Scale}
We vary the fraction of AV-FABLE training data while retaining the
same epoch schedule. This measures the practical benefit of a larger
training resource under a fixed-epoch schedule.
\begin{table}[!htbp]
\centering
\AVCtablefont
\caption{Training-fraction sensitivity: native-track mAP (\%). Fractions use the
same epoch count, so sample count and optimizer updates vary together.
OOD denotes the OOD pool. Entries are single-run results.}
\label{tab:avc_scale}
\begin{tabular}{@{}lrrrr@{}}
\toprule
\rowcolor{AVCTableHead}
Training fraction & ID A & OOD A & ID V & OOD V \\
\midrule
5\% & 55.7 & 16.8 & 5.6 & 5.1 \\
10\% & 70.5 & 50.4 & 56.1 & 17.4 \\
25\% & 83.6 & 74.7 & 80.0 & 39.3 \\
50\% & 88.3 & 79.8 & 85.7 & 51.2 \\
100\% & 92.1 & 84.6 & 90.7 & 55.2 \\
\bottomrule
\end{tabular}
\end{table}
Both tracks improve throughout the sweep (Table~\ref{tab:avc_scale}),
with particularly weak visual localization at the smallest fraction.
Increasing training from 50\% to the full set still raises pooled OOD
audio/visual mAP by 4.8/4.0 points. Additional data and the associated
optimizer updates therefore remain useful beyond the initial gains
obtained with a small subset.

\paragraph{Generator coverage.}
In three generator-holdout folds, the held-out generator's route remains
covered by other generators in training, yet localization mAP drops by
16.6, 17.0, and 32.8 points. Coverage of an editing route therefore does
not by itself ensure transfer to the held-out generators tested here.

\subsection{External Comparison with Official MSLoc}
\label{app:avc_external}
To test transfer under another benchmark's task definition, we evaluate
MSLoc-PG and its released refinement checkpoint
\citep{feng2026tasle} alongside AV-FABLE-trained NFSR
on the same 1,442 TASLE/TVSum videos with the official evaluator.
$F1_{\mathrm{Det}}$ denotes its \texttt{Loc\_F1} on a 0.01-second grid;
$F1_{\mathrm{Loc}}$ averages interval F1 at tIoU 0.1/0.3/0.5/0.7.
These are distinct from AV-FABLE video-level detection and mAP.
\begin{table}[!htbp]
\centering
\AVCtablefont
\caption{Cross-dataset evaluation under the official TASLE protocol:
$F1_{\mathrm{Det}}/F1_{\mathrm{Loc}}$ (\%). Counts are manipulated/pristine.
Commercial subsets overlap. Evaluation uses the full TASLE official test;
the main source-transfer evaluation uses a separate source-disjoint subset.}
\label{tab:avc_external}
\begingroup
\adjustbox{max width=\linewidth}{%
\begin{tabular}{@{}lrrr@{}}
\toprule
\rowcolor{AVCTableHead}
Subset & NFSR & MSLoc-PG & PG + refine \\
\midrule
TASLE official (631/662) & 51.01 / 50.80 & 67.24 / 63.62 & 36.14 / 36.93 \\
Commercial pool (208/662) & 46.33 / 47.40 & 58.14 / 49.28 & 43.79 / 47.09 \\
Contains Kling/Jimeng (94/662) & 44.10 / 46.61 & 50.55 / 36.46 & 35.77 / 38.92 \\
Pure Kling/Jimeng (78/662) & 48.02 / 54.03 & 42.26 / 30.85 & 35.23 / 39.76 \\
TVSum (72/77) & 38.42 / 43.75 & 48.01 / 38.96 & 36.24 / 39.22 \\
\bottomrule
\end{tabular}
}
\endgroup
\end{table}
MSLoc-PG leads on its TASLE source test, whereas NFSR leads both
metrics on the pure Kling/Jimeng subset and localization F1 on TVSum
(Table~\ref{tab:avc_external}). NFSR thus transfers without TASLE
training, while source-trained MSLoc-PG remains stronger on the full
test. The released refinement checkpoint does not improve aggregate
TASLE scores in this evaluation.

\subsection{Robustness to Input Degradation}
\label{app:avc_robustness}
We test whether degradation of one input track disrupts localization
on that track or propagates to the other. Thresholds remain fixed by
clean source-domain calibration. Table~\ref{tab:avc_stress} reports the
most severe tested condition from each corruption family on 512 videos
(256 pairs), and missing-modality results on the 2,004-video OOD pool.
Each block includes its own clean reference.
\begin{table}[!htbp]
\centering
\AVCtablefont
\caption{Robustness to media corruption and missing modalities (\%).
Recall and FAR use fixed source-domain thresholds. Each evaluation
population has its own unperturbed reference.}
\label{tab:avc_stress}
\begingroup
\adjustbox{max width=\linewidth}{%
\begin{tabular}{@{}lrrrrrr@{}}
\toprule
\rowcolor{AVCTableHead}
Condition & AUROC & Recall & FAR & Attr. F1 & A-mAP & V-mAP \\
\midrule
\multicolumn{7}{l}{\textit{Media corruptions: 512-video subset}} \\
Clean & 88.6 & 58.2 & 2.7 & 61.0 & 85.6 & 53.1 \\
H.264 CRF 35 & 85.1 & 46.9 & 2.0 & 54.0 & 85.6 & 26.0 \\
360p & 80.7 & 40.6 & 1.2 & 47.1 & 85.6 & 10.9 \\
AAC 48 kb/s & 83.4 & 39.8 & 2.3 & 56.0 & 65.5 & 53.1 \\
Audio noise 0.2 & 76.2 & 30.1 & 4.3 & 42.8 & 29.1 & 53.1 \\
Gain $-12$ dB & 87.3 & 53.1 & 2.3 & 60.7 & 81.9 & 53.1 \\
A/V offset 500 ms & 88.5 & 56.6 & 3.1 & 61.8 & 82.3 & 53.1 \\
\multicolumn{7}{l}{\textit{Missing modalities: 2,004-video OOD pool}} \\
Audio + visual & 88.9 & 57.6 & 3.0 & 58.5 & 84.6 & 55.2 \\
Audio missing & 67.0 & 18.5 & 1.8 & 32.6 & 0.4 & 55.2 \\
Visual missing & 81.2 & 39.6 & 1.3 & 40.0 & 84.6 & 3.9 \\
\bottomrule
\end{tabular}
}
\endgroup
\end{table}
Single-track degradation affects the corresponding branch: 360p
reduces visual mAP from 53.1 to 10.9 without changing audio mAP;
audio noise similarly preserves visual localization. Removing a modality
also preserves the other track, consistent with separate evidence paths.
Gain changes and A/V offsets are less damaging in the tested ranges.
Reduced recall can coexist with low FAR, so fewer false alarms alone
do not imply that manipulated events remain detectable.

\subsection{Cross-Track False Alarms}
\label{app:avc_cross_track}
Audio-only speech replacement provides a paired-input control for modality
specificity. In the training set, speech replacements co-occur with
visual edits; the control instead changes the soundtrack while preserving
the visual content. A visual prediction should therefore respond in the
same way to each edited video and its pristine twin.

We compare NFSR's separated readouts with a cross-modal readout whose audio
and visual branches exchange information through gated cross-modal
attention at each shared pyramid level; each model uses its own
source-calibrated 5\% threshold. The separated visual readout fires on
2.9\% of both audio-edited videos and pristine twins for all three TTS
systems. The cross-modal readout's visual head instead fires on 4.6\% of
pristine twins but 11.8--15.1\% of audio-only edits. This sensitivity to
soundtrack changes indicates cross-modal coupling in the cross-modal readout's
visual decisions. Separate branches preserve visual specificity when
manipulation labels co-occur during training.

\FloatBarrier
\section{Language Verification and Temporal Decisions}
\label{app:avc_language}
We examine whether language judgments improve temporal prediction
through direct outputs, fixed-candidate verification, or score fusion.
These analyses use an earlier training release that contains event-sound
and ambience manipulations but no speech replacement, so HDTF here
combines source and manipulation-type shifts.
Event prediction, verification, and rescoring use event F1, classification
AP, and candidate-set localization AP, respectively.

\subsection{Zero-Shot and Direct-Transfer Results}
We first compare direct language outputs with a trained temporal readout
on frozen BEATs+CLIP features. This tests whether transferring perceptual
features to temporal prediction is useful beyond the Qwen-based readouts
in the main comparison.
\begin{table}[!htbp]
\centering
\AVCtablefont
\caption{Zero-shot (ZS) and direct-transfer (DT) evaluation (earlier release).
Triples are detection Macro-F1 / visual Event-F1@0.5 / audio Event-F1@0.5
(\%). HDTF combines source and manipulation-type shifts in this protocol.}
\label{tab:avc_direct_transfer}
\begingroup
\adjustbox{max width=\linewidth}{%
\begin{tabular}{@{}lrrrrr@{}}
\toprule
\rowcolor{AVCTableHead}
Method & Source & Generator & TI2V & OOD pool & Attr. F1 \\
\midrule
Qwen2.5-Omni (ZS) & 46.3 / 0.0 / 3.4 & 45.1 / 4.4 / 3.8 & 46.3 / 3.7 / 2.2 & 46.0 / 4.0 / 2.8 & 8.0 \\
Qwen3-Omni (ZS) & 35.6 / 0.0 / 0.5 & 42.9 / 0.6 / 2.3 & 42.4 / 2.4 / 0.4 & 42.6 / 1.7 / 1.0 & 18.6 \\
video-SALMONN 2+ (ZS) & 15.5 / 0.0 / 0.0 & 1.2 / 0.0 / 0.0 & 10.5 / 0.0 / 0.0 & 7.9 / 0.0 / 0.0 & 5.6 \\
MSLoc-PG (DT) & 35.1 / 0.0 / -- & 52.4 / 11.6 / -- & 43.4 / 4.6 / -- & 46.6 / 7.6 / -- & -- \\
BEATs+CLIP / joint readout & 38.4 / 22.2 / 3.5 & 70.2 / 48.6 / 53.8 & 54.8 / 51.0 / 3.4 & 60.2 / 49.9 / 20.5 & 41.9 \\
\bottomrule
\end{tabular}}
\endgroup
\end{table}
Direct language outputs yield low event F1 across shifts
(Table~\ref{tab:avc_direct_transfer}). The BEATs+CLIP readout recovers
more events, but audio transfer is weaker on unseen HDTF speech
replacements than generator-shifted event sounds. In this protocol,
transfer varies with both the prediction path and training manipulation
coverage. The
MSLoc row evaluates proposals only; Table~\ref{tab:avc_external}
evaluates the complete released pipeline.

\subsection{Fixed-Candidate Verification}
Holding candidate intervals fixed separates the ability to judge an event
from the ability to propose its boundaries. We compare zero-shot
verification, candidate-slot SFT, and decision-stage GRPO~\citep{shao2024deepseekmath} on 130
development and 228 OOD-probe examples. Candidate authenticity AP
measures verification quality; integration into the temporal pipeline
tests whether better verification improves the final localized events.
\begin{table}[!htbp]
\centering
\AVCtablefont
\caption{Candidate verification and complete-system evaluation (\%).
Top: slot AP is audio / visual / mean, and video metrics are AUROC / AP.
Bottom: each track reports mAP / Event-F1@0.5 / positive-video mIoU /
pristine FAR. Mean mAP averages the two native-track mAP values.}
\label{tab:avc_verification}
\begingroup
\adjustbox{max width=\linewidth}{%
\begin{tabular}{@{}lrrrr@{}}
\toprule
\rowcolor{AVCTableHead}
Verifier & Dev slot AP & Dev video & OOD slot AP & OOD video \\
\midrule
Zero-shot & 56.9 / 24.3 / 40.6 & 71.7 / 75.2 & 24.7 / 53.5 / 39.1 & 64.1 / 63.1 \\
Slot SFT & 92.9 / 76.6 / 84.7 & 89.0 / 89.8 & 44.5 / 62.3 / 53.4 & 73.3 / 71.9 \\
Decision GRPO & 92.8 / 81.5 / 87.1 & 91.0 / 91.3 & 43.8 / 63.4 / 53.6 & 72.0 / 71.6 \\
\midrule
System & Visual metrics & Audio metrics & Mean mAP & Attr. F1 / Acc. \\
\midrule
Verifier off & 69.08 / 65.23 / 54.22 / 4.13 & 78.15 / 73.79 / 67.94 / 7.87 & 73.61 & 66.56 / 77.79 \\
+ Zero-shot & 69.08 / 65.23 / 54.22 / 4.13 & 78.15 / 73.52 / 66.84 / 7.68 & 73.61 & 66.41 / 77.59 \\
+ Slot SFT & 69.10 / 64.86 / 53.79 / 4.33 & 78.10 / 72.53 / 64.50 / 6.50 & 73.60 & 66.23 / 77.18 \\
+ Decision GRPO & 69.08 / 65.23 / 54.22 / 4.13 & 78.17 / 72.44 / 65.36 / 7.48 & 73.62 & 66.71 / 77.48 \\
\bottomrule
\end{tabular}}
\endgroup
\end{table}
SFT raises mean OOD candidate AP from 39.1 to 53.4; GRPO adds a
development-set gain but little OOD improvement
(Table~\ref{tab:avc_verification}). Complete-system mean mAP remains
within 73.60--73.62 despite improved candidate discrimination. Better
candidate verification therefore does not translate into better
localization: the verifier can only rescore candidates whose boundaries
and coverage are fixed by the temporal readout.

\subsection{Language Rescoring and Score Fusion}
We next retain the candidate intervals and compare their numerical
readout scores with language-model scores, within-video standardization,
and weighted fusion. This tests whether language judgments provide
useful ranking information even when they do not change event boundaries.
\begin{table}[!htbp]
\centering
\AVCtablefont
\caption{Candidate-set AP at tIoU 0.5 (earlier release, \%). Readout:
numerical readout score; Readout-z: within-video standardized readout
score; LM: Qwen3-Omni language-model score; Equal: equal-weight fusion;
z+LM: standardized fusion; 0.8-Readout: fusion with readout weight 0.8.}
\label{tab:avc_rescoring}
\begingroup
\adjustbox{max width=\linewidth}{%
\begin{tabular}{@{}llrrrrrr@{}}
\toprule
\rowcolor{AVCTableHead}
Slice & Track & Readout & Readout-z & LM & Equal & z+LM & 0.8-Readout \\
\midrule
ID & A & 91.4 & 80.9 & 35.6 & 88.6 & 89.1 & 90.8 \\
ID & V & 92.9 & 71.5 & 50.0 & 88.6 & 87.7 & 92.0 \\
OOD pool & A & 42.8 & 46.3 & 15.5 & 37.2 & 51.4 & 41.6 \\
OOD pool & V & 63.6 & 41.1 & 31.3 & 56.6 & 56.5 & 63.2 \\
HDTF & A & 54.4 & 61.2 & 19.0 & 40.2 & 59.6 & 47.5 \\
ANet OOD & A & 40.6 & 17.9 & 10.8 & 35.8 & 34.7 & 39.7 \\
ANet OOD & V & 65.3 & 48.2 & 32.6 & 57.3 & 58.1 & 64.5 \\
Omni2Sound & A & 43.0 & 20.0 & 16.7 & 38.3 & 40.5 & 42.2 \\
Wan-FLF2V & V & 47.9 & 15.9 & 30.3 & 45.4 & 40.4 & 48.2 \\
OmniWeaving & V & 32.4 & 13.0 & 18.7 & 28.5 & 24.1 & 32.6 \\
TI2V & V & 65.1 & 36.1 & 35.0 & 58.1 & 54.3 & 64.5 \\
\bottomrule
\end{tabular}}
\endgroup
\end{table}
Numerical readout scores outperform language scores on every slice,
and equal-weight fusion lowers AP throughout
(Table~\ref{tab:avc_rescoring}). Standardized fusion improves pooled OOD
audio AP from 42.8 to 51.4 but reduces both ID tracks and OOD visual
AP. Language judgments supplement particular domains without providing
a consistently better ranking rule.

\section{Explanation Baselines}
\label{app:avc_explanation}
We evaluate zero-shot and LoRA-tuned Qwen3-Omni explanation baselines
conditioned on fixed events. We measure output validity, temporal agreement with reference
intervals, operation/target identification, and abstention.
These results show that event-conditioned explanation remains challenging:
output validity and content correctness diverge,
semantic identification remains low, and the tuned model abstains on
most manipulated inputs.
\subsection{Training and Evaluation Setup}
For a conditioning event $(m_i,b_i,e_i,c_i)$ in a video of duration $T$,
the Qwen3-Omni Thinker receives its modality, interval, confidence, and
the audio-visual crop
\begin{equation}
 W_i=[\max(0,b_i-\delta),\;\min(T,e_i+\delta)],
 \qquad \delta=2\,\mathrm{s}.
 \label{eq:explanation_window}
\end{equation}
Here, $b_i,e_i$ are absolute timestamps supplied by the selected candidate
or the annotated event, depending on the evaluation condition. Clipping
keeps the context window within the video, while both modalities remain
available for contextual explanation. A structured response carries
event identity, track, absolute start/end times, operation, target type,
inside/context evidence, judgment, and a support verdict.
Identity, modality, and interval are copied from the conditioning event;
complete abstention requires the designated none-valued semantic fields
and a false support verdict.
Manipulated-event examples supply evidence records; their corresponding
pristine windows supply structured abstention targets. LoRA training updates
the language adapter independently of NFSR's temporal readouts.
Generated text leaves predicted intervals, confidence scores,
detection decisions, and modality attribution unchanged.

SFT uses 4,000 examples (2,000 event--pristine pairs from 1,921 hosts),
500 steps, and LoRA rank eight/alpha 32 on all linear layers
\citep{hu2022lora}. Supervision is drawn from the 13,628 training events
with candidate evidence records and their corresponding pristine windows. Training and evaluation hosts are disjoint.

Evaluation uses 64 ID pairs and 64 OOD pairs with duration at most
240 seconds, stratified by track, source, and generator.
Zero-shot and SFT-predicted modes receive identical NFSR candidates.
GT conditioning supplies annotated intervals to evaluate explanation
generation with ground-truth temporal input.
Strict schema scoring checks the full raw JSON. Events match on track
and tIoU at least 0.5; operation/target macro-F1 retains parse failures
in the fixed matched-event denominator. Temporal agreement requires
both endpoints within one second of ground truth. Clean abstention
measures complete abstention on pristine inputs, whereas clean support
counts erroneous support verdicts; malformed or incomplete outputs can
satisfy neither condition.
\begin{table}[!htbp]
\centering
\AVCtablefont
\caption{Event-conditioned explanation. Scores are percentages except
matched-event counts.
Op./target denote operation/target-type Macro-F1; time measures boundary
agreement with the reference interval. Clean support denotes erroneous
support verdicts on pristine videos. SFT-GT conditions on annotated
intervals, separating explanation evaluation from predicted localization.}
\label{tab:avc_explanation}
\begingroup
\adjustbox{max width=\linewidth}{%
\begin{tabular}{@{}lrrrrrrr@{}}
\toprule
\rowcolor{AVCTableHead}
Mode & Schema & Matched & Op./target & Time & Clean abst. & Clean support & Fake abst. \\
\midrule
Zero / ID & 82.81 & 62 / 64 & 6.00 / 11.04 & 72.58 & 20.31 & 1.56 & 18.75 \\
Zero / OOD & 91.41 & 49 / 64 & 3.10 / 19.85 & 87.76 & 17.19 & 3.13 & 17.19 \\
SFT / ID & 99.22 & 62 / 64 & 16.38 / 20.91 & 79.03 & 87.50 & 10.94 & 67.19 \\
SFT / OOD & 100.00 & 49 / 64 & 4.44 / 15.11 & 93.88 & 93.75 & 6.25 & 76.56 \\
SFT-GT / ID & 100.00 & 64 / 64 & 13.91 / 13.47 & 100.00 & 89.06 & 10.94 & 76.56 \\
SFT-GT / OOD & 100.00 & 64 / 64 & 1.28 / 13.64 & 100.00 & 92.19 & 7.81 & 79.69 \\
\bottomrule
\end{tabular}
}
\endgroup
\end{table}
\subsection{Structured Output and Semantic Identification}
SFT produces valid raw structured outputs on 99.22\% of ID and
100.00\% of OOD inputs (Table~\ref{tab:avc_explanation}). On ID events,
operation/target F1 increases from 6.00/11.04 to 16.38/20.91 relative
to zero-shot generation. The corresponding OOD results are mixed:
operation F1 increases slightly while target F1 decreases. Format
compliance and event-label accuracy therefore respond differently to
supervision.

\paragraph{Abstention and event support.}
SFT raises pristine abstention to 87.50\% on ID and 93.75\% on OOD,
but also abstains on 67.19\% and 76.56\% of manipulated inputs.
It thus learns strong abstention without reliably distinguishing
unsupported candidates from manipulated events.

\paragraph{Conditioning on annotated intervals.}
With annotated events, SFT-GT achieves complete matching and 100\%
boundary agreement by construction, while operation/target F1 remains
low and abstention frequent. Accurate temporal conditioning alone is
therefore insufficient for semantic identification.

\paragraph{Prompt and formatting sensitivity.}
Zero-shot schema failures stem from outer Markdown code fences; removing
them makes every zero-shot response schema-compliant without changing its
content. Clarifying numeric types and absolute timestamps in the SFT
prompt yields no reliable operation-F1 change ($+3.44$ on ID,
pair-bootstrap interval $[-0.84,10.67]$; $0.00$ on OOD, $[-4.52,4.43]$).

\subsection{Media Dependence of Support Decisions}
\label{app:avc_media_dependence}
To test whether support decisions depend on the supplied media, we
compare three input conditions using the same SFT-GT model and the
64 ID and 64 OOD twin pairs. Prompts, event identities, modalities,
and annotated intervals are held fixed. The original condition supplies
the event-centered crop. The shifted condition supplies an event-free
window from the same video, using matching windows in each twin;
shorter windows are padded with repeated terminal frames and silence
to preserve crop duration. The target-track-masked condition retains
the original crop but replaces the queried modality with silence or
black frames, leaving the other modality unchanged.

\begin{table}[!htbp]
\centering
\AVCtablefont
\caption{\textbf{Support decisions under media interventions.}
Rates (\%) are measured separately on the manipulated and pristine
members of each twin pair, with 64 pairs per split. Support and complete
abstention require valid structured outputs and unchanged conditioning
fields. Original crops use the SFT-GT setting of
Table~\ref{tab:avc_explanation}.}
\label{tab:avc_media_dependence}
\begingroup
\adjustbox{max width=\linewidth}{%
\begin{tabular}{@{}lrrrr@{}}
\toprule
\rowcolor{AVCTableHead}
& \multicolumn{2}{c}{Manipulated-twin support}
& \multicolumn{2}{c}{Pristine-twin abstention} \\
\cmidrule(lr){2-3}\cmidrule(l){4-5}
Media condition & ID & OOD & ID & OOD \\
\midrule
Original crop & 23.44 & 20.31 & 89.06 & 92.19 \\
Event-free shifted crop & 9.38 & 4.69 & 92.19 & 98.44 \\
Target track masked & 0.00 & 0.00 & 100.00 & 100.00 \\
\bottomrule
\end{tabular}}
\endgroup
\end{table}

Replacing the event crop with an event-free window reduces support
by 14.06 percentage points on ID and 15.63 on OOD
(Table~\ref{tab:avc_media_dependence}). The respective 95\% paired-bootstrap
intervals are $[-28.13,-1.56]$ and $[-26.56,-6.25]$, using 1,000
resamples of twin pairs shared across input conditions. Masking the
queried track eliminates support on both splits. Thus, the supplied
event prompt alone does not determine the support verdict, which is
sensitive to event content and the availability of the queried modality.
Sensitivity to the media does not establish that the verdicts or evidence
statements are correct; the operation/target scores in
Table~\ref{tab:avc_explanation} remain low.

\section{Qualitative Examples}
\label{app:avc_cases}
\begin{figure}[!htbp]
\centering
\includegraphics[width=\linewidth]{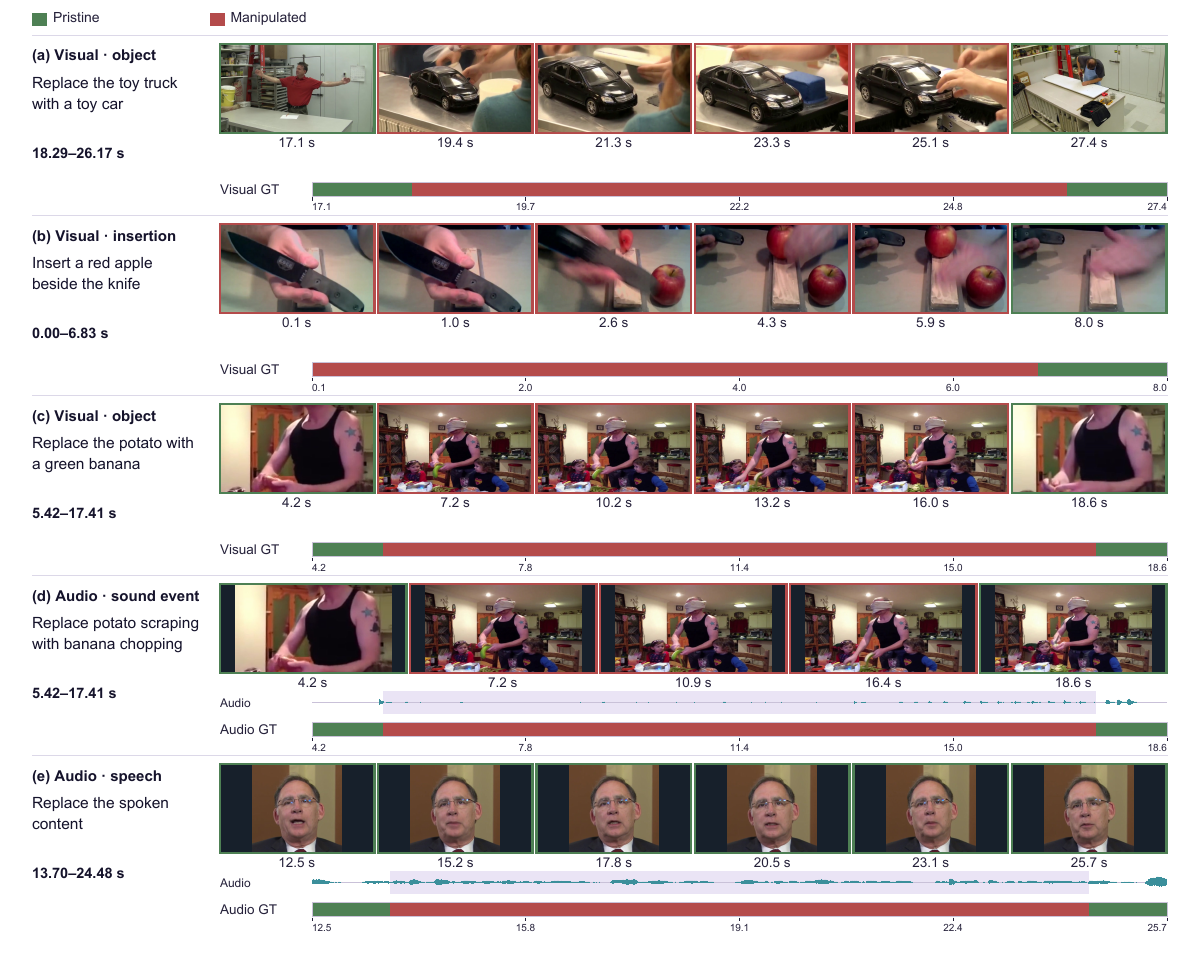}
\caption{Manipulation examples in AV-FABLE. Each row gives a short
description of the edit, the annotated interval, frames sampled across the
event boundaries, and the local ground-truth (GT) timeline of the
manipulated track; audio rows add the manipulated waveform with the event
shaded. Green and red denote pristine and manipulated content, and frame
borders follow the visual track. Rows (c) and (d) are co-timed events in
one both-track video.}
\label{fig:avc_types}
\end{figure}

Figure~\ref{fig:avc_types} illustrates manipulated events on both tracks:
object replacement and insertion on the visual track, and sound-event and
speech replacement on the audio track. Rows (c) and (d) show co-timed
visual and audio events in the same both-track video.

\begin{figure}[!htbp]
\centering
\includegraphics[width=\linewidth]{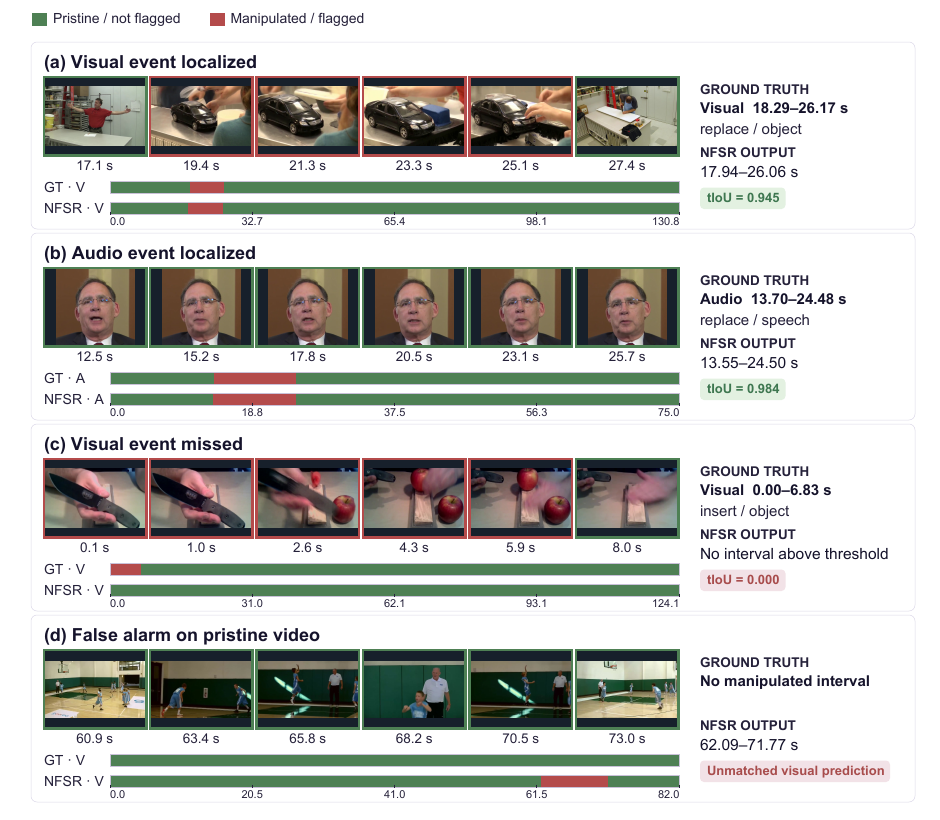}
\caption{Localization successes and errors. (a) A visual-only object
replacement and (b) an audio-only speech replacement are localized;
(c) a visual object insertion is missed; (d) a pristine video yields a
visual false alarm. Each row shows sampled frames, full-video GT and NFSR
timelines of the displayed track, the annotated and predicted intervals,
and the best tIoU. NFSR intervals are those above the fixed track
threshold; frame borders follow the visual-track GT.}
\label{fig:avc_localization}
\end{figure}

\begin{figure}[!htbp]
\centering
\includegraphics[width=\linewidth]{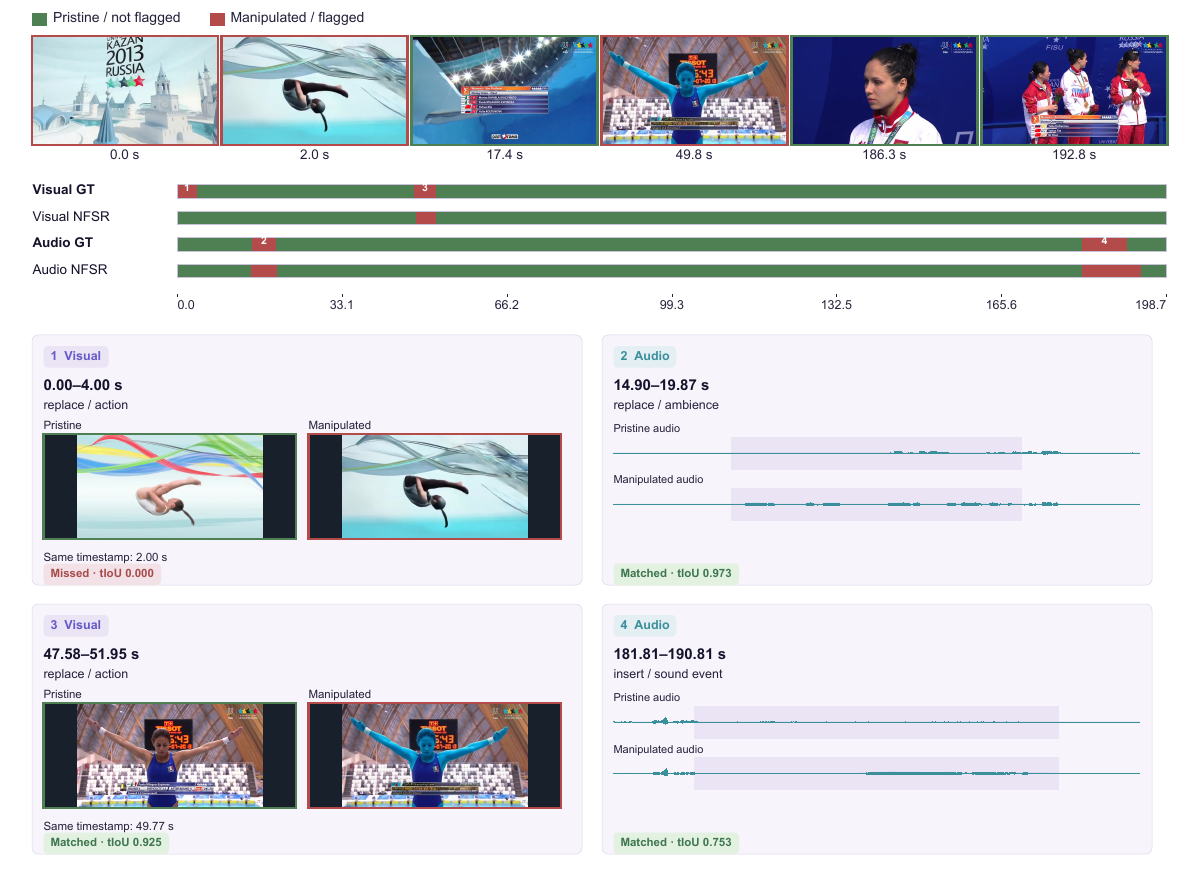}
\caption{Four asynchronous events in one both-track video. Top: sampled
frames and full-video GT and NFSR timelines for each track, with numbered
GT events. Bottom: each event's interval and operation/target, pristine
and manipulated frames at the same timestamp (visual) or pristine and
manipulated waveforms over the event window (audio), and the best tIoU.
NFSR matches events 2--4 (tIoU $\geq 0.5$) and misses event 1.}
\label{fig:avc_multi}
\end{figure}

Figures~\ref{fig:avc_localization} and~\ref{fig:avc_multi} show NFSR
predictions at the fixed source-calibrated track thresholds.
Figure~\ref{fig:avc_localization} contains a localized visual-only object
replacement, a localized audio-only speech replacement, a missed visual
object insertion, and a false alarm on a pristine video.
Figure~\ref{fig:avc_multi} shows a both-track video with four asynchronous
events, two on each track. NFSR matches both audio events and the second
visual event (tIoU 0.753--0.973) but misses the first visual event.

\begin{figure}[!htbp]
\centering
\includegraphics[width=\linewidth]{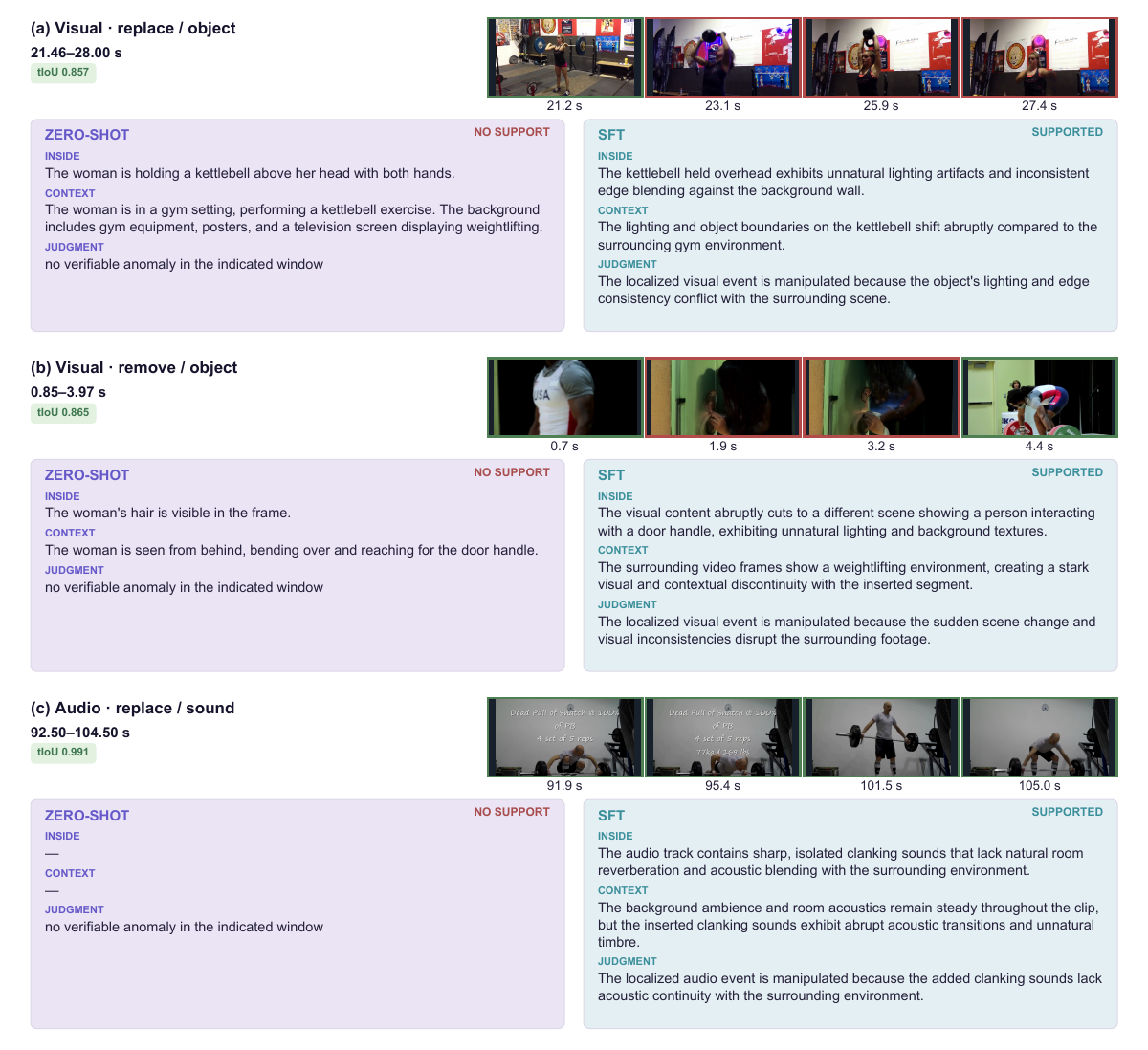}
\caption{Explanation outputs on fixed NFSR candidates. Each example gives
the annotated track, operation, target, and interval, the tIoU of the
candidate supplied to both modes, and sampled frames. Zero-shot and SFT
panels show the support verdict and the generated inside, context, and
judgment fields; dashes mark empty fields.}
\label{fig:avc_explanations}
\end{figure}

Figure~\ref{fig:avc_explanations} shows three ID examples from the
evaluation in Table~\ref{tab:avc_explanation}. Zero-shot and SFT-predicted
modes receive the same top-ranked NFSR candidate, taken before the
video-level decision threshold is applied; candidate tIoU is 0.857--0.991.
Zero-shot generation withholds support in all three examples, whereas SFT
supports each event with inside, context, and judgment fields. These
supported cases are selected for illustration; in aggregate, SFT often
abstains on manipulated inputs (Table~\ref{tab:avc_explanation}). The
displayed fields are model outputs rather than verified evidence; they
illustrate the distinction between locating an event and describing its
forensic evidence.

\FloatBarrier